\documentclass{article}

\usepackage{microtype}
\usepackage{graphicx}
\usepackage{subfigure}
\usepackage{booktabs} 
\usepackage{xcolor}
\usepackage{enumitem}

\usepackage{hyperref}

\usepackage[accepted]{icml2025}

\usepackage{amsmath}
\usepackage{amssymb}
\usepackage{mathtools}
\usepackage{amsthm}

\usepackage[capitalize,noabbrev]{cleveref}

\usepackage{multirow}
\usepackage{graphicx}

\newcommand{\name}{\textsc{{L-STEP}}}

\def\nodefeat#1{\mathbf{s}_{#1}}
\def\edgefeat#1{\mathbf{e}^{#1}}
\def\neighborhood{\mathcal{N}}
\def\adj{\mathbf{A}}
\def\degree{\mathbf{D}}
\def\laplacian{\mathbf{L}}
\def\eigenvecs{\mathbf{U}}
\def\eigenvec{\mathbf{u}}
\def\h{\mathbf{h}}
\def\pe{\mathbf{p}}
\def\mlp{\mathbf{W}}
\def\losslp{\mathcal{L}_{lp}}
\def\losspe{\mathcal{L}_{pe}}
\def\approxpe{\widetilde{\pe}}

\theoremstyle{plain}
\newtheorem{theorem}{Theorem}[section]

\newtheorem{lemma}[theorem]{Lemma}

\theoremstyle{definition}

\theoremstyle{remark}

\usepackage[textsize=tiny]{todonotes}

\icmltitlerunning{Learnable Spatial-Temporal Positional Encoding for Link Prediction}

\begin{document}

\twocolumn[
\icmltitle{Learnable Spatial-Temporal Positional Encoding for Link Prediction}



\icmlsetsymbol{equal}{*}

\begin{icmlauthorlist}
\icmlauthor{Katherine Tieu}{equal,uiuc}
\icmlauthor{Dongqi Fu}{equal,meta}
\icmlauthor{Zihao Li}{uiuc}
\icmlauthor{Ross Maciejewski}{asu}
\icmlauthor{Jingrui He}{uiuc}
\end{icmlauthorlist}

\icmlaffiliation{uiuc}{University of Illinois Urbana-Champaign}
\icmlaffiliation{meta}{Meta AI}
\icmlaffiliation{asu}{Arizona State University}

\icmlcorrespondingauthor{Jingrui He}{jingrui@illinois.edu}

\icmlkeywords{Machine Learning, ICML}

\vskip 0.3in
]



\printAffiliationsAndNotice{\icmlEqualContribution} 

\begin{abstract}
Accurate predictions rely on the expressiveness power of graph deep learning frameworks like graph neural networks and graph transformers, where a positional encoding mechanism has become much more indispensable in recent state-of-the-art works to record the canonical position information.
However, the current positional encoding is limited in three aspects: (1) most positional encoding methods use pre-defined, and fixed functions, which are inadequate to adapt to the complex attributed graphs; (2) a few pioneering works proposed the learnable positional encoding but are still limited to the structural information, not considering the real-world time-evolving topological and feature information; (3) most positional encoding methods are equipped with transformers' attention mechanism to fully leverage their capabilities, where the dense or relational attention is often unaffordable on large-scale structured data.
Hence, we aim to develop \underline{\textbf{L}}earnable \underline{\textbf{S}}patial-\underline{\textbf{T}}emporal \underline{\textbf{P}}ositional \underline{\textbf{E}}ncoding in an effective and efficient manner and propose a simple temporal link prediction model named \textbf{\name}.
Briefly, for \name, we (1) prove the proposed positional learning scheme can preserve the graph property from the spatial-temporal spectral viewpoint, (2) verify that MLPs can fully exploit the expressiveness and reach transformers' performance on that encoding, (3) change different initial positional encoding inputs to show robustness, (4) analyze the theoretical complexity and obtain less empirical running time than SOTA, and (5) demonstrate its temporal link prediction out-performance on 13 classic datasets and with 10 algorithms in both transductive and inductive settings using 3 different sampling strategies. Also, \name\ obtains the leading performance in the newest large-scale TGB benchmark. Our code is available at~\url{https://github.com/kthrn22/L-STEP}.
\end{abstract}

\section{Introduction} 

Link prediction is an important research topic in the graph learning community~\citep{kumar2020link}, especially in complex networks~\citep{DBLP:journals/corr/abs-1010-0725, DBLP:journals/csur/MartinezBT17}. Learning from the time-evolving topological structures and time-evolving node and edge features, and determining whether the link (i.e., edge or interaction between two nodes) exists at a certain timestamp are challenging \cite{tieu2025invariant} but have attracted much research interest from dynamic protein-protein interactions~\citep{taylor2009dynamic, gupta2015dynamic, DBLP:conf/bigdataconf/FuH22, DBLP:conf/cikm/ZhouZF0H22}, recommender systems~\citep{huang2005link, DBLP:conf/iclr/CongZKYWZTM23}, social network analysis~\citep{daud2020applications, DBLP:conf/kdd/FuZH20, DBLP:conf/www/Fu0MCBH23}, and many more, as real-world graphs often exhibit temporality \cite{DBLP:conf/kdd/FuFMTH22, DBLP:conf/www/LiFH23, DBLP:conf/nips/Lin0FQT24, DBLP:conf/nips/BanZLQFKTH24, DBLP:conf/nips/TieuFZHH24, DBLP:conf/kdd/LiFAH25, DBLP:journals/corr/abs-2502-08942}, and heterogeneity \cite{DBLP:conf/icde/LeZWW24, DBLP:conf/icml/0002QZYZ0ZWHT24, DBLP:conf/icml/LeZLXCZX24, DBLP:conf/icml/ZhongLLJYZYZXMX24, DBLP:conf/iclr/0003FTMH25}.

Recently, with the success of the attention mechanism~\citep{DBLP:conf/nips/VaswaniSPUJGKP17}, graph transformers have been proposed for the structured data to obtain the out-performance in many graph applications~\citep{DBLP:conf/nips/RampasekGDLWB22,DBLP:conf/nips/KimNMCLLH22, DBLP:conf/iclr/FuHXFZSWMHL24}.
To preserve the exact ordering information, the corresponding positional encoding plays an important role in boosting the downstream task performance like temporal link prediction. Recent studies show that it can not only help graph transformers~\citep{DBLP:conf/nips/0004S0L23} but also can contribute to graph neural networks~\citep{DBLP:conf/iclr/CongZKYWZTM23}.
In spite of the great potential, the research on positional encoding on graphs is limted from at least three aspects.
\textbf{First}, following~\cite{DBLP:conf/iclr/XuRKKA20}, most positional encoding methods are still based on pre-defined, and fixed functions, which either heavily depends on the domain knowledge of understanding the topological distribution of graphs or not easy to adapt to different complex attributed graphs.
\textbf{Second}, to the best of our knowledge, a pioneering work~\cite{DBLP:conf/iclr/DwivediL0BB22} proposed the learnable position encoding to tackle this problem, but it only considers the static structural information and does not accommodate the realistic time-evolving structural and feature information.
\textbf{Third}, most, if not all, of the positional encoding methods need to rely on the transformer's attention mechanism to fully release the expressiveness power. To be more specific, in order to preserve the long-distance relation into the representation vectors, graph transformers usually need high time complexity due to the dense attention or relational attention~\citep{DBLP:journals/corr/abs-2302-04181, DBLP:journals/corr/abs-2202-08455}, i.e., each pair of nodes need to be attended with $O(n^{2})$ time complexity, which can even approach $O(n^{3})$ in a more complex relational setting~\citep{DBLP:conf/iclr/DiaoL23}, and the prolonged time-aware features also increase the size of neural architectures. A detailed theoretical time complexity and empirical running time comparison with SOTA can be found in Appendix~\ref{sec:time_complexity} and Appendix~\ref{app:scalability}.


In light of the above discussion, we aim to propose learnable positional encoding evolving over time, such that a simple neural architecture can achieve competitive or even the same performance as graph transformers.
Thus, we first propose two major novel techniques: \textbf{Learnable Positional Encoding Module} (LPE) and \textbf{Node-Link-Positional Encoder} based on Discrete Fourier Transform. Then, we combine these two into an end-to-end temporal link prediction framework named \textbf{\name}, as illustrated in Section~\ref{sec: framework}.

To be specific, we first prove that the proposed positional encoding method LPE can preserve the temporal graph topology from the spatial-temporal spectral viewpoint and \name\ having better theoretical runtime complexity compared to SOTAs in Section~\ref{sec:theory}. Moreover, in Section~\ref{sec:exp}, we evaluate the empirical performance of \name\ extensively, including (1) the comprehensive effectiveness comparison with 13 classic datasets, 10 algorithms, 2 learning settings (transductive and inductive), and 3 sampling strategies, where our \name\ performs the best across the board, (2) the verification of the role of MLPs and transformers upon \name, (3) the learning robustness of \name\ in terms of different initial positional encoding inputs, (4) parameter analysis, running time comparison, and ablation studies, and (5) the leading performance in the large-scale TGB open benchmark~\citep{DBLP:journals/corr/abs-2406-09639}.

\section{Preliminaries} 
\label{sec:prelim}
Here, we introduce preliminaries to pave the way for deriving our \name\ in the next section.

\textbf{Temporal Graph Snapshot}. A snapshot of a temporal graph $G$ is a collection of temporal interactions (i.e., edges) at time $t \in \{1, \ldots, T\}$, which is defined as $G^t = \{(u, v, t) \big| u \in V^{t}; v \in V^{t}; (u, v, t) \in E^{t}\}$. Each event $(u, v, t)$ denotes an interaction between nodes $u$ and $v$ that occurs at time $t$, and the node and edge sets of $G^t$ are denoted as $V^t$ and $E^t$ respectively. Additionally, for node $u \in V^t$, we denote the node features as $\nodefeat{u}^t \in \mathbb{R}^{d_N}$, and the edge features associated with an event $(u, v, t)$ as $\edgefeat{t}_{u, v} \in \mathbb{R}^{d_E}$, where $d_N$ and $d_E$ are the dimensions of the node and edge features.

\textbf{Temporal Neighbors}. The temporal neighborhood of a node $u \in V^{t}$ at time $t \in \{1, \ldots, T\}$, denoted as $\neighborhood^{t}_u$, is the set of all interactions that involve $u$ at time $t$. Formally, $\neighborhood^{t}_u$ can be mathematically expressed as $\neighborhood^{t}_u = \{(u, v, t) \big| v \in V^t; (u, v, t) \in E^t\}$.

\textbf{Time Encoder}. A time encoder is a function to obtain the vector representation of time $t$~\citep{DBLP:conf/iclr/XuRKKA20, DBLP:conf/iclr/WangCLL021, DBLP:conf/iclr/CongZKYWZTM23}. Let $d_T$ be the dimension of the time encoding vector. We define our time encoding function as $f_{T}: \mathbb{R}^+ \rightarrow \mathbb{R}^{d_T}$. In \name, for $t \in \mathbb{R}^+$, $f_T(t)=\cos(t \cdot \omega)$, where $\mathbf{\omega} \in \mathbb{R}^{d_T}$ with the $i^{\textrm{th}}$ entry computed as $\mathbf{\omega}_i = \alpha^{-(i - 1) / \beta}$, and $\alpha, \beta \in \mathbb{R}$ are hyper-parameters. In order to enhance the ability of $d_T$ in distinguishing all timestamps, given that $t_{max}$ is the maximum timestamp being considered, $\alpha$ and $\beta$ should be selected in such a way that $t_{max} \cdot \alpha^{-(i - 1) / \beta} \rightarrow 0$ as $i$ goes towards $d_T$. Moreover, $\alpha, \beta$, and $\omega$ are fixed during the training process of \name. 

\textbf{Positional Encoding for Graphs}.
Originated from the transformer~\citep{DBLP:conf/nips/VaswaniSPUJGKP17}, positional encoding is proposed to add node feature information to support the attention mechanism by indicating the relative position of a node within the input graph~\citep{DBLP:journals/jmlr/DwivediJL0BB23}, and the common positional encoding methods include computing the Laplacian eigenvector or Personalized PageRannk vector for each node ~\citep{DBLP:conf/nips/RampasekGDLWB22, DBLP:conf/iclr/ChenGL023}.
Mathematically, given an $n$-node (static) graph $G = (V, E)$ and its adjacency matrix $\mathbf{A} \in \mathbb{R}^{n \times n}$, the normalized Laplacian of $G$ is computed as $\laplacian = \mathbf{I}_n - \degree^{-1/2} \adj \degree^{-1/2}$, where $\mathbf{I}_n$ is the $n \times n$ identity matrix, and $\degree$ is the diagonal degree matrix of $G$. Let the eigen-decomposition of $\laplacian$ be $\laplacian = \eigenvecs \mathbf{\Lambda} \eigenvecs^\top$, where $\mathbf{\Lambda} = \text{diag}(\lambda_1, \lambda_2, \dots, \lambda_n)$ is a diagonal matrix consisting of $\laplacian$'s eigenvalues and $\eigenvecs$ consists of $n$ eigenvectors of $\mathbf{L}$ as its columns. Let $\eigenvec_i \in \mathbb{R}^n$ be the $i^{\textrm{th}}$ eigenvector of $\mathbf{L}$ and its $j^{\textrm{th}}$ entry denoted as $\eigenvec_{i,j}$. Then, the positional encoding of node $i \in V$ is defined as $\pe_i = [\eigenvec_{1, i}, \dots, \eigenvec_{n, i}]^\top \in \mathbb{R}^n$. In practice, the dimension $d_P$ of positional encoding is usually set to be less than $n$ for computational efficiency~\citep{DBLP:journals/corr/abs-2012-09699}, i.e., $d_P \ll n$ and $\pe_i = [\eigenvec_{1, i}, \dots, \eigenvec_{d_P, i}]^\top \in \mathbb{R}^{d_P}$.

\textbf{Discrete Fourier Transform}.
To pave the way for learnable positional encoding, we need to first introduce the basics of Discrete Fourier Transform (DFT)~\citep{sundararajan2001discrete}. In brief, DFT is designed to convert a finite-length sequence of samples into another sequence in the frequency domain (with the same length). Formally, given a sequence with length $N$, $\{\mathbf{x}_j\}_{j = 1}^N$, the DFT operation $\mathcal{F}(\{\mathbf{x}_j\}_{j = 1}^N)$ converts $\{\mathbf{x}_j\}_{j = 1}^N$ to the complex-valued sequence $\{\mathbf{X}_j\}_{j = 1}^N$ as $\mathbf{X}_j = \sum_{k = 1}^N \mathbf{x}_k e^{-i2\pi \frac{j}{N} k}$. Also, to re-construct the original sequence $\{\mathbf{x}_{j}\}_{j = 1}^N$, inverse DFT (IDFT) $\mathcal{F}^{-1}(\{ \mathbf{X}_{j} \}_{j = 1}^N)$ is defined as $\mathbf{x}_j = \sum_{k = 1}^N \mathbf{X}_k e^{i 2\pi \frac{j}{N} k}$.






\section{Proposed Method: \name} 
\label{sec: framework}

To begin with, we focus on a general research question, i.e., \textit{given two nodes $u$ and $v$, how can we predict whether the link (interaction or edge) exists between $u$ and $v$ at a certain timestamp $t$, effectively and efficiently}?

To answer this question, we propose a simple temporal link prediction model named \name. It retrieves the interaction history of two nodes and decides whether the current link exists or not.
Without heavy neural architectures like dense attention~\citep{DBLP:conf/nips/RampasekGDLWB22} or relational attention~\citep{DBLP:conf/iclr/DiaoL23}, \name\ only depends on two of our proposed simple yet effective techniques, i.e., \textit{Learnable Positional Encoding Module} (LPE) and \textit{Node-Link-Positional Encoder}, such that \name\ can preserve the spatial-temporal topology rigorously (as shown in Section~\ref{sec:theory}) and outperform attention-based state-of-the-art link prediction methods (as shown in Section~\ref{sec:exp}).

\begin{figure*}[h]
    \includegraphics[width=1\linewidth]{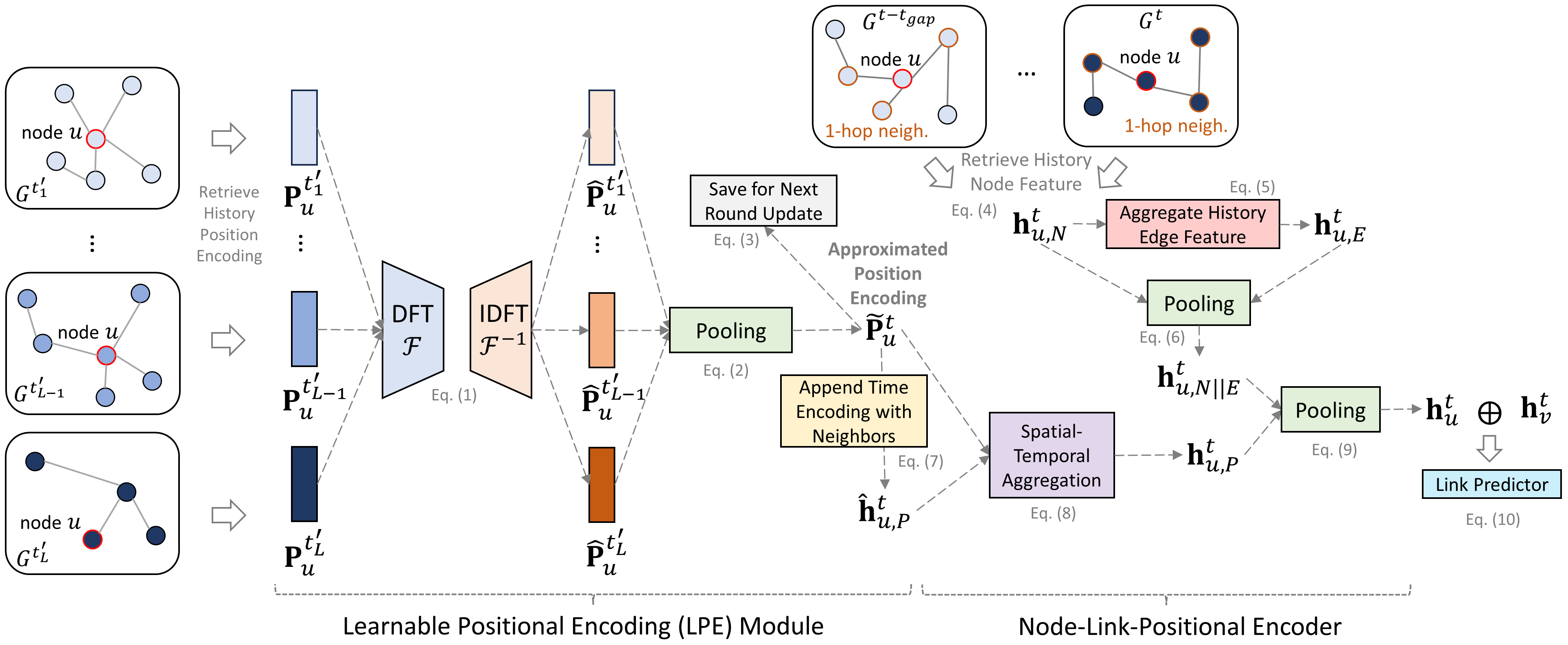}
    \centering
    \caption{Overall Framework of \name}
    \label{fig: framework}
\end{figure*}

The overall framework of \name\ is shown in Figure~\ref{fig: framework} about how to retrieve and learn from history to make the current decision with two proposed techniques. Next, we will introduce the details of the proposed techniques, i.e., \textit{Learnable Positional Encoding Module} in Section~\ref{sec:LPE} and \textit{Node-Link-Positional Encoder} in Section~\ref{sec: encoder}.

\subsection{Learnable Positional Encoding (LPE) Module} \label{sec:LPE}








The main idea of Learnable Positional Encoding (LPE) is to leverage historical positional encoding to determine whether the link exists at the current timestamp.
However, learning from the history data in an effective and efficient way and only through positional encoding is challenging, because SOTA methods usually need a complex attention mechanism~\citep{DBLP:conf/nips/PoursafaeiHPR22, DBLP:conf/nips/0004S0L23}.

To this end, we aim to only use positional encoding to carry informative and useful knowledge to record what had happened in the past (e.g., what interactions occurred and what the node and edge features looked like around the interaction).
Existing simple hand-crafted positional encoding can be efficient but tend to induce suboptimal performance~\citep{DBLP:conf/iclr/WangCLL021, DBLP:conf/iclr/CongZKYWZTM23}.

Thus, we propose a ``\textbf{learnable}'' positional encoding module (LPE), enabling it to reflect complex and non-linear knowledge.
As shown in Figure~\ref{fig: framework}, in our \name, LPE estimates (or approximates) the positional encoding at current time $t$ by applying the Fourier Transform (with learnable parameters) on the past positional encoding. It is worth noting that obtaining the positional encoding through LPE: (1) records the complex spatial-temporal information (proved from the spectral viewpoint in Section~\ref{sec:theory}), (2) and avoids computing the Laplacian eigendecomposition for each timestamp from scratch, which is computationally expensive (e.g., $O(|V|^{3})$~\citep{DBLP:conf/iclr/ChenGL023} or $O(|E|^{3/2})$~\citep{DBLP:journals/jmlr/DwivediJL0BB23})


Because of no peeking at the current graph topology structure and features, our LPE is an iterative updating module. Next, we introduce LPE in two-folds: (1) given the past positional encoding, how LPE approximates the current positional encoding (that will be used to determine links by \textit{Node-Link-Positional Encoder} in Section~\ref{sec: encoder}), and (2) how to store the approximate positional encoding and update/optimize it in the future round with current topology structure and features observed.

\textbf{Approximating current position encoding}.
First of all, we denote the positional encoding of $u$ to be learned for time $t$ as $\pe^t_u \in \mathbb{R}^{d_P}$.
Suppose there are $L$ most recent distinct timestamps prior to $t$, i.e., $t'_1, \dots, t'_L$ (where $t'_1 < \dots < t'_L < t$)
, and we can retrieve the already learned $\{\pe_u^{t'_1}, \dots, \pe_u^{t'_L}\}$ that is regarded as a $L$-length sequence. Additionally, considering computational efficiency, we only retrieve $L$ most recent timestamps. Then, in this LPE module, we leverage the Discrete Fourier Transform as a way to capture the temporal dependencies among $\pe_u^{t'_1}, \dots, \pe_u^{t'_L}$ and derive an estimated positional encoding of $u$ at time $t$ denoted by $\widetilde{\pe}_u^{t} \in \mathbb{R}^{d_P}$.


To be specific, we start by applying Discrete Fourier Transform (DFT) on the sequence $\{\pe^{t'}_u \}_{t' = t'_1}^{t'_L}$ to obtain the corresponding sequence in the frequency domain. Then, we employ a complex-valued learnable filter, $\mlp_{filter} \in \mathbb{C}^{d_P \times L}$, to filter out noises of the sequence in the frequency domain, (detailed discussion on how positional encoding could be regarded as signals and what kind of noises that our LPE module filters can be founded in Appendix~\ref{app:pe-signal-filter}) and finally transform back to the original domain. The process can be mathematically expressed as
\begin{equation}
    \label{eq:transform_pe}
    \{\widehat{\pe}_u^{t'_j} \}_{j= 1}^{L} = \mathcal{F}^{-1}(\mlp_{filter} \odot \mathcal{F}(\{\pe_u^{t'_j} \}_{j = 1}^L))
\end{equation}





Then, we obtain the approximate positional encoding $\widetilde{\pe}_u^{t}$ by applying sum-pooling with learnable weights on $\{\widehat{\pe}_u^{t'_j}\}_{j = 1}^L$
\begin{equation}
\label{eq:approx_pe}
    \widetilde{\pe}_u^{t} = \begin{bmatrix}
        \widehat{\pe}_u^{t'_1}  & \dots & \widehat{\pe}_u^{t'_L}
    \end{bmatrix} \mlp^{sum}_{P}
\end{equation}
where $\mlp^{sum}_{P} \in \mathbb{R}^{L \times 1}$. Note that the approximate positional encoding $\widetilde{\pe}_u^{t}$ will be sent to the \textit{Node-Link-positional Encoder} (in Section~\ref{sec: encoder}) for making predictions for current time $t$.

\textbf{Storing approximate positional encoding for future retrieval and optimization}.
Since \name\ does not peek at the current topology structure and features when making the prediction for now, i.e., only using approximate positional encoding $\widetilde{\pe}_u^{t}$, we elaborate on how the current approximate positional encoding gets updated and optimized in the next round.

After making the current link predictions at time $t$ (as shown in Section~\ref{sec: encoder}), \name\ can then ``peek'' (or ``reveal'') the ground-truth links at time $t$, such that it can update the approximation $\widetilde{\pe}_u^{t}$ into updated $\pe^t_u$ for future retrieval and predictions. 

Then, we need to update the positional encoding of for any node $u$ at $t$. We sample the most $K$ recent interactions, \textit{including those that occur at time $t$ if the node has interactions at time $t$}. Suppose they are indexed as $\{(\bar{u}_1, u, \bar{t}_1), \dots, (\bar{u}_K, u, \bar{t}_K)\}$ where $\bar{t}_1 \leq \dots \leq \bar{t}_K \leq t$. Leveraging the approximate positional encoding from $\bar{u}_1, \dots, \bar{u}_K$, $\pe^{t}_u$ is derived as
\begin{equation}
    \label{eq:pe_updated}
    \pe^{t}_u = \widetilde{\pe}^{t}_{u} + tanh(\mlp^{(self)}_P \widetilde{\pe}^{t}_{u}  + \mlp^{(2)}_{P}\text{RELU}(\mlp^{(1)}_{P} \widehat{\mathbf{\mathbf{q}}}^{t}_{u})) \in \mathbb{R}^{d_P}
\end{equation}
where $\widehat{\mathbf{\mathbf{q}}}^{t}_{u} = \sum_{j = 1}^K [(f_T(t - \bar{t}_j))^\top ~||~ (\widetilde{\pe}^{t}_{\bar{u}_j})^\top]^\top \in \mathbb{R}^{d_T + d_P}$, $f_T$ is the time encoder, and $\mlp^{(1)}_P \in \mathbb{R}^{d_P \times (d_P + d_T)}, \mlp^{(2)}_P \in \mathbb{R}^{d_P \times d_P},\mlp^{(self)}_P \in \mathbb{R}^{d_P \times d_P}$ are MLP parameters.

At the first time $t_0$, we denote the initial positional encoding as the Laplacian of the initial graph snapshot $G^{t_0}$.
Formally, let $\laplacian$ be the Laplacian of $G^{t_0}$ and the eigen-decomposition of $\laplacian$ be $\mathbf{U} \Lambda \mathbf{U}^T$, then $\pe^{t_0}_u = [\eigenvec_{1, u}, \dots, \eigenvec_{d_P, u}]^T \in \mathbb{R}^{d_P}$, where $\eigenvec_{i}$ is the $i^{\textrm{th}}$ eigenvector of $\laplacian$, as defined in Section~\ref{sec:prelim}.

\subsection{Node-Link-Positional Encoder}
\label{sec: encoder}


For node $u$, the Node-Link-Positional Encoder is designed to aggregate its historical node and edge features with the positional encoding of its recent 1-hop neighbors prior to time $t$, and, eventually, derive a compact representation associated with $u$ that encodes all the information.


In general, Node-Link-Positional Encoder works as follows.
We first use $\h^{t}_{u, N}$ and $\h^{t}_{u, E}$ to denote the node and link encoding of recent $1$-hop neighbors of node $u$.
Then, we apply MLP transformations on $\h^{t}_{u, N}$ and $\h^{t}_{u, E}$ to obtain the encoding, denoted by $\h^t_{u, N\|E}$, which represents the combination of node-link information.
For adding positional information, we use $\h^{t}_{u, P}$ to denote the aggregated positional encoding of recent $1$-hop neighbors, and further transform 
them by MLP layers. Eventually, $\h^{t}_{u, N\|E}$ and $\h^t_{u, P}$ are combined by a $1$-layer MLP to derive a temporal representation at time $t$ for $u$, denoted by $\h^{t}_u$, which will be sent together with $\h^{t}_v$ to the link predictor in Section~\ref{sec:optimization} to determine whether the link exists between nodes $u$ and $v$ at time $t$.

The detailed computational procedures of the Node-Link-Positional Encoder are explained below.


First, the node encoding of recent 1-hop neighbors of $u$ can be obtained by aggregating features of nodes that interact with $u$ in the time interval $[t - t_{gap}; t)$, where $t_{gap}$ is a dataset-dependent hyper-parameter. Mathematically, this process can be described as

\begin{equation}
    \label{eq:node_encoding}
    \begin{split}
    \colorbox{red!20}{$\h_{u, N}^t$} = \nodefeat{u}^t + &\text{Mean}(\{\nodefeat{u'}^{t'} \big| (u', u, t') \in \neighborhood_{u}^{t - t_{gap}} \cup \dots \cup \neighborhood_{u}^{t}\}
    \end{split}
\end{equation}

where  $\nodefeat{u}^t$ is the input node features, and $\text{Mean}(\cdot)$ is the mean pooling operation.


Then, we derive the link encoding for $u$ by extracting the $K$ most recent interactions involving $u$ that occur prior to time $t$, combine the edge features and the time encoding of these interactions, and further apply MLP transformations on these representations. Specifically, suppose the $K$ most recent interactions of $u$ are indexed as $\{(u_1, u, t'_1), \dots, (u_K, u, t'_K)\}$, where $t'_1 \leq \dots \leq t'_K < t$. Let $\mathbf{H}^t_{u, E} \in \mathbb{R}^{K \times (d_T + d_E)}$ be a matrix, whose $k^{\textrm{th}}$ row is defined as $[\mathbf{H}^t_{u, E}]_{k} = [(f_T(t - t'_k))^\top ~||~ (\edgefeat{t}_{u, u'_k})^\top] \in \mathbb{R}^{1 \times (d_T + d_E)}$, where $[\cdot || \cdot]$ denotes concatenation, $f_T$ is the time encoder, and $\edgefeat{t}_{u, u'_k}$ denotes the edge features. If $u$ does not involve at least $K$ interactions before $t$ then $\mathbf{H}^{t}_{u, E}$ is zero-padded so that the matrix has $K$ rows. We first apply a $1$-layer MLP on the last dimension of $\mathbf{H}^{t}_{u, E}$, sum-pool over $K$ rows of $\mathbf{H}^{t}_{u, E}$ with learnable weights to obtain a vector of dimension $d_T + d_E$, and finalize the encoding with an additional $1$-layer MLP transformation as
\begin{equation}
\label{eq:link_encoding_1}
    \colorbox{green!20}{$\h^t_{u, E}$} = \mlp^{(2)}_{link}(\text{RELU}((\mathbf{H}^t_{u, E} \mlp^{(1)}_{link})^\top \mlp^{sum}_{link})) \in \mathbb{R}^{d_T + d_E}
\end{equation}
where $\mlp^{(1)}_{link}, \mlp^{(2)}_{link} \in \mathbb{R}^{(d_T + d_E) \times (d_T + d_E)}$, and $\mlp^{sum}_{link} \in \mathbb{R}^{K \times 1}$.





After that, $\h^t_{u,  N\|E}$ is obtained by a $1$-layer MLP transformation as
\begin{equation}
    \label{eq:node-link-encoding}
    \colorbox{orange!20}{$\h^t_{u, N\|E}$} = \mlp_{N\|E}([(\colorbox{red!20}{$\h^t_{u, N}$})^\top ~||~ (\colorbox{green!20}{$\h^t_{u, E}$})^T])^\top \in \mathbb{R}^{d_N}
\end{equation}
where $\mlp_{N\|E} \in \mathbb{R}^{d_N \times (d_N + d_T + d_E)}$.

Next, for historical positional information, we apply sum aggregation over the positional information from the aforementioned $K$ most recent interactions involving $u$ as
\begin{equation}
    \label{eq: agg_neighbor_pe}
    \colorbox{blue!20}{$\widehat{\h}^{t}_{u, P}$} = \sum_{j = 1}^K [(f_T(t - t'_j))^\top ~||~ (\widetilde{\pe}^{t}_{u_j})^\top]^\top \in \mathbb{R}^{d_T + d_P}
\end{equation}
where $f_T$ is the time encoder. Let $\mlp^{(1)}_P \in \mathbb{R}^{d_P \times (d_P + d_T)}$, $\mlp^{(2)}_P \in \mathbb{R}^{d_P \times d_P}$, $\mlp^{(self)}_P \in \mathbb{R}^{d_P \times d_P}$, and $tanh$ be the hyperbolic tangent function. Then, $\h^{t}_{u, P}$ is obtained through the neighborhood-aggregated positional encoding $\widehat{\h}^{t}_{u, P}$ and $\widetilde{\pe}^{t}_{u}$ as
\begin{equation}
    \label{eq: pe}
    \colorbox{yellow!30}{$\h^{t}_{u, P}$} = \widetilde{\pe}^{t}_{u} + tanh(\mlp^{(self)}_P \widetilde{\pe}^{t}_{u}  + \mlp^{(2)}_{P}\text{RELU}(\mlp^{(1)}_{P} \colorbox{blue!20}{$\widehat{\h}^{t}_{u, P}$})) 
\end{equation}


Finally, the temporal representation at time $t$ of $u$, $\h^{t}_u$, is derived as
\begin{equation}
\label{eq:temporal_rep}
    \h^{t}_u = \mlp [(\colorbox{orange!20}{$\h^{t}_{u, N\|E}$})^\top ~||~ (\colorbox{yellow!30}{$\h^{t}_{u, P}$})^\top]^\top \in \mathbb{R}^{d_N}
\end{equation}
where $\mlp \in \mathbb{R}^{d_N \times (d_N + d_P)}$.


\subsection{Optimization} \label{sec:optimization}
With the two embedding vectors $\h^{t}_u$ and $\h^{t}_v$, we can now let \name\ decide whether a link exists between nodes $u$ and $v$ at time $t$ and train \name\ against the ground truth connections before $t$.

\textbf{Link Predictor}.
Given a node pair $(u, v)$ and timestamp $t$, \name's goal is to predict whether there is a link between $(u, v)$ at time $t$. Specifically, the Link Predictor would give the probability $\hat{y} \in (0, 1)$ of whether $u$ and $v$ interact at time $t$ by
\begin{equation}
    \label{eq: link_prediction}
    \hat{y} = sigmoid(\mlp^{(2)} \text{RELU}(\mlp^{(1)} [(\h^{t}_u)^\top ~||~ (\h^{t}_v)^\top]^\top)) 
\end{equation}
where $sigmoid$ denotes the sigmoid function mapping all real values to the range $(0, 1)$, and $\mlp^{(1)} \in \mathbb{R}^{(2d_N) \times d_N}, \mlp^{(2)} \in \mathbb{R}^{d_N \times 1}$ are $1$-layer MLPs.

\textbf{Loss Functions}.
To begin with, we point out the necessary components of the objective loss function for the training process of \name, give formal definitions for those components, and derive the final loss function used in the training process of \name.

Suppose there are $B$ positive (i.e., existing edges) samples $(u^{(pos)}_1, v^{(pos)}_1, t_1), \dots, (u^{(pos)}_B, v^{(pos)}_B, t_B)$ and $B$ negative (i.e., non-existing edges) samples $(u^{(neg)}_1, v^{(neg)}_1, t_1), \dots, (u^{(neg)}_B, v^{(neg)}_B, t_B)$. We employed $3$ different negative sampling strategies (NSS): random, historical, and inductive; more details can be found in Appendix~\ref{app:nss}. Specifically, following \cite{DBLP:conf/nips/0004S0L23}, during the model training stage, we employ random NSS for the loss function computation, and then evaluate our model with all $3$ NSS. Thus the ground truth label for the positive samples should be $1$, and $0$ for the negative samples. As the link prediction can be regarded as a binary classification problem, we employ the binary cross-entropy loss function to obtain the link prediction loss $\losslp$ as
\begin{equation}
    \label{eq:loss_func_lp}
    \begin{split}
    &\losslp = \frac{-1}{2B} \bigg[ \bigg(\sum_{i = 1}^B \log \big(\hat{y}_{u^{(pos)}_i, v^{(pos)}_i} \big) \bigg) + \\
    &\bigg(\sum_{i = 1}^B \log \big(1- \hat{y}_{u^{(neg)}_i, v^{(neg)}_i} \big) \bigg)\bigg]
    \end{split}
\end{equation}
where $\hat{y}_{u^{(pos)}_i, v^{(pos)}_i}$ denotes the predicted probability of a link existing between $u^{(pos)}_i, v^{(pos)}_i$ at time $t_i$, as defined in Eq.~\ref{eq: link_prediction}, and $\hat{y}_{u^{(neg)}_i, v^{(neg)}_i}$ is the predicted probability of a link existing between $u^{(neg)}_i, v^{(neg)}_i$ at time $t_i$. Thus, ideally, $\hat{y}_{u^{(pos)}_i, v^{(pos)}_i}$ should be close to $1$, and $\hat{y}_{u^{(neg)}_i, v^{(neg)}_i}$ should be close to zero.

In addition, we introduce the positional encoding loss $\losspe$ to assess the learned positional encoding, defined in Eq.~\ref{eq:approx_pe}.
Specifically, we would expect $\widetilde{\pe}^{t_i}_{u^{(pos)}_i}$ be close to $\widetilde{\pe}^{t_i}_{v^{(pos)}_i}$, as there is a link between $u^{(pos)}_i, v^{(pos)}_i$ at time $t_i$.
Thus, the difference $|| \widetilde{\pe}^{t_i}_{u^{(pos)}_i} - \widetilde{\pe}^{t_i}_{v^{(pos)}_i} ||_2$ (where $||\cdot||_2$ denotes the $2$-norm of a real-valued vector) should be small, so that the learned positional encoding correctly reflects the nature of positional encoding.
Moreover, in order to avoid \name\ giving similar positional encoding to all nodes to minimize the aforementioned $2$-norm difference of positive samples, we add another constraint, i.e., maximizing the difference $\alpha_{neg}|| \widetilde{\pe}^{t_i}_{u^{(neg)}_i} - \widetilde{\pe}^{t_i}_{v^{(neg)}_i} ||_2$, where $\alpha_{neg}$ is a positive scaling factor. As $(u^{(neg)}_1, v^{(neg)}_1, t_1), \dots, (u^{(neg)}_B, v^{(neg)}_B, t_B)$ are not all possible negative links (i.e., not connected), we let $0< \alpha_{neg} < 1$ to avoid \name\ emphasizing the negativity of these $B$ negative samples. Therefore, we define the positional encoding loss as
\begin{equation}
\label{eq:loss_func_pe}
\begin{split}
    &\losspe = \frac{1}{B} \bigg[ \bigg(\sum_{i = 1}^B || \widetilde{\pe}^{t_i}_{u^{(pos)}_i} - \widetilde{\pe}^{t_i}_{v^{(pos)}_i} ||_2 \bigg) - \\
    &\alpha_{neg}\bigg(\sum_{i = 1}^B || \widetilde{\pe}^{t_i}_{u^{(neg)}_i} - \widetilde{\pe}^{t_i}_{v^{(neg)}_i} ||_2 \bigg)\bigg]\\
\end{split}
\end{equation}

Finally, the overall loss function is
\begin{equation}
    \label{eq:loss_func}
    \mathcal{L} = (1 - \alpha_{pe}) \losslp + \alpha_{pe} \losspe
\end{equation}
where $\alpha_{pe} \in (0, 1)$ is the weight for $\mathcal{L}_{pe}$.

The training procedures of \name\ can be found in Appendix~\ref{pseudo-code}.




\section{Theoretical Analysis}
\label{sec:theory}
In this section, we first theoretically show that, at time $t$, the approximated positional encoding $\widetilde{\pe}^{t}$, defined by Eq.~\ref{eq:approx_pe}, can be a good approximation of the positional encoding of graph snapshot $G^t$ by establishing a bound for the approximation error.
\begin{theorem}
\label{thm1}
    Suppose the temporal graph $G$ is ``slowly changing'' (i.e., the ground truth positional encoding $\pe^{0} \approx \dots \approx \pe^{t-1}$), then the corresponding approximated positional encoding of \name\ satisfies
\begin{equation}
\label{eq:theoretical_analysis_claim}
    ||\widetilde{\pe}^{t}_u - \widetilde{\pe}^{t''}_u||_2 \leq \epsilon
\end{equation}
where $t'' > t$ is the future distinct timestamp of $t$, and $\epsilon$ is a bounding term that depends on $L$, which is the number of recent graph snapshots utilized in obtaining the approximated positional encoding, as stated in Eqs.~\ref{eq:transform_pe} and~\ref{eq:approx_pe}. (Proof in Appendix~\ref{app:proofs})
\end{theorem}

Briefly, Theorem~\ref{thm1} indicates that when the graph is slowly changing, the positional encoding learned at the previous timestamp can well preserve the next close future timestamp’s positional encoding. In other words, as Theorem~\ref{thm1} suggests, the positional encoding function of \name\ can be a good representation of the positional information of future graph snapshots without leveraging the information about the future graph topology.

Note that the above statements do not necessarily mean that \name\ does not need to update the positional encoding in every case. Actually, \name\ indeed updates the positional encoding when new interaction happens. The role of Theorem~\ref{thm1} is to demonstrate the effectiveness of our positional encoding in the inductive setting, where the future scenario is usually not given or difficult to observe.



\textbf{Complexity Analysis}. Next, we provide the analysis on \name's computational complexity. \name\ consists of two main components: LPE Module and Node-Link-Positional Encoder. In LPE Module, the approximated positional encoding is derived by applying Discrete Fourier Transform on a $L$-length sequence of previous positional encoding, so the complexity is $\mathcal{O}(L \log(L))$, achieved by utilizing Fast Fourier Transform. Then, we update the positional encoding by sampling the $K$ most recent interactions, adding $\mathcal{O}(K)$ time. Therefore, for $n$ nodes, LPE Module costs $\mathcal{O}(n(L \log(L) + K))$. For Node-Link-Positional Encoder, for each node we retrieve node-level information from its neighborhood in the interval $[t - t_{gap}; t]$ and sample its $K$ most recent interactions, so the complexity is $\mathcal{O}(n\cdot(t_{gap}+ K))$, which can be achieved by pre-computing all temporal neighbors in the range $[t - t_{gap}; t]$ of each node. Thus, the total time complexity is $\mathcal{O}(n(t_{gap} + K + L\log(L))$, which scales linearly with the number of nodes, as $L, K, t_{gap}$ are constants independent of $n$. 

Compared to recent SOTA, DyGFormer \citep{DBLP:conf/nips/0004S0L23} and FreeDyG \citep{DBLP:conf/iclr/TianQG24}, \name~is more efficient as both of these methods employ neighborhood co-occurrence for each node pair of interaction, making their worst-case complexity scale with the number of edges, $|E|$; while \name's complexity scales with $n$, the number of nodes, and it is quite common that $n << |E|$
, showing \name's advantage in efficiency. 
Detailed theoretical time complexity comparison can be found in Appendix~\ref{sec:time_complexity}, and detailed empirical running time comparison can be found in Appendix~\ref{app:scalability}.

\begin{table*}[t]
\centering
\caption{Performance comparison in the \textit{transductive setting} with \textit{random negative sampling strategy}.}
\vspace{1mm}
\resizebox{1.01\textwidth}{!}
{
\setlength{\tabcolsep}{0.9mm}
{
\begin{tabular}{c|c|cccccccccc}
\hline
Metric                     & Datasets    & JODIE            & DyRep            & TGAT             & TGN                                              & CAWN             & TCL              & GraphMixer       & DyGFormer & FreeDyG                                        & \name\                                           \\ \hline
& Wikipedia   & 96.50 $\pm$ 0.14 & 94.86 $\pm$ 0.06 & 96.94 $\pm$ 0.06 & 98.45 $\pm$ 0.06 & 98.76 $\pm$ 0.03 & 96.47 $\pm$ 0.16 & 97.25 $\pm$ 0.03 &  99.03 $\pm$ 0.02 & {\color[HTML]{329A9D} \textbf{99.26 $\pm$ 0.01}} & {\color[HTML]{9A0000} \textbf{99.34 $\pm$ 0.04}} \\
                           
& Reddit      & 98.31 $\pm$ 0.14 & 98.22 $\pm$ 0.04 & 98.52 $\pm$ 0.02 & 98.63 $\pm$ 0.06                                 & 99.11 $\pm$ 0.01 & 97.53 $\pm$ 0.02 & 97.31 $\pm$ 0.01 & 99.22 $\pm$ 0.01 & {\color[HTML]{9A0000}\textbf{99.48 $\pm$ 0.01}} & {\color[HTML]{329A9D} \textbf{99.37 $\pm$ 0.04}} \\
                           
& MOOC        & 80.23 $\pm$ 2.44 & 81.97 $\pm$ 0.49 & 85.84 $\pm$ 0.15 & {\color[HTML]{329A9D} \textbf{89.15 $\pm$ 1.60}} & 80.15 $\pm$ 0.25 & 82.38 $\pm$ 0.24 & 82.78 $\pm$ 0.15 &  87.52 $\pm$ 0.49 & {\color[HTML]{9A0000}\textbf{89.61 $\pm$ 0.19}} & 86.94 $\pm$ 0.34                                 \\
                           
& LastFM      & 70.85 $\pm$ 2.13 & 71.92 $\pm$ 2.21 & 73.42 $\pm$ 0.21 & 77.07 $\pm$ 3.97                                 & 86.99 $\pm$ 0.06 & 67.27 $\pm$ 2.16 & 75.61 $\pm$ 0.24 & {\color[HTML]{329A9D} \textbf{93.00 $\pm$ 0.12}} & 92.15 $\pm$ 0.16 & {\color[HTML]{9A0000} \textbf{96.06 $\pm$ 0.30}} \\
                           
& Enron       & 84.77 $\pm$ 0.30 & 82.38 $\pm$ 3.36 & 71.12 $\pm$ 0.97 & 86.53 $\pm$ 1.11                                 & 89.56 $\pm$ 0.09 & 79.70 $\pm$ 0.71 & 82.25 $\pm$ 0.16 & 92.47 $\pm$ 0.12 & {\color[HTML]{329A9D} \textbf{92.51 $\pm$ 0.05}} & {\color[HTML]{9A0000} \textbf{93.96 $\pm$ 0.36}} \\
                           
& Social Evo. & 89.89 $\pm$ 0.55 & 88.87 $\pm$ 0.30 & 93.16 $\pm$ 0.17 & 93.57 $\pm$ 0.17 & 84.96 $\pm$ 0.09 & 93.13 $\pm$ 0.16 & 93.37 $\pm$ 0.07 & {\color[HTML]{329A9D} \textbf{94.73 $\pm$ 0.01}} & {\color[HTML]{9A0000} \textbf{94.91 $\pm$ 0.01}} & 92.22 $\pm$ 0.25                                 \\
                           
& UCI         & 89.43 $\pm$ 1.09 & 65.14 $\pm$ 2.30 & 79.63 $\pm$ 0.70 & 92.34 $\pm$ 1.04 & 95.18 $\pm$ 0.06 & 89.57 $\pm$ 1.63 & 93.25 $\pm$ 0.57 & 95.79 $\pm$ 0.17 & {\color[HTML]{329A9D}\textbf{96.28 $\pm$ 0.11}} & {\color[HTML]{9A0000} \textbf{96.67 $\pm$ 0.47}} \\
                           
& Flights     & 95.60 $\pm$ 1.73 & 95.29 $\pm$ 0.72 & 94.03 $\pm$ 0.18 & 97.95 $\pm$ 0.14                                 & 98.51 $\pm$ 0.01 & 91.23 $\pm$ 0.02 & 90.99 $\pm$ 0.05 & {\color[HTML]{329A9D} \textbf{98.91 $\pm$ 0.01}} & 98.12 $\pm$ 0.19 & {\color[HTML]{9A0000} \textbf{98.94 $\pm$ 0.10}} \\
                           
& Can. Parl.  & 69.26 $\pm$ 0.31 & 66.54 $\pm$ 2.76 & 70.73 $\pm$ 0.72 & 70.88 $\pm$ 2.34                                 & 69.82 $\pm$ 2.34 & 68.67 $\pm$ 2.67 & 77.04 $\pm$ 0.46 & {\color[HTML]{329A9D} \textbf{97.36 $\pm$ 0.45}} & 72.22 $\pm$ 2.47 & {\color[HTML]{9A0000} \textbf{98.24 $\pm$ 0.11}} \\
                           
& US Legis.   & 75.05 $\pm$ 1.52 & 75.34 $\pm$ 0.39 & 68.52 $\pm$ 3.16 & {\color[HTML]{329A9D} \textbf{75.99 $\pm$ 0.58}} & 70.58 $\pm$ 0.48 & 69.59 $\pm$ 0.48 & 70.74 $\pm$ 1.02 & 71.11 $\pm$ 0.59                                 & 69.94 $\pm$ 0.59 & {\color[HTML]{9A0000} \textbf{76.74 $\pm$ 0.60}} \\
                           
& UN Trade    & 64.94 $\pm$ 0.31 & 63.21 $\pm$ 0.93 & 61.47 $\pm$ 0.18 & 65.03 $\pm$ 1.37                                 & 65.39 $\pm$ 0.12 & 62.21 $\pm$ 0.03 & 62.61 $\pm$ 0.27 & {\color[HTML]{329A9D} \textbf{66.46 $\pm$ 1.29}} & - & {\color[HTML]{9A0000} \textbf{75.84 $\pm$ 2.08}} \\
                           
& UN Vote     & 63.91 $\pm$ 0.81 & 62.81 $\pm$ 0.80 & 52.21 $\pm$ 0.98 & {\color[HTML]{329A9D} \textbf{65.72 $\pm$ 2.17}} & 52.84 $\pm$ 0.10 & 51.90 $\pm$ 0.30 & 52.11 $\pm$ 0.16 & 55.55 $\pm$ 0.42                                  & 51.99 $\pm$ 1.13 & {\color[HTML]{9A0000} \textbf{73.40 $\pm$ 0.03}} \\
                           
& Contact     & 95.31 $\pm$ 1.33 & 95.98 $\pm$ 0.15 & 96.28 $\pm$ 0.09 & 96.89 $\pm$ 0.56                                 & 90.26 $\pm$ 0.28 & 92.44 $\pm$ 0.12 & 91.92 $\pm$ 0.03 & {\color[HTML]{9A0000} \textbf{98.29 $\pm$ 0.01}} & 97.99 $\pm$ 0.03 & {\color[HTML]{329A9D} \textbf{98.16 $\pm$ 0.09}} \\ \cline{2-12} 

\multirow{-14}{*}{AP}      & 


Avg. Rank & 6.77 & 7.38 & 7.23 & 4.08 & 5.85 & 8.46 & 6.77 & {\color[HTML]{329A9D}{\textbf{2.77}}} & 3.85 & {\color[HTML]{9A0000}{\textbf{1.85}}} \\

\hline

& Wikipedia   & 96.33 $\pm$ 0.07 & 94.37 $\pm$ 0.09 & 96.67 $\pm$ 0.07 & 98.37 $\pm$ 0.07                                 & 98.54 $\pm$ 0.04 & 95.84 $\pm$ 0.18 & 96.92 $\pm$ 0.03 & 98.91 $\pm$ 0.02 & {\color[HTML]{329A9D} \textbf{99.41 $\pm$ 0.01}} & {\color[HTML]{9A0000} \textbf{99.48 $\pm$ 0.03}} \\

& Reddit      & 98.31 $\pm$ 0.05 & 98.17 $\pm$ 0.05 & 98.47 $\pm$ 0.02 & 98.60 $\pm$ 0.06                                 & 99.01 $\pm$ 0.01 & 97.42 $\pm$ 0.02 & 97.17 $\pm$ 0.02 & 99.15 $\pm$ 0.01 & {\color[HTML]{9A0000} \textbf{99.50 $\pm$ 0.01}} &  {\color[HTML]{329A9D} \textbf{99.49 $\pm$ 0.03}} \\
                           
& MOOC        & 83.81 $\pm$ 2.09 & 85.03 $\pm$ 0.58 & 87.11 $\pm$ 0.19 & {\color[HTML]{9A0000} \textbf{91.21 $\pm$ 1.15}} & 80.38 $\pm$ 0.26 & 83.12 $\pm$ 0.18 & 84.01 $\pm$ 0.17 & 87.91 $\pm$ 0.58                               & {\color[HTML]{329A9D} \textbf{89.93 $\pm$ 0.35}} & 89.28 $\pm$ 0.28 \\
                           
& LastFM      & 70.49 $\pm$ 1.66 & 71.16 $\pm$ 1.89 & 71.59 $\pm$ 0.18 & 78.47 $\pm$ 2.94                                 & 85.92 $\pm$ 0.10 & 64.06 $\pm$ 1.16 & 73.53 $\pm$ 0.12 & 93.05 $\pm$ 0.10 & {\color[HTML]{329A9D} \textbf{93.42 $\pm$ 0.15}} & {\color[HTML]{9A0000} \textbf{97.52 $\pm$ 0.17}} \\
                           
& Enron       & 87.96 $\pm$ 0.52 & 84.89 $\pm$ 3.00 & 68.89 $\pm$ 1.10 & 88.32 $\pm$ 0.99                                 & 90.45 $\pm$ 0.14 & 75.74 $\pm$ 0.72 & 84.38 $\pm$ 0.21 & 93.33 $\pm$ 0.13 & {\color[HTML]{329A9D} \textbf{94.01 $\pm$ 0.11}} & {\color[HTML]{9A0000} \textbf{95.98 $\pm$ 0.23}} \\
                           
& Social Evo. & 92.05 $\pm$ 0.46 & 90.76 $\pm$ 0.21 & 94.76 $\pm$ 0.16 & {\color[HTML]{329A9D} \textbf{95.39 $\pm$ 0.17}} & 87.34 $\pm$ 0.08 & 94.84 $\pm$ 0.17 & 95.23 $\pm$ 0.07 & 96.30 $\pm$ 0.01 & {\color[HTML]{9A0000} \textbf{96.59 $\pm$ 0.04}} & 94.33 $\pm$ 0.21                                 \\
                           
& UCI         & 90.44 $\pm$ 0.49 & 68.77 $\pm$ 2.34 & 78.53 $\pm$ 0.74 & 92.03 $\pm$ 1.13                                 & 93.87 $\pm$ 0.08 & 87.82 $\pm$ 1.36 & 91.81 $\pm$ 0.67 & 94.49 $\pm$ 0.26 & {\color[HTML]{329A9D} \textbf{95.00 $\pm$ 0.21}} & {\color[HTML]{9A0000} \textbf{97.62 $\pm$ 0.23}} \\
                           
& Flights     & 96.21 $\pm$ 1.42 & 95.95 $\pm$ 0.62 & 94.13 $\pm$ 0.17 & 98.22 $\pm$ 0.13                                 & 98.45 $\pm$ 0.01 & 91.21 $\pm$ 0.02 & 91.13 $\pm$ 0.01 & {\color[HTML]{329A9D} \textbf{98.93 $\pm$ 0.01}} & 98.03 $\pm$ 0.17 & {\color[HTML]{9A0000} \textbf{99.38 $\pm$ 0.04}} \\
                           
& Can. Parl.  & 78.21 $\pm$ 0.23 & 73.35 $\pm$ 3.67 & 75.69 $\pm$ 0.78 & 76.99 $\pm$ 1.80                                 & 75.70 $\pm$ 3.27 & 72.46 $\pm$ 3.23 & 83.17 $\pm$ 0.53 & {\color[HTML]{329A9D} \textbf{97.76 $\pm$ 0.41}} & 81.09 $\pm$ 2.28 & {\color[HTML]{9A0000} \textbf{98.97 $\pm$ 0.06}} \\
                           
& US Legis.   & 82.85 $\pm$ 1.07 & 82.28 $\pm$ 0.32 & 75.84 $\pm$ 1.99 & {\color[HTML]{329A9D} \textbf{83.34 $\pm$ 0.43}} & 77.16 $\pm$ 0.39 & 76.27 $\pm$ 0.63 & 76.96 $\pm$ 0.79 & 77.90 $\pm$ 0.58                                 & 77.26 $\pm$ 0.50 & {\color[HTML]{9A0000} \textbf{83.89 $\pm$ 0.43}} \\
                           
& UN Trade    & 69.62 $\pm$ 0.44 & 67.44 $\pm$ 0.83 & 64.01 $\pm$ 0.12 & 69.10 $\pm$ 1.67                                 & 68.54 $\pm$ 0.18 & 64.72 $\pm$ 0.05 & 65.52 $\pm$ 0.51 & {\color[HTML]{329A9D} \textbf{70.20 $\pm$ 1.44}} & - & {\color[HTML]{9A0000} \textbf{83.17 $\pm$ 1.28}} \\
                           
& UN Vote     & 68.53 $\pm$ 0.95 & 67.18 $\pm$ 1.04 & 52.83 $\pm$ 1.12 & {\color[HTML]{329A9D} \textbf{69.71 $\pm$ 2.65}} & 53.09 $\pm$ 0.22 & 51.88 $\pm$ 0.36 & 52.46 $\pm$ 0.27 & 57.12 $\pm$ 0.62                                 & 52.79 $\pm$ 1.56 & {\color[HTML]{9A0000} \textbf{79.59 $\pm$ 0.02}} \\
                           
& Contact     & 96.66 $\pm$ 0.89 & 96.48 $\pm$ 0.14 & 96.95 $\pm$ 0.08 & 97.54 $\pm$ 0.35                                 & 89.99 $\pm$ 0.34 & 94.15 $\pm$ 0.09 & 93.94 $\pm$ 0.02 & {\color[HTML]{329A9D} \textbf{98.53 $\pm$ 0.01}} & 98.39 $\pm$ 0.02 & {\color[HTML]{9A0000} \textbf{98.72 $\pm$ 0.06}} \\ \cline{2-12} 
                           
\multirow{-14}{*}{ROC-AUC} & Avg. Rank   


& 6.08 & 7.31 & 7.46 & 3.92& 6.0& 8.69& 7.15& {\color[HTML]{329A9D}\textbf{3.0}} & 3.69& {\color[HTML]{9A0000}\textbf{1.69}} \\

\hline
\end{tabular}
}
}
\label{tab:transductive_random}
\end{table*}

\section{Experiments}
\label{sec:exp}
Here, we provide the main results showing the outperformance of \name\ in classic and large-scale benchmark datasets. Furthermore, we leave
\begin{itemize}[noitemsep,topsep=0pt,parsep=0pt,partopsep=0pt,leftmargin=*]
    \item the effectiveness comparison in more learning and sampling settings in Appendix~\ref{app:link_prediction_nss},
    \item MLPs and Transformers performance on the proposed LPE positional encoding method in Appendix~\ref{sec:exp_ablation_study},
    \item running time comparison with SOTAs in Appendix~\ref{app:scalability},
    \item LPE ablation studies wrt input node and edge features in Appendix~\ref{app:ablation_study},
    \item analysis of important hyperparameters in Appendix~\ref{app:param_analysis},
    \item robustness of \name\ with different initial positional encoding input in Appendix~\ref{app:different_pe},
    \item detailed reproducibility in Appendix~\ref{app:reproducibility}.
\end{itemize}




\subsection{Experimental Settings}

\textbf{Datasets and baselines.} We assess the ability of \name\ in performing link prediction with $13$ datasets covering various domains and  collected by \citep{DBLP:conf/nips/PoursafaeiHPR22}: \textit{Wikipedia, Reddit, MOOC, LastFM, Enron, Social Evo., UCI, Flights, Can. Parl., US Legis., UN Trade, UN Vote,} and \textit{Contact}. Details about the dataset statistics are shown in Appendix~\ref{app:datasets}. We compare \name\ with $8$ state-of-the-art baselines, including JODIE \citep{DBLP:conf/kdd/KumarZL19}, DyRep \citep{DBLP:conf/iclr/TrivediFBZ19}, TGAT \citep{DBLP:conf/iclr/XuRKKA20}, TGN \citep{DBLP:journals/corr/abs-2006-10637}, CAWN \citep{DBLP:conf/iclr/WangCLL021}, TCL \citep{DBLP:journals/corr/abs-2105-07944}, GraphMixer \citep{DBLP:conf/iclr/CongZKYWZTM23}, DyGFormer \citep{DBLP:conf/nips/0004S0L23}, and FreeDyG~\citep{DBLP:conf/iclr/TianQG24}. Across all $13$ datasets, training/validation/testing sets are following the standard library~\citep{DBLP:conf/nips/0004S0L23} by chronological splits with ratios $70\% / 15\% / 15\%$. The large-scale datasets with pre-defined splits are publicly available at TGB Benchmark~\citep{DBLP:journals/corr/abs-2406-09639}

\textbf{Evaluation metrics and settings.} Following existing works \citep{DBLP:conf/iclr/XuRKKA20, DBLP:journals/corr/abs-2006-10637, DBLP:conf/nips/0004S0L23} in evaluating models for link prediction, we assess \name\ under two settings: \textbf{transductive} setting and \textbf{inductive} setting. Under the transductive setting, models will predict the link occurrence in future timestamps between nodes that had been observed during the training process, while the inductive setting involves predicting future links between unseen nodes in the training process. For each setting, we use two evaluation metrics: Average Precision (AP) and Area Under the Receiver Operating Characteristic Curve (AUC-ROC). In addition, following \citep{DBLP:conf/nips/PoursafaeiHPR22}, for more robust evaluation, each baseline is assessed with three negative sampling strategies (NSS): random, historical, and inductive (Details in Appendix~\ref{app:reproducibility}). 


\begin{table*}[t]
\centering
\caption{\textit{Test MRR} and \textit{Validation MRR} in the worldwide leaderboard of TGB benchmark}
\vspace{1mm}
\resizebox{1\textwidth}{!}
{
\setlength{\tabcolsep}{0.9mm}
{
\begin{tabular}{cccc|cccc}
\hline
\multicolumn{4}{c|}{tgbl-review} & \multicolumn{4}{c}{tgbl-coin} \\
\hline
Rank &	Method &	Test MRR &	Validation MRR  & Rank &	Method &	Test MRR &	Validation MRR \\
\hline
1 &	GraphMixer &	0.521 $\pm$ 0.015 &	0.428 $\pm$ 0.019 & 1 &	TNCN &	0.762 $\pm$ 0.004 & 0.740 $\pm$ 0.002 \\
2 & \textbf{\name} (Ours) & \textbf{0.411 $\pm$ 0.011} & \textbf{0.330 $\pm$ 0.017} & 2 &	DyGFormer &	0.752 $\pm$ 0.004 &	0.730 $\pm$ 0.002	\\
3 &	TNCN &	0.377 $\pm$ 0.010 &	0.325 $\pm$ 0.003 & 3 &	\textbf{\name} (Ours) & \textbf{0.711 $\pm$ 0.0185}	 & \textbf{0.674 $\pm$ 0.021}		\\
4 &	TGAT &	0.355 $\pm$ 0.012 &	0.324 $\pm$ 0.006 & 4 &	TGN &	0.586 $\pm$ 0.037 &	0.607 $\pm$ 0.014	\\
5 &	TGN &	0.349 $\pm$ 0.020 &	0.313 $\pm$ 0.012 & 5 &	EdgeBank (tw) &	0.580 &	0.492	\\
6 &	NAT &	0.341 $\pm$ 0.020 &	0.302 $\pm$ 0.011 & 6 &	DyRep &	0.452 $\pm$ 0.046 &	0.512 $\pm$ 0.014 \\	
7 &	DyGFormer &	0.224 $\pm$ 0.015 &	0.219 $\pm$ 0.017 & 7 &	EdgeBank (unlimited) &	0.359 &	0.315	\\	
8 &	DyRep &	0.220 $\pm$ 0.030 &	0.216 $\pm$ 0.031	& & & & \\
9 &	CAWN&	0.193 $\pm$ 0.001 &	0.200 $\pm$ 0.001	& & & & \\
10 &	TCL	&0.193 $\pm$ 0.009 &	0.199 $\pm$ 0.007	& & & & \\
11 &	EdgeBank (tw) &	0.025 &	0.024	& & & & \\
12 &	EdgeBank (unlimited) &	0.023 &	0.023 & & & & \\

\hline
\end{tabular}
}
}
\label{tab:tgb}
\end{table*}

\subsection{Temporal Link Prediction Performance on 13 Classic Datasets}
\label{sec:exp_link_prediction}

Due to the limited space, we present all baseline methods with respect to AP and AUC-ROC metrics with \textit{random} negative sampling strategy here for the transductive setting in Table~\ref{tab:transductive_random} and the inductive setting in Table~\ref{tab:inductive_random} in Appendix ~\ref{app:link_prediction_nss}. The performance of \textit{historical} and \textit{inductive} negative sampling strategies on both learning settings are placed in Appendix~\ref{app:link_prediction_nss}. The first and second results are highlighted with ${\color[HTML]{9A0000} \textbf{Red}}$ and ${\color[HTML]{329A9D} \textbf{Blue}}$.
For the \textit{random} negative sampling strategy, in Tables~\ref{tab:transductive_random} and~\ref{tab:inductive_random}, \name\ achieves competitive results over $13$ datasets, outperforming most of the baselines with average rankings close to $1$ under both settings. Moreover, the second-best results of \name\ closely approach the corresponding best results by a small gap. Notably, on some datasets, such as $\textit{UN Trade, UN Vote}$, the difference between the second-best and \name's performance is substantial, suggesting that \name\ makes great improvements over existing methods.
Regarding the \textit{historical} and \textit{inductive} sampling strategies, \name\ also attains competitive results, as observed in Tables~\ref{tab:transductive_hist}, ~\ref{tab:inductive_hist}, ~\ref{tab:transductive_ind}, and ~\ref{tab:inductive_ind}. In most cases, \name\ surpasses the second-best by a substantial margin. 





\subsection{Temporal Link Prediction Performance on the Large-Scale Benchmark -- TGB}
\label{sec:scalability}

In this section, we assess the scalability of our method by evaluating \name~on large-scale temporal graphs from the TGB Benchmark~\citep{DBLP:journals/corr/abs-2406-09639} with link prediction task. Specifically, we report \name's performance on \textit{tgbl-review} and \textit{tgbl-coin} datasets in Table~\ref{tab:tgb}, following the given pre-defined splits and metrics. Moreover, \textit{tgbl-review} has \textbf{352,637 nodes} and \textbf{4,873,540 edges}, while \textit{tgbl-coin} has \textbf{638,486 nodes} and \textbf{22,809,486 edges}.
As shown in Table~\ref{tab:tgb}, our method achieves competitive performance, ranking the $2^{\textrm{nd}}$ place on \textit{tgbl-review}~\footnote{\url{https://tgb.complexdatalab.com/docs/leader_linkprop/\#tgbl-review-v2}} and the $3^{\textrm{rd}}$ place on \textit{tgbl-coin}~\footnote{\url{https://tgb.complexdatalab.com/docs/leader_linkprop/\#tgbl-coin-v2}}, according to the worldwide open leaderboard.
To further demonstrate the efficiency of \name, we provide runtime comparison between our method and some SOTA methods, DyGFormer \citep{DBLP:conf/nips/0004S0L23} and FreeDyG~\citep{DBLP:conf/iclr/TianQG24}, on \textit{US Legis} and \textit{UN Trade} datasets in Table~\ref{tab:runtime_comparison} in Appendix~\ref{app:scalability}. 

\section{Related Work}

Recently, many efforts have been made to develop the next-generation graph foundation models~\citep{zheng2024drgnn, DBLP:journals/corr/abs-2412-21151, DBLP:journals/corr/abs-2410-02296, DBLP:journals/corr/abs-2412-08174, DBLP:conf/iclr/WangHZFYCHWYL25, DBLP:journals/corr/abs-2504-01346}.
As an important pillar, current most temporal graph learning methods are designed to learn node temporal representations at a certain timestamp, and then obtain the link prediction between a pair of nodes $(u, v)$ at time $t$. They typically start by concatenating the temporal representations of $u, v$ at time $t$, and then process the concatenation with MLP layers to estimate the link occurrence probability. Usually, the temporal representations of nodes are obtained by aggregating intrinsic node/edge features from recent temporal neighbors and interactions, which are then processed using more complex architectures.

For example, TGN \citep{DBLP:journals/corr/abs-2006-10637} and JODIE \citep{DBLP:conf/kdd/KumarZL19} employ the recurrent architecture, TGAT \citep{DBLP:conf/iclr/XuRKKA20} utilizes the self-attention mechanism, DyRep \citep{DBLP:conf/iclr/TrivediFBZ19} uses Temporal Point Processes, and more recently, GraphMixer \citep{DBLP:conf/iclr/CongZKYWZTM23} leverages Mixer-MLP layers, while DyGFormer \citep{DBLP:conf/nips/0004S0L23} harnesses the power of transformers and FreeDyG \citep{DBLP:conf/iclr/TianQG24} utilizes Fourier Transform to encode neighborhood co-occurrence information.
From another aspect, some methods aggregate information from higher-order graph structures. For example, TGN \citep{DBLP:journals/corr/abs-2006-10637} and TGAT \citep{DBLP:conf/iclr/XuRKKA20} both aggregate information from $k$-hop neighborhoods, and CAWN \citep{DBLP:conf/iclr/WangCLL021} samples random walks and encode them to derive the node representations. PINT~\cite{DBLP:conf/nips/SouzaMKG22} leverages positional features constructed from temporal random walks to enhance the model's expressiveness. A detailed discussion and comparison between our method and PINT~\cite{DBLP:conf/nips/SouzaMKG22}'s positional features can be found in Appendix~\ref{app:related-work-pint}.

Our \name\ achieves effectiveness with efficiency. First, our \name\ only considers $1$-hop neighborhood for information aggregation and processes representations with simple MLP transformations. More importantly, \name\ integrates node and edge features along with evolving positional encodings, which is informative to support task out-performance.The integration is intuitive and does not really rely on heavy attention mechanisms. 

\section{Conclusion}
\label{others}
In this paper, we propose a simple temporal link prediction model, \name, which only relies on learnable positional encoding and MLPs to achieve out-performance over SOTA graph transformers in temporal link prediction tasks. In addition to theoretical analysis of the effectiveness and efficiency, we design comprehensive experiments to demonstrate the empirical performance of \name.

\section*{Acknowledgments}

This work is supported by National Science Foundation under Award No. IIS-2117902, and the U.S. Department of Homeland Security under Grant Award Number 17STQAC00001-08-00. The views and conclusions are those of the authors and should not be interpreted as representing the official policies of the funding agencies or the government.

\section*{Impact Statement}


This paper presents work whose goal is to advance the field of 
Machine Learning. There are many potential societal consequences 
of our work, none which we feel must be specifically highlighted here.


\nocite{langley00}

\bibliography{reference}
\bibliographystyle{icml2025}

\newpage
\appendix
\onecolumn

\section{Appendix Contents}
The appendix contains the following information:
\begin{itemize}
    \item Appendix~\ref{pseudo-code}: Pseudo-code of \name\
    \item Appendix~\ref{app:proofs}: Proof of LPE
    \item Appendix~\ref{app:pe-signal-filter}: Elaborations on the Learnable Filter
    \item Appendix~\ref{sec:time_complexity}: Theoretical Time Complexity of \name\
    \item Appendix~\ref{app:related-work-pint}: Detailed Theoretical and Empirical Comparison with Provably Expressive Temporal Graph Networks
    \item Appendix~\ref{app:datasets}: Dataset Details
    \item Appendix~\ref{sec: app_experiments}: More Experiments
    \begin{itemize}
        \item Appendix~\ref{app:link_prediction_nss}: The effectiveness comparison in more learning and sampling settings
        \item Appendix~\ref{sec:exp_ablation_study}: MLPs and Transformers performance on the proposed LPE positional encoding method
        \item Appendix~\ref{app:scalability}: Running time comparison with SOTAs
        \item Appendix~\ref{app:ablation_study}: LPE ablation studies wrt input node and edge features
        \item Appendix~\ref{app:param_analysis}: Analysis of important hyperparameters
        \item Appendix~\ref{app:different_pe}: Robustness of \name\ with different initial positional encoding input
    \end{itemize}
    \item Appendix~\ref{app:reproducibility}: Reproducibility of \name
\end{itemize}

\section{Pseudo-code for \name}
\label{pseudo-code}

In this section, we provide the pseudo-code for \name. Specifically, Algorithm.~\ref{algo:temporal_representation} summarizes the computational process of the Node-Link-Positional Encoder, Algorithm.~\ref{algo:loss} demonstrates the computation of the objective loss function, and finally, Algorithm.~\ref{algo:train} briefly describe the training procedure of \name\ as a neural architecture.


\begin{algorithm}
\caption{Temporal Representation}
\label{algo:temporal_representation}
\begin{algorithmic}
\REQUIRE $\text{node } u, \text{time } t$
\ENSURE $\text{Temporal Representation for } u \text{ at } t$
\STATE $\{\widehat{\pe}^{t'_j}_{u}\}_{j = 1}^L \gets \text{Eq.~\ref{eq:transform_pe}}$

\STATE $\approxpe^{t}_u \gets  \text{Eq.~\ref{eq:approx_pe}}$ 

\STATE $\h^{t}_{u, N} \gets \text{Eq.~\ref{eq:node_encoding}}$

\STATE $\mathbf{H}^{t}_{u, E} \gets \text{Eq.~\ref{eq:link_encoding_1}}$ 

\STATE $\h^{t}_{u, E} \gets  \text{Eq.~\ref{eq:link_encoding_1}}$ 

\STATE $\h^{t}_{u, N \| E} \gets \text{Eq.~\ref{eq:node-link-encoding}}$

\STATE $\widehat{\h}^{t}_{u, P} \gets \text{Eq.~\ref{eq: agg_neighbor_pe}}$

\STATE $\h^{t}_{u, P} \gets \text{Eq.~\ref{eq: pe}}$ 

\STATE $\h^{t}_{u} \gets \text{Eq.~\ref{eq:temporal_rep}}$ \\

\STATE $\textbf{return } \h^{t}_{u}$

\end{algorithmic}
\end{algorithm}

\begin{algorithm}
\caption{Loss Function}
\label{algo:loss}
\begin{algorithmic}
\REQUIRE $B \text{ positive samples, } B \text{ negative samples}$
\ENSURE $\mathcal{L}$

\STATE $\mathcal{L}_{lp} \leftarrow 0$
\STATE $\mathcal{L}_{pe} \gets 0$

\FOR {$i \in [1, \dots, B]$}

\STATE $\h^{t}_{u^{(pos)}_i} \gets \text{Algorithm.~\ref{algo:temporal_representation}}(u^{(pos)}_i, t)$

\STATE $\h^{t}_{v^{(pos)}_i} \gets \text{Algorithm.~\ref{algo:temporal_representation}}(v^{(pos)}_i, t)$

\STATE $\h^{t}_{u^{(neg)}_i} \gets \text{Algorithm.~\ref{algo:temporal_representation}}(u^{(neg)}_i, t)$

\STATE $\h^{t}_{v^{(neg)}_i} \gets \text{Algorithm.~\ref{algo:temporal_representation}}(v^{(neg)}_i, t)$

\STATE $\hat{y}^{(pos)}_{i} \gets \text{Eq.~\ref{eq: link_prediction} with }\h^{t}_{u^{(pos)}_i}, \h^{t}_{v^{(pos)}_i}$ 

\STATE $\hat{y}^{(neg)}_{i} \gets \text{Eq.~\ref{eq: link_prediction} with }\h^{t}_{u^{(neg)}_i}, \h^{t}_{v^{(neg)}_i}$

\STATE $\mathcal{L}_{lp} \gets \mathcal{L}_{lp} + \log(\hat{y}^{(pos)}_{i}) + \log(1 - \hat{y}^{(neg)}_{i})$

\STATE $\mathcal{L}_{pe} \gets \mathcal{L}_{pe} + ||\approxpe^{t_i}_{u^{(pos)}_i} - \approxpe^{t_i}_{v^{(pos)}_i}||_2 - \alpha_{neg}||\approxpe^{t_i}_{u^{(neg)}_i} - \approxpe^{t_i}_{v^{(neg)}_i}||_2$
\ENDFOR

\STATE $\mathcal{L}_{lp} \gets -\mathcal{L}_{lp} / (2B)$
\STATE $\mathcal{L}_{pe} \gets \mathcal{L}_{pe} / B$
\STATE $\mathcal{L} \gets \text{Eq.~\ref{eq:loss_func}}$

\end{algorithmic}
\end{algorithm}

\begin{algorithm}
\caption{Training Pipeline of \name}
\label{algo:train}
\begin{algorithmic}
\REQUIRE $\text{Data-loader over } G \text{, number of epochs (\# Epochs)}$
\ENSURE $\text{Link Predictions}$
\STATE $\text{Initialize the PE of the initial snapshot with its Laplacian PE: } \pe^{t_0} \gets \text{ Laplacian PE of }G^{t_0}$
\FOR{epoch in [1, \dots, \# Epochs]}
\FOR{($B$ positive samples, $B$ negative samples) in Data-loader}
\STATE $\mathcal{L} \gets$ Algorithm.~\ref{algo:loss}($B$ positive samples, $B$ negative samples)
\STATE Back-propagate $\mathcal{L}$
\FOR{$t$ in [$t_1, \dots, t_B$]}
\IF{Finish link prediction at time $t$ ($t \in [t_1, t_B]$)}
\STATE reveal new links and update positional encoding: $\forall u, \pe^{t}_u \gets $ Eq.~\ref{eq:pe_updated}
\ENDIF
\ENDFOR
\ENDFOR
\ENDFOR
\end{algorithmic}
\end{algorithm}

\newpage

\section{Proof of Theorem~\ref{thm1}}
\label{app:proofs}

We start our proof for Eq. \ref{eq:theoretical_analysis_claim} by first stating the following Lemma about the eigenvalues of $n$-node ring graph, denoted by $R_n$.

\begin{lemma} 
\label{lemma:eigenvalues}
The real-valued part of the Fourier basis is the eigenvector of the Laplacian of the ring graph. Specifically, for a $n$-node ring graph $R_n$, the $k-$th eigenvector of its Laplacian, $\laplacian(R_n)$, is 

\begin{equation}
\label{eq:ring_graph_eigenvec}
    \mathbf{u}^{ring}_k = 
    \begin{bmatrix}
        \cos(2 \pi \cdot 0 / n) \\
        \cos(2 \pi \cdot 1 / n) \\
        \dots \\
        \cos(2 \pi \cdot (n - 1)) / n
    \end{bmatrix}
    \in \mathbb{R}^n
\end{equation}

and its corresponding eigenvalue is $\lambda^{ring}_k = 2 - \cos(2\pi k / n)$.

\end{lemma}

Lemma.~\ref{lemma:eigenvalues} is motivated by Lemma 5.2.1 from the lecture~\footnote{\url{https://www.cs.yale.edu/homes/spielman/561/lect05-15.pdf}}. Based on that, we can then extend to prove our Theorem.~\ref{thm1} as follows.

\begin{proof}

For short of notation, in the rest of this proof, suppose the positional encoding of snapshots at $L + 1$ distinct timestamps before $t$ are denoted as $\pe^{0}, \pe^{1}, \dots, \pe^{L}$, which are the Laplacian eigenvector of graph snapshots at $t_0, \dots, t_L$, and $\approxpe^{t}_u$ and $\approxpe^{t'}_u$ are denoted as $\approxpe^{L + 1}$ and $\approxpe^{L + 2}$, respectively. Again, for short of notation, we omit the subscript $u$ in the positional encoding notations. In addition, we denote $\laplacian(R_n)$ as the (un-normalized) Laplacian of a $n$-node ring graph, $R_n$.

We start by giving the definition of ``slow-changing'' network. We suppose that our network is ``slow-changing'', in terms of graph topology, thus, we can assume the eigenvector of the Laplacian does not vary across timestamps, i.e., $\pe^{0} \approx \dots \approx \pe^{L}$, so now we focus on bounding the left-hand side of Eq.~\ref{eq:theoretical_analysis_claim}.

We start by considering the following equation:
\begin{equation}
\begin{split}
    &\approxpe^{L + 1} - \approxpe^{L + 2} + \approxpe^{L + 1} - \approxpe^{L} + \big(\begin{bmatrix}
        \pe^0 - \pe^L & 0 & 0 & \dots & \pe^{L + 1} - \pe^1
    \end{bmatrix} \big)\mathbf{F} \mlp_{filter}\mathbf{F}^{-1} \begin{bmatrix}
        \alpha_1 \\
        \dots \\
        \alpha_{t - 1}
     \end{bmatrix}
    = \\
    &= \big(\begin{bmatrix}
        \pe^1 & \pe^2 & \dots & \pe^L
    \end{bmatrix} - \begin{bmatrix}
        \pe^2 & \pe^3 & \dots & \pe^{L + 1}
    \end{bmatrix} \\
    &+ \begin{bmatrix}
        \pe^1 & \pe^2 & \dots & \pe^L
    \end{bmatrix} - \begin{bmatrix}
        \pe^0 & \pe^1 & \dots & \pe^{L - 1}
    \end{bmatrix}  \\
    & + \begin{bmatrix}
        \pe^0 - \pe^L & 0 & 0 & \dots & \pe^{L + 1} - \pe^1
    \end{bmatrix} \big)\mathbf{F} \mlp_{filter}\mathbf{F}^{-1} \begin{bmatrix}
        \alpha_1 \\
        \dots \\
        \alpha_{t - 1}
     \end{bmatrix} \\
    &= \big(\begin{bmatrix}
        \pe^1 - \pe^2 + \pe^1 - \pe^0 & \dots & \pe^i - \pe^{i + 1} + \pe^i - \pe^{i - 1} & \dots & \pe^L - \pe^{L + 1} + \pe^L - \pe^{L - 1}
    \end{bmatrix}\\
    & + \begin{bmatrix}
        \pe^0 - \pe^L & 0 & 0 & \dots & \pe^{L + 1} - \pe^1
    \end{bmatrix}\big) \mathbf{F} \mlp_{filter}\mathbf{F}^{-1} \begin{bmatrix}
        \alpha_1 \\
        \dots \\
        \alpha_{t - 1}
     \end{bmatrix} \\
    &=\big(\begin{bmatrix}
        \pe^1 - \pe^2 + \pe^1 - \pe^L & \dots & \pe^i - \pe^{i + 1} + \pe^i - \pe^{i - 1} & \dots & \pe^L - \pe^1 + \pe^L - \pe^{L - 1}
    \end{bmatrix}\big)\mathbf{F} \mlp_{filter}\mathbf{F}^{-1} \begin{bmatrix}
        \pe^1 \\
        \dots \\
        \pe^L
     \end{bmatrix}\\
    &= \begin{bmatrix}
        \pe^{1} & \dots & \pe^{L}
    \end{bmatrix}L(R_{L}) \mathbf{F} \mlp_{filter}\mathbf{F}^{-1} \begin{bmatrix}
        \alpha_1 \\
        \dots \\
        \alpha_{L}
     \end{bmatrix} \\
\end{split}
\end{equation}

The eigen-decomposition of $L(R_{L})$ is $L(R_{L}) = F \Lambda F^{-1}$, where $\Lambda$ is a diagonal matrix with the $i-$th entri is $\lambda^{ring}_i$. Thus

\begin{equation}
\begin{split}
    &\begin{bmatrix}
        \pe^{1} & \dots & \pe^{L}
    \end{bmatrix}L(R_{L}) \mathbf{F} \mlp_{filter}\mathbf{F}^{-1} \begin{bmatrix}
        \alpha_1 \\
        \dots \\
        \alpha_{L}
     \end{bmatrix} \\
     &= \begin{bmatrix}
        \pe^{1} & \dots & \pe^{L}
    \end{bmatrix} \mathbf{F} \Lambda \mathbf{F}^{-1} \mathbf{F} \mlp_{filter}\mathbf{F}^{-1} \begin{bmatrix}
        \alpha_1 \\
        \dots \\
        \alpha_{L}
    \end{bmatrix} \\
    &= \begin{bmatrix}
       \pe^{1} & \dots & \pe^{L} 
    \end{bmatrix} \mathbf{F} (\Lambda \mlp_{filter}) \mathbf{F}^{-1} \begin{bmatrix}
        \alpha_1 \\
        \dots \\
        \alpha_{L}
    \end{bmatrix}
\end{split}
\end{equation}

For simplicity, let $\mlp_{filter}$ be a diagonal matrix, thus $(\Lambda \mlp_{filter})$ is a diagonal matrix.

Therefore, we obtain

\begin{equation}
\begin{split}
    &\begin{bmatrix}
       \pe^{1} & \dots & \pe^{L} 
    \end{bmatrix} \mathbf{F} (\Lambda \mlp_{filter}) \mathbf{F}^{-1} \begin{bmatrix}
        \alpha_1 \\
        \dots \\
        \alpha_{L}
    \end{bmatrix} \\
    &= \begin{bmatrix}
       \pe^{1} & \dots & \pe^{L} 
    \end{bmatrix} \sum_{i = 1}^{L} (\Lambda W_{filter})_i \mathbf{F}_{:, i} (\mathbf{F}^{-1}_{:, i})^T \begin{bmatrix}
        \alpha_1 \\
        \dots \\
        \alpha_{L}
    \end{bmatrix} \\
\end{split}
\end{equation}

Thus,

\begin{equation}
    \begin{split}
        &\approxpe^{L + 1} - \approxpe^{L + 2} + \approxpe^{L + 1} - \approxpe^{L} + \big(\begin{bmatrix}
        \pe^0 - \pe^L & 0 & 0 & \dots & \pe^{L + 1} - \pe^1
    \end{bmatrix} \big)\mathbf{F} \mlp_{filter}\mathbf{F}^{-1} \begin{bmatrix}
        \alpha_1 \\
        \dots \\
        \alpha_{t - 1}
     \end{bmatrix}
    = \\
    &= \begin{bmatrix}
       \pe^{1} & \dots & \pe^{L} 
    \end{bmatrix} \sum_{i = 1}^{L} (\Lambda W_{filter})_i \mathbf{F}_{:, i} (\mathbf{F}^{-1}_{:, i})^T \begin{bmatrix}
        \alpha_1 \\
        \dots \\
        \alpha_{L}
    \end{bmatrix} \\
    &= (\sum_{i = 1}^{L} (\Lambda \mlp_{filter})_i) (\sum_{j = 1}^{L} \alpha_j \pe^{j})\\
    &\Rightarrow \approxpe^{L + 1} - \approxpe^{L + 2} = \approxpe^{L} - \approxpe^{L + 1} - \big(\begin{bmatrix}
        \pe^0 - \pe^L & 0 & 0 & \dots & \pe^{L + 1} - \pe^1
    \end{bmatrix} \big)\mathbf{F} \mlp_{filter}\mathbf{F}^{-1} \begin{bmatrix}
        \alpha_1 \\
        \dots \\
        \alpha_{t - 1}
     \end{bmatrix}\\
     &+ (\sum_{i = 1}^{L} (\Lambda \mlp_{filter})_i) (\sum_{j = 1}^{L} \alpha_j \pe^{j})
    \end{split}
\end{equation}

Therefore,

\begin{equation}
\label{eq:last_equation}
\begin{split}
    &||\approxpe^{L + 1} - \approxpe^{L + 2}||_2 \leq ||(\sum_{i = 1}^{L} (\Lambda \mlp_{filter})_i) (\sum_{j = 1}^{L} \alpha_j \pe^{j})||_2
    + ||\approxpe^{L+1} - \approxpe^{L}||_2 \\
    &\leq 
    ||(\sum_{i = 1}^{L} (\Lambda \mlp_{filter})_i) (\sum_{j = 1}^{L} \alpha_j \pe^{j})||_2 + ||(\sum_{i = 1}^{L} (\Lambda \mlp_{filter})_i) (\sum_{j = 0}^{L - 1} \alpha_j \pe^{j})||_2 + 
    ||\approxpe^{L} - \approxpe^{L - 1}||_2 \\
    &\leq \dots \\
    &\leq (\sum_{i = 1}^{L} (\Lambda \mlp_{filter})_i) ||(\sum_{j = 1}^{L} \alpha_j \pe^{j})||_2 + (\sum_{i = 1}^{L} (\Lambda \mlp_{filter})_i) || \sum_{j = 0}^{L - 1}(L - 1 - j)\alpha_{j + 1} \pe^{j}||_2
\end{split}
\end{equation}

where the first inequality holds due to the assumption of ``slow-changing'' network.

For simplicity, we assume $\alpha_1, \alpha_2, \dots, \alpha_{t - 1}$ and the length of positional encoding $P^{(t)}$ is normalized, i.e $\alpha_1 + \dots + \alpha_{t - 1} = 1$

Thus, 

\begin{equation}
     (\ref{eq:last_equation})\leq \big(\sum_{i = 1}^{L} |(\Lambda\mlp_{filter})_i| \big) \cdot (2L - 2) 
\end{equation}

Since $L \ll N$, thus choosing $t$ small enough and choosing the filter such that $\sum_{i = 1}^L (\Lambda \mlp_{filter})_i < 1$ shows that the past positional encoding can be a well approximation of the positional encoding of the future snapshot.

\end{proof}

\section{Positional Encodings, Signals, and Noises Filtering} \label{app:pe-signal-filter}

To start with, we pave the way for understanding our learnable filter by unraveling how positional encodings could be regarded as signals on graphs.

Let the Laplacian eigein-decomposition of a graph snapshot $G^t$ be $\laplacian^t = \eigenvecs \mathbf{\Lambda} \eigenvecs^\top$. Suppose $G^t$ has $n$ nodes, then $\eigenvecs \in \mathbb{R}^{n \times n}$, where $\eigenvecs$'s columns are $G^t$'s eigenvectors, and $\mathbf{\Lambda} = \text{diag}(\lambda_1, \lambda_2, \dots, \lambda_n)$, where $\lambda_1 \leq \lambda_2 \dots \lambda_n$, is a diagonal matrix consisting of $\laplacian^t$'s eigenvalues. In addition, let $\eigenvec_i \in \mathbb{R}^n$ be the $i^{\textrm{th}}$ eigenvector of $\mathbf{L}$ and its $j^{\textrm{th}}$ entry denoted as $\eigenvec_{i,j}$, then, the positional encoding of node $i$ is defined as $\pe_i = [\eigenvec_{1, i}, \dots, \eigenvec_{n, i}]^\top \in \mathbb{R}^n$. In our case, the graph signal constructed by $\pe_i$ is a multi-dimensional signal. It is noteworthy that our paper does not consider graph signal constructed by the eigenvectors, and thus our multi-dimensional signals formed by positional encodings, $\pe_i$, are not associated with signals yielded by eigenvalues $\lambda_i$.

Next, we discuss the essence of using DFT and learnable filters to filer noises by elaborating what kinds of noises would be filtered by our design. (1) First, it is the change is caused by randomness. To be more specific, DFT is applied on the sequence of $u$’s positional encodings at previous timestamps and the learnable filter is designed to filter noises across dimensions in the spectral domain. If the signal changes from high-frequency to low-frequency (or vice versa) over time by randomness, then the learnable filter filters out noise in the spectral domain of the sequence. (2) Second, it is the changes caused by entity activities. It is also notable that the change pattern may reflect the historical properties of a batch of certain nodes, which is part of evidence for the current time decision making. Because our SimpleTLP updates positional encoding at each timestamp as shown in Eq.~\ref{eq:pe_updated}, this pattern will be recorded into the representation learning process constrained by the loss functions as shown in Eq.~\ref{eq:loss_func_lp} and Eq.~\ref{eq:loss_func_pe}.

\section{Theoretical time complexity comparison with SOTAs}
\label{sec:time_complexity}
Now, we give an in-depth analysis of the complexity of DyGFormer and FreeDyG as follows. 

Suppose we are given $E$ query links: $\{(u_i, v_i, t_i)\}_{i = 1}^E$. As we move to new timestamps, new connections and new nodes will emerge, resulting in the changes in some nodes’ neighborhoods; then, both \name\ and DyGFormer need to update historical neighbors for some nodes. (Note that for a node, both \name\ and DyGFormer consider its historical neighbors, so both methods would need to perform an update if its neighborhood changes).

For each link $(u_i, v_i, t_i)$, DyGFormer and FreeDyG first need to obtain the neighbor co-occurrence information of the node pair $(u_i, v_i)$ as follows. Suppose $u_i$ has $p$ historical neighbors, $a_1, \dots, a_p$, and $v_i$ has $q$ historical neighbors, $b_1, \dots, b_q$, then the neighbor co-occurrence scheme of DyGFormer and FreeDyG requires counting (1) the number of times $a_1, \dots, a_p$ are involved in interactions with $v_i$ before time $t_i$ and (2) the number of times $b_1, \dots, b_q$ are involved in interactions with $u_i$ before time $t_i$. In other words, for an arbitrary link $(u_i, v_i, t_i)$, the neighbor co-occurrence framework of DyGFormer and FreeDyG makes the computation of $u_i$ depend on $v_i$’s information, as they have to count how many times historical neighbors of $u_i$ are also $v_i$’s historical neighbors, and vice versa. Thus, in order to make predictions for the aforementioned  query links, DyGFormer and FreeDyG require at least $\mathcal{O}(E)$ complexity.


On the other hand, for \name, we first consider the unique nodes among $u_1, \dots, u_E, v_1, \dots, v_E$. Suppose there are $n$ unique nodes, and let us denote them as $c_1, \dots, c_n$, then we could compute their temporal representation as stated in Eq.~\ref{eq:temporal_rep}. Moreover, in the process of obtaining the necessary components for Eq.~\ref{eq:temporal_rep}, for any node $c_i$, we only use $c_i$’s related information (most recent temporal neighbors / links and its previous positional encoding). Thus, it takes \name~$\mathcal{O}(n)$ time to compute the representation for these $n$ nodes. To predict an arbitrary link $(u_i, v_i, t_i)$, we simply retrieve the representation for $u_i, v_i$ that we just computed and apply Eq.~\ref{eq: link_prediction}


In summary, DyGFormer and FreeDyG have $\mathcal{O}(E)$ since they process a node pair $(u, v)$-related information, while \name's complexity linearly scales with the number of nodes.

\section{Detailed theoretical and empirical comparison with related works}
\label{app:related-work-pint}
In this section, we provide a comparative discussion and experimental comparison between our method~\name~and PINT \citep{DBLP:conf/nips/SouzaMKG22}, a method for temporal link prediction that also leverages positional features.

To be more specific, PINT~\citep{DBLP:conf/nips/SouzaMKG22} proposes relative positional features that count the number of temporal walks of a given length between $2$ nodes. As stated in PINT~\citep{DBLP:conf/nips/SouzaMKG22}, temporal walks are defined as follows. An $(L - 1)$-length temporal walk is $W = \{(w_1, t_1), (w_2, t_2), \dots, (w_L, t_L)\}$, where $t_1 > t_2 > \dots t_L$ and $(w_{i - 1}, w_i, t_i)$ is an interaction in the given temporal graph. In summary, PINT proposes a relative positional encoding that represents pair-wise node “distance” (here, the “distance” is the number of $L$-length temporal walks, assuming $L$ is given). On the other hand, our \name\ learnable positional encodings (LPE) fall into the category of global positional encodings, which encode the position of the node relative to the global structure of the graph.

\begin{table}[h]
\centering
\caption{Performance comparison between PINT and \name\ in \textit{transductive setting} and \textit{inductive setting} with random negative sampling strategy}
\vspace{1mm}
\resizebox{0.99\textwidth}{!}
{
\setlength{\tabcolsep}{0.9mm}
{
\begin{tabular}{c|c|c|c|c|c}
\hline
Dataset & Method & Transductive AP &	Transductive ROC-AUC &	Inductive AP & 	Inductive ROC-AUC \\
\hline
\multirow{2}{*}{Enron}   & PINT &	81.60 ± 0.67&	84.50 ± 0.60&	67.01 ± 2.76&	70.41 ± 2.34 \\

& \name\ (Ours) &	\textbf{93.96 ± 0.36}&	\textbf{95.98 ± 0.23}&	\textbf{89.17 ± 0.88}	& \textbf{92.30 ± 0.57}
 \\
\hline
\multirow{2}{*}{UCI}   & PINT &	95.77 ± 0.11&	94.89 ± 0.15&	94.14 ± 0.05&	92.51 ± 0.07 \\
& \name\ (Ours) &	\textbf{96.67 ± 0.47}&	\textbf{97.62 ± 0.23}&	\textbf{94.60 ± 0.41}&	\textbf{95.88 ± 0.17}\\
\hline
\end{tabular}
}
}
\label{tab:pint}
\end{table}

Next, we present the empirical comparison between \name\ and PINT \citep{DBLP:conf/nips/SouzaMKG22} in Table~\ref{tab:pint}. As shown in Table~\ref{tab:pint}, our \name\ outperforms PINT in all experimental settings (transductive and inductive). Regarding PINT's empirical evaluation, we adopt PINT's official implementation (via the GitHub link provided in the \citep{DBLP:conf/nips/SouzaMKG22}) and use the exact hyperparameters presented in the PINT paper to reproduce their results on these datasets. Moreover, PINT \citep{DBLP:conf/nips/SouzaMKG22} and \name\ use the same data split.

\section{Datasets}
\label{app:datasets}
Here, in Table~\ref{tab:data_statistics}, we report a detailed regarding $13$ datasets in our empirical assessment, which includes the domain, number of nodes, number of links, dimension of raw node and edge features. The statistics of Benchmark TGB are available at \url{https://tgb.complexdatalab.com/docs/linkprop/}.

\begin{table}[!htbp]
\centering
\caption{Statistics of the datasets.}
\vspace{1mm}
\resizebox{1\textwidth}{!}
{
\setlength{\tabcolsep}{0.45mm}
{
\begin{tabular}{c|cccccccc}
\hline
Datasets    & Domains     & \#Nodes & \#Links   & \#N\&L Feat & Bipartite & Duration  & Unique Steps & Time Granularity    \\ \hline
Wikipedia   & Social      & 9,227  & 157,474   & -- \& 172                & True      & 1 month    &  152,757      &  Unix timestamps   \\
Reddit      & Social      & 10,984 & 672,447   & -- \& 172                & True      & 1 month    &  669,065      &  Unix timestamps          \\
MOOC        & Interaction & 7,144  & 411,749   & -- \& 4                  & True      & 17 months    &  345,600      & Unix timestamps         \\
LastFM      & Interaction & 1,980  & 1,293,103 & -- \& --                 & True      & 1 month    &   1,283,614     & Unix timestamps           \\
Enron       & Social      & 184    & 125,235   & -- \& --                 & False     & 3 years    &   22,632     &    Unix timestamps        \\
Social Evo. & Proximity   & 74     & 2,099,519 & -- \& 2                  & False     & 8 months    &  565,932      &  Unix timestamps         \\
UCI         & Social      & 1,899  & 59,835    & -- \& --                 & False     & 196 days    &  58,911      &  Unix timestamps         \\
Flights     & Transport   & 13,169 & 1,927,145 & -- \& 1                  & False     & 4 months    &  122      &   days        \\
Can. Parl.  & Politics    & 734    & 74,478    & -- \& 1                  & False     & 14 years    &   14     &   years        \\
US Legis.   & Politics    & 225    & 60,396    & -- \& 1                  & False     & 12 congresses    &   12     &  congresses    \\
UN Trade    & Economics   & 255    & 507,497   & -- \& 1                  & False     & 32 years    &    32    &   years        \\
UN Vote     & Politics    & 201    & 1,035,742 & -- \& 1                  & False     & 72 years    &    72    &     years      \\
Contact     & Proximity   & 692    & 2,426,279 & -- \& 1                  & False     & 1 month    &    8,064    &   5 minutes         \\ \hline
\end{tabular}
}
}
\label{tab:data_statistics}
\end{table}

\section{More experimental results}
\label{sec: app_experiments}

\subsection{Empirical results for different negative sampling strategies} \label{app:link_prediction_nss}


We provide empirical results for (1) random NSS for inductive setting in Table~\ref{tab:inductive_random} (2) historical NSS for transductive setting in Table~\ref{tab:transductive_hist} and inductive setting in Table~\ref{tab:inductive_hist}, (3) inductive NSS under transductive setting in Table~\ref{tab:transductive_ind} and inductive setting in Table~\ref{tab:inductive_ind}.

\begin{table*}[!htbp]
\centering
\caption{Performance Comparison in the \textit{inductive setting} with \textit{random negative sampling strategy}.}
\vspace{1mm}
\resizebox{1.01\textwidth}{!}
{
\setlength{\tabcolsep}{0.9mm}
{
\begin{tabular}{c|c|cccccccccc}
\hline
Metric                     & Datasets    & JODIE                                            & DyRep            & TGAT             & TGN                                              & CAWN                                             & TCL              & GraphMixer       & DyGFormer             & FreeDyG                           & \name\                                           \\ \hline
                           & Wikipedia   & 94.82 $\pm$ 0.20                                 & 92.43 $\pm$ 0.37 & 96.22 $\pm$ 0.07 & 97.83 $\pm$ 0.04                                 & 98.24 $\pm$ 0.03                                 & 96.22 $\pm$ 0.17 & 96.65 $\pm$ 0.02 & 98.59 $\pm$ 0.03 & {\color[HTML]{329A9D} \textbf{98.97 $\pm$ 0.01}} & {\color[HTML]{9A0000} \textbf{99.15 $\pm$ 0.04}} \\
                           & Reddit      & 96.50 $\pm$ 0.13                                 & 96.09 $\pm$ 0.11 & 97.09 $\pm$ 0.04 & 97.50 $\pm$ 0.07                                 & 98.62 $\pm$ 0.01                                 & 94.09 $\pm$ 0.07 & 95.26 $\pm$ 0.02 & {\color[HTML]{329A9D} \textbf{98.84 $\pm$ 0.02}} & {\color[HTML]{9A0000} \textbf{98.91 $\pm$ 0.01}} & 98.02 $\pm$ 0.09 \\
                           & MOOC        & 79.63 $\pm$ 1.92                                 & 81.07 $\pm$ 0.44 & 85.50 $\pm$ 0.19 & {\color[HTML]{9A0000} \textbf{89.04 $\pm$ 1.17}} & 81.42 $\pm$ 0.24                                 & 80.60 $\pm$ 0.22 & 81.41 $\pm$ 0.21 & 86.96 $\pm$ 0.43                                 & 87.75 $\pm$ 0.62 & {\color[HTML]{329A9D} \textbf{88.49 $\pm$ 0.24}} \\
                           & LastFM      & 81.61 $\pm$ 3.82                                 & 83.02 $\pm$ 1.48 & 78.63 $\pm$ 0.31 & 81.45 $\pm$ 4.29                                 & 89.42 $\pm$ 0.07                                 & 73.53 $\pm$ 1.66 & 82.11 $\pm$ 0.42 & 94.23 $\pm$ 0.09 & {\color[HTML]{329A9D} \textbf{94.89 $\pm$ 0.01}} & {\color[HTML]{9A0000} \textbf{96.25 $\pm$ 0.43}} \\
                           & Enron       & 80.72 $\pm$ 1.39                                 & 74.55 $\pm$ 3.95 & 67.05 $\pm$ 1.51 & 77.94 $\pm$ 1.02                                 & 86.35 $\pm$ 0.51                                 & 76.14 $\pm$ 0.79 & 75.88 $\pm$ 0.48 & {\color[HTML]{9A0000} \textbf{89.76 $\pm$ 0.34}} & {\color[HTML]{329A9D} \textbf{89.69 $\pm$ 0.17}} & 89.17 $\pm$ 0.88 \\
                           & Social Evo. & 91.96 $\pm$ 0.48 & 90.04 $\pm$ 0.47 & 91.41 $\pm$ 0.16 & 90.77 $\pm$ 0.86                                 & 79.94 $\pm$ 0.18                                 & 91.55 $\pm$ 0.09 & 91.86 $\pm$ 0.06 & {\color[HTML]{329A9D} \textbf{93.14 $\pm$ 0.04}} & {\color[HTML]{9A0000} \textbf{94.76 $\pm$ 0.05}} & 91.71 $\pm$ 0.32                                 \\
                           & UCI         & 79.86 $\pm$ 1.48                                 & 57.48 $\pm$ 1.87 & 79.54 $\pm$ 0.48 & 88.12 $\pm$ 2.05                                 & 92.73 $\pm$ 0.06                                 & 87.36 $\pm$ 2.03 & 91.19 $\pm$ 0.42 & 94.54 $\pm$ 0.12 & {\color[HTML]{9A0000} \textbf{94.85 $\pm$ 0.10}} & {\color[HTML]{329A9D} \textbf{94.60 $\pm$ 0.41}} \\
                           & Flights     & 94.74 $\pm$ 0.37                                 & 92.88 $\pm$ 0.73 & 88.73 $\pm$ 0.33 & 95.03 $\pm$ 0.60                                 & 97.06 $\pm$ 0.02                                 & 83.41 $\pm$ 0.07 & 83.03 $\pm$ 0.05 & {\color[HTML]{9A0000} \textbf{97.79 $\pm$ 0.02}} & 96.42 $\pm$ 0.32 & {\color[HTML]{329A9D} \textbf{97.43 $\pm$ 0.15}} \\
                           & Can. Parl.  & 53.92 $\pm$ 0.94                                 & 54.02 $\pm$ 0.76 & 55.18 $\pm$ 0.79 & 54.10 $\pm$ 0.93                                 & 55.80 $\pm$ 0.69                                 & 54.30 $\pm$ 0.66 & 55.91 $\pm$ 0.82 & {\color[HTML]{329A9D} \textbf{87.74 $\pm$ 0.71}} & 52.96 $\pm$ 1.05 & {\color[HTML]{9A0000} \textbf{92.25$\pm$ 0.24}}  \\
                           & US Legis.   & 54.93 $\pm$ 2.29                                 & 57.28 $\pm$ 0.71 & 51.00 $\pm$ 3.11 & {\color[HTML]{329A9D} \textbf{58.63 $\pm$ 0.37}} & 53.17 $\pm$ 1.20                                 & 52.59 $\pm$ 0.97 & 50.71 $\pm$ 0.76 & 54.28 $\pm$ 2.87                                 & 53.05 $\pm$ 2.59 & {\color[HTML]{9A0000} \textbf{61.98 $\pm$ 1.31}} \\
                           & UN Trade    & 59.65 $\pm$ 0.77                                 & 57.02 $\pm$ 0.69 & 61.03 $\pm$ 0.18 & 58.31 $\pm$ 3.15                                 & {\color[HTML]{329A9D} \textbf{65.24 $\pm$ 0.21}} & 62.21 $\pm$ 0.12 & 62.17 $\pm$ 0.31 & 64.55 $\pm$ 0.62                                & - & {\color[HTML]{9A0000} \textbf{76.80 $\pm$ 2.01}} \\
                           & UN Vote     & 56.64 $\pm$ 0.96                                 & 54.62 $\pm$ 2.22 & 52.24 $\pm$ 1.46 & {\color[HTML]{329A9D} \textbf{58.85 $\pm$ 2.51}} & 49.94 $\pm$ 0.45                                 & 51.60 $\pm$ 0.97 & 50.68 $\pm$ 0.44 & 55.93 $\pm$ 0.39                                 & 50.51 $\pm$ 1.02  & {\color[HTML]{9A0000} \textbf{75.09 $\pm$ 0.34}} \\
                           & Contact     & 94.34 $\pm$ 1.45                                 & 92.18 $\pm$ 0.41 & 95.87 $\pm$ 0.11 & 93.82 $\pm$ 0.99                                 & 89.55 $\pm$ 0.30                                 & 91.11 $\pm$ 0.12 & 90.59 $\pm$ 0.05 & {\color[HTML]{9A0000} \textbf{98.03 $\pm$ 0.02}} & 97.66 $\pm$ 0.04 & {\color[HTML]{329A9D} \textbf{97.25 $\pm$ 0.07}} \\ \cline{2-12} 
\multirow{-14}{*}{AP}      & Avg. Rank   


& 6.38& 7.54& 7.08& 5.31& 5.38& 7.54& 6.92& {\color[HTML]{329A9D}\textbf{2.62}}& 4.15& {\color[HTML]{9A0000}\textbf{2.08}} \\

\hline
                           & Wikipedia   & 94.33 $\pm$ 0.27                                 & 91.49 $\pm$ 0.45 & 95.90 $\pm$ 0.09 & 97.72 $\pm$ 0.03                                 & 98.03 $\pm$ 0.04                                 & 95.57 $\pm$ 0.20 & 96.30 $\pm$ 0.04 & 98.48 $\pm$ 0.03 & {\color[HTML]{329A9D} \textbf{99.01 $\pm$ 0.02}} & {\color[HTML]{9A0000} \textbf{99.30 $\pm$ 0.03}} \\
                           & Reddit      & 96.52 $\pm$ 0.13                                 & 96.05 $\pm$ 0.12 & 96.98 $\pm$ 0.04 & 97.39 $\pm$ 0.07                                 & 98.42 $\pm$ 0.02                                 & 93.80 $\pm$ 0.07 & 94.97 $\pm$ 0.05 & {\color[HTML]{329A9D} \textbf{98.71 $\pm$ 0.01}} & {\color[HTML]{9A0000} \textbf{98.84 $\pm$ 0.01}} & 98.49 $\pm$ 0.06 \\
                           & MOOC        & 83.16 $\pm$ 1.30                                 & 84.03 $\pm$ 0.49 & 86.84 $\pm$ 0.17 & {\color[HTML]{9A0000} \textbf{91.24 $\pm$ 0.99}} & 81.86 $\pm$ 0.25                                 & 81.43 $\pm$ 0.19 & 82.77 $\pm$ 0.24 & 87.62 $\pm$ 0.51                                 & 87.01 $\pm$ 0.74 & {\color[HTML]{329A9D} \textbf{90.63 $\pm$ 0.21}} \\
                           & LastFM      & 81.13 $\pm$ 3.39                                 & 82.24 $\pm$ 1.51 & 76.99 $\pm$ 0.29 & 82.61 $\pm$ 3.15                                 & 87.82 $\pm$ 0.12                                 & 70.84 $\pm$ 0.85 & 80.37 $\pm$ 0.18 & 94.08 $\pm$ 0.08 & {\color[HTML]{329A9D} \textbf{94.32 $\pm$ 0.03}} & {\color[HTML]{9A0000} \textbf{97.66 $\pm$ 0.21}} \\
                           & Enron       & 81.96 $\pm$ 1.34                                 & 76.34 $\pm$ 4.20 & 64.63 $\pm$ 1.74 & 78.83 $\pm$ 1.11                                 & 87.02 $\pm$ 0.50                                 & 72.33 $\pm$ 0.99 & 76.51 $\pm$ 0.71 & {\color[HTML]{329A9D} \textbf{90.69 $\pm$ 0.26}} & 89.51 $\pm$ 0.20 & {\color[HTML]{9A0000} \textbf{92.30 $\pm$ 0.57}} \\
                           & Social Evo. & 93.70 $\pm$ 0.29                                 & 91.18 $\pm$ 0.49 & 93.41 $\pm$ 0.19 & 93.43 $\pm$ 0.59                                 & 84.73 $\pm$ 0.27                                 & 93.71 $\pm$ 0.18 & 94.09 $\pm$ 0.07 & {\color[HTML]{329A9D} \textbf{95.29 $\pm$ 0.03}} & {\color[HTML]{9A0000} \textbf{96.41 $\pm$ 0.07}} & 94.09 $\pm$ 0.24 \\
                           & UCI         & 78.80 $\pm$ 0.94                                 & 58.08 $\pm$ 1.81 & 77.64 $\pm$ 0.38 & 86.68 $\pm$ 2.29                                 & 90.40 $\pm$ 0.11                                 & 84.49 $\pm$ 1.82 & 89.30 $\pm$ 0.57 & 92.63 $\pm$ 0.13 & {\color[HTML]{329A9D} \textbf{93.01 $\pm$ 0.08}} & {\color[HTML]{9A0000} \textbf{95.88 $\pm$ 0.17}} \\
                           & Flights     & 95.21 $\pm$ 0.32                                 & 93.56 $\pm$ 0.70 & 88.64 $\pm$ 0.35 & 95.92 $\pm$ 0.43                                 & 96.86 $\pm$ 0.02                                 & 82.48 $\pm$ 0.01 & 82.27 $\pm$ 0.06 & {\color[HTML]{329A9D} \textbf{97.80 $\pm$ 0.02}} & 96.05 $\pm$ 0.29 & {\color[HTML]{9A0000} \textbf{98.46 $\pm$ 0.07}} \\
                           & Can. Parl.  & 53.81 $\pm$ 1.14                                 & 55.27 $\pm$ 0.49 & 56.51 $\pm$ 0.75 & 55.86 $\pm$ 0.75                                 & 58.83 $\pm$ 1.13                                 & 55.83 $\pm$ 1.07 & 58.32 $\pm$ 1.08 & {\color[HTML]{329A9D} \textbf{89.33 $\pm$ 0.48}} & 52.89 $\pm$ 1.61 & {\color[HTML]{9A0000} \textbf{95.06 $\pm$ 0.11}} \\
                           & US Legis.   & 58.12 $\pm$ 2.35                                 & 61.07 $\pm$ 0.56 & 48.27 $\pm$ 3.50 & {\color[HTML]{329A9D} \textbf{62.38 $\pm$ 0.48}} & 51.49 $\pm$ 1.13                                 & 50.43 $\pm$ 1.48 & 47.20 $\pm$ 0.89 & 53.21 $\pm$ 3.04                                  & 53.01 $\pm$ 3.14 & {\color[HTML]{9A0000} \textbf{66.18 $\pm$ 1.19}} \\
                           & UN Trade    & 62.28 $\pm$ 0.50                                 & 58.82 $\pm$ 0.98 & 62.72 $\pm$ 0.12 & 59.99 $\pm$ 3.50                                 & 67.05 $\pm$ 0.21                                 & 63.76 $\pm$ 0.07 & 63.48 $\pm$ 0.37 & {\color[HTML]{329A9D} \textbf{67.25 $\pm$ 1.05}} & - & {\color[HTML]{9A0000} \textbf{83.95 $\pm$ 1.76}} \\
                           & UN Vote     & 58.13 $\pm$ 1.43                                 & 55.13 $\pm$ 3.46 & 51.83 $\pm$ 1.35 & {\color[HTML]{329A9D} \textbf{61.23 $\pm$ 2.71}} & 48.34 $\pm$ 0.76                                 & 50.51 $\pm$ 1.05 & 50.04 $\pm$ 0.86 & 56.73 $\pm$ 0.69                                 & 49.97 $\pm$ 1.22 & {\color[HTML]{9A0000} \textbf{82.01 $\pm$ 0.21}} \\
                           & Contact     & 95.37 $\pm$ 0.92                                 & 91.89 $\pm$ 0.38 & 96.53 $\pm$ 0.10 & 94.84 $\pm$ 0.75                                 & 89.07 $\pm$ 0.34                                 & 93.05 $\pm$ 0.09 & 92.83 $\pm$ 0.05 & {\color[HTML]{9A0000} \textbf{98.30 $\pm$ 0.02}} & {\color[HTML]{329A9D} \textbf{98.11 $\pm$ 0.02}} & 97.99 $\pm$ 0.07 \\ \cline{2-12} 
\multirow{-14}{*}{ROC-AUC} & Avg. Rank   


& 6.38& 7.54& 7.08& 4.92& 5.77& 7.77& 7.08& {\color[HTML]{329A9D}\textbf{2.62}}& 4.31& {\color[HTML]{9A0000}\textbf{1.54}} \\

\hline
\end{tabular}
}
}
\label{tab:inductive_random}
\end{table*}

\begin{table}[h]
\centering
\caption{Performance comparison in the \textit{transductive setting} with \textit{historical negative sampling strategy}.}
\vspace{1mm}
\resizebox{1.01\textwidth}{!}
{
\setlength{\tabcolsep}{0.9mm}
{
\begin{tabular}{c|c|cccccccccc}
\hline
Metric                     & Datasets    & JODIE                                            & DyRep                                            & TGAT                                             & TGN                                              & CAWN             & TCL              & GraphMixer                                       & DyGFormer & FreeDyG                                         & \name\                                           \\ \hline
                           & Wikipedia   & 83.01 $\pm$ 0.66                                 & 79.93 $\pm$ 0.56                                 & 87.38 $\pm$ 0.22                                 & 86.86 $\pm$ 0.33                                 & 71.21 $\pm$ 1.67 & 89.05 $\pm$ 0.39 & 90.90 $\pm$ 0.10 & 82.23 $\pm$ 2.54                                & {\color[HTML]{329A9D} \textbf{91.59 $\pm$ 0.57}} & {\color[HTML]{9A0000} \textbf{99.10 $\pm$ 0.51}} \\
                           
                           & Reddit      & 80.03 $\pm$ 0.36                                 & 79.83 $\pm$ 0.31                                 & 79.55 $\pm$ 0.20                            & 81.22 $\pm$ 0.61                                 & 80.82 $\pm$ 0.45 & 77.14 $\pm$ 0.16 & 78.44 $\pm$ 0.18                                 & 81.57 $\pm$ 0.67 & {\color[HTML]{329A9D} \textbf{85.67 $\pm$ 1.01}} &{\color[HTML]{9A0000} \textbf{93.25 $\pm$ 1.13}} \\

                           & MOOC        & 78.94 $\pm$ 1.25                                 & 75.60 $\pm$ 1.12                                 & 82.19 $\pm$ 0.62                                 & {\color[HTML]{329A9D} \textbf{87.06 $\pm$ 1.93}} & 74.05 $\pm$ 0.95 & 77.06 $\pm$ 0.41 & 77.77 $\pm$ 0.92                                 & 85.85 $\pm$ 0.66                                 & 86.71 $\pm$ 0.81 & {\color[HTML]{9A0000} \textbf{96.52 $\pm$ 1.58}} \\
                           & LastFM      & 74.35 $\pm$ 3.81                                 & 74.92 $\pm$ 2.46                                 & 71.59 $\pm$ 0.24                                 & 76.87 $\pm$ 4.64                                 & 69.86 $\pm$ 0.43 & 59.30 $\pm$ 2.31 & 72.47 $\pm$ 0.49                                 & {\color[HTML]{329A9D} \textbf{81.57 $\pm$ 0.48}} & 79.71 $\pm$ 0.51 &{\color[HTML]{9A0000} \textbf{94.88 $\pm$ 1.35}} \\
                           & Enron       & 69.85 $\pm$ 2.70                                 & 71.19 $\pm$ 2.76                                 & 64.07 $\pm$ 1.05                                 & 73.91 $\pm$ 1.76                                 & 64.73 $\pm$ 0.36 & 70.66 $\pm$ 0.39 & 77.98 $\pm$ 0.92 & 75.63 $\pm$ 0.73                                 & {\color[HTML]{329A9D} \textbf{78.87 $\pm$ 0.82}} &{\color[HTML]{9A0000} \textbf{95.42 $\pm$ 0.53}} \\
                           & Social Evo. & 87.44 $\pm$ 6.78                                 & 93.29 $\pm$ 0.43                                 & 95.01 $\pm$ 0.44 & 94.45 $\pm$ 0.56                                 & 85.53 $\pm$ 0.38 & 94.74 $\pm$ 0.31 &  94.93 $\pm$ 0.31                                 & {\color[HTML]{329A9D} \textbf{97.38 $\pm$ 0.14}} & {\color[HTML]{9A0000} \textbf{97.79 $\pm$ 0.23}} & 90.19 $\pm$ 2.01                                 \\
                           
                           & UCI         & 75.24 $\pm$ 5.80                                 & 55.10 $\pm$ 3.14                                 & 68.27 $\pm$ 1.37                                 & 80.43 $\pm$ 2.12                                 & 65.30 $\pm$ 0.43 & 80.25 $\pm$ 2.74 & {\color[HTML]{329A9D} \textbf{84.11 $\pm$ 1.35}} & 82.17 $\pm$ 0.82 & {\color[HTML]{9A0000} \textbf{86.10 $\pm$ 1.19}} & 80.76 $\pm$ 1.88                                 \\
                           
                           & Flights     & 66.48 $\pm$ 2.59                                 & 67.61 $\pm$ 0.99                                 & {\color[HTML]{329A9D} \textbf{72.38 $\pm$ 0.18}} & 66.70 $\pm$ 1.64                                 & 64.72 $\pm$ 0.97 & 70.68 $\pm$ 0.24 & 71.47 $\pm$ 0.26                                 & 66.59 $\pm$ 0.49                                 & 66.13 $\pm$ 1.23 & {\color[HTML]{9A0000} \textbf{82.02 $\pm$ 2.39}} \\
                           
                           & Can. Parl.  & 51.79 $\pm$ 0.63                                 & 63.31 $\pm$ 1.23                                 & 67.13 $\pm$ 0.84                                 & 68.42 $\pm$ 3.07                                 & 66.53 $\pm$ 2.77 & 65.93 $\pm$ 3.00 & 74.34 $\pm$ 0.87                                 & {\color[HTML]{9A0000} \textbf{97.00 $\pm$ 0.31}} & 72.22 $\pm$ 2.47 & {\color[HTML]{329A9D} \textbf{75.03 $\pm$ 1.76}} \\
                           
                           & US Legis.   & 51.71 $\pm$ 5.76                                 & {\color[HTML]{9A0000} \textbf{86.88 $\pm$ 2.25}} & 62.14 $\pm$ 6.60                                 & 74.00 $\pm$ 7.57                                 & 68.82 $\pm$ 8.23 & 80.53 $\pm$ 3.95 & 81.65 $\pm$ 1.02                                 & {\color[HTML]{329A9D} \textbf{85.30 $\pm$ 3.88}} & 69.94 $\pm$ 0.59 & 62.66 $\pm$ 3.54                                 \\
                           
                           & UN Trade    & 61.39 $\pm$ 1.83                                 & 59.19 $\pm$ 1.07                                 & 55.74 $\pm$ 0.91                                 & 58.44 $\pm$ 5.51                                 & 55.71 $\pm$ 0.38 & 55.90 $\pm$ 1.17 & 57.05 $\pm$ 1.22                                 & {\color[HTML]{329A9D} \textbf{64.41 $\pm$ 1.40}} & - & {\color[HTML]{9A0000} \textbf{80.00 $\pm$ 1.55}} \\
                           
                           & UN Vote     & {\color[HTML]{9A0000} \textbf{70.02 $\pm$ 0.81}} & 69.30 $\pm$ 1.12                                 & 52.96 $\pm$ 2.14                                 & {\color[HTML]{329A9D} \textbf{69.37 $\pm$ 3.93}} & 51.26 $\pm$ 0.04 & 52.30 $\pm$ 2.35 & 51.20 $\pm$ 1.60                                 & 60.84 $\pm$ 1.58                                 & 50.61 $\pm$ 2.46 & 61.43 $\pm$ 1.07                                 \\
                           
                           & Contact     & 95.31 $\pm$ 2.13                                 & 96.39 $\pm$ 0.20 & 96.05 $\pm$ 0.52                                 & 93.05 $\pm$ 2.35                                 & 84.16 $\pm$ 0.49 & 93.86 $\pm$ 0.21 & 93.36 $\pm$ 0.41                                 & {\color[HTML]{9A0000} \textbf{97.57 $\pm$ 0.06}} & {\color[HTML]{329A9D} \textbf{97.32 $\pm$ 0.48}} & 80.24 $\pm$ 1.98                                 \\ \cline{2-12} 
\multirow{-14}{*}{AP}      & Avg. Rank   

& 6.62& 6.0& 6.31& 4.85& 8.62& 6.62& 5.08& {\color[HTML]{329A9D}\textbf{3.38}}& 4.23& {\color[HTML]{9A0000}\textbf{3.31}} \\

\hline
                           
                           & Wikipedia   & 80.77 $\pm$ 0.73                                 & 77.74 $\pm$ 0.33                                 & 82.87 $\pm$ 0.22                                 & 82.74 $\pm$ 0.32                                 & 67.84 $\pm$ 0.64 & 85.76 $\pm$ 0.46 & {\color[HTML]{329A9D} \textbf{87.68 $\pm$ 0.17}} & 78.80 $\pm$ 1.95                                 &  82.78 $\pm$ 0.30 & {\color[HTML]{9A0000} \textbf{99.21 $\pm$ 0.48}} \\
                           & Reddit      & 80.52 $\pm$ 0.32                                 & 80.15 $\pm$ 0.18                                 & 79.33 $\pm$ 0.16                                 & 81.11 $\pm$ 0.19 & 80.27 $\pm$ 0.30 & 76.49 $\pm$ 0.16 & 77.80 $\pm$ 0.12                                 & 80.54 $\pm$ 0.29                                 & {\color[HTML]{329A9D} \textbf{85.92 $\pm$ 0.10}} & {\color[HTML]{9A0000} \textbf{96.40 $\pm$ 0.59}} \\
                           & MOOC        & 82.75 $\pm$ 0.83                                 & 81.06 $\pm$ 0.94                                 & 80.81 $\pm$ 0.67                                 & 88.00 $\pm$ 1.80 & 71.57 $\pm$ 1.07 & 72.09 $\pm$ 0.56 & 76.68 $\pm$ 1.40                                 & 87.04 $\pm$ 0.35                                 & {\color[HTML]{329A9D} \textbf{88.32 $\pm$ 0.99}} & {\color[HTML]{9A0000} \textbf{97.69 $\pm$ 0.94}} \\
                           & LastFM      & 75.22 $\pm$ 2.36                                 & 74.65 $\pm$ 1.98                                 & 64.27 $\pm$ 0.26                                 & 77.97 $\pm$ 3.04                                 & 67.88 $\pm$ 0.24 & 47.24 $\pm$ 3.13 & 64.21 $\pm$ 0.73                                 & {\color[HTML]{329A9D} \textbf{78.78 $\pm$ 0.35}} & 73.53 $\pm$ 0.12 & {\color[HTML]{9A0000} \textbf{97.66 $\pm$ 1.96}} \\
                           & Enron       & 75.39 $\pm$ 2.37                                 & 74.69 $\pm$ 3.55                                 & 61.85 $\pm$ 1.43                                 & {\color[HTML]{329A9D} \textbf{77.09 $\pm$ 2.22}} & 65.10 $\pm$ 0.34 & 67.95 $\pm$ 0.88 & 75.27 $\pm$ 1.14                                 & 76.55 $\pm$ 0.52                                 & 75.74 $\pm$ 0.72 & {\color[HTML]{9A0000} \textbf{95.79 $\pm$ 0.33}} \\
                           & Social Evo. & 90.06 $\pm$ 3.15                                 & 93.12 $\pm$ 0.34                                 & 93.08 $\pm$ 0.59                                 & 94.71 $\pm$ 0.53 & 87.43 $\pm$ 0.15 & 93.44 $\pm$ 0.68 & 94.39 $\pm$ 0.31                                 & {\color[HTML]{329A9D} \textbf{97.28 $\pm$ 0.07}} & {\color[HTML]{9A0000} \textbf{97.42 $\pm$ 0.02}} & 93.60 $\pm$ 2.27                                 \\
                           & UCI         & 78.64 $\pm$ 3.50 & 57.91 $\pm$ 3.12                                 & 58.89 $\pm$ 1.57                                 & 77.25 $\pm$ 2.68                                 & 57.86 $\pm$ 0.15 & 72.25 $\pm$ 3.46 & 77.54 $\pm$ 2.02                                 & 76.97 $\pm$ 0.24                                 & {\color[HTML]{329A9D} \textbf{80.38 $\pm$ 0.26}} & {\color[HTML]{9A0000} \textbf{90.59 $\pm$ 1.60}} \\
                           & Flights     & 68.97 $\pm$ 1.87                                 & 69.43 $\pm$ 0.90                                 & {\color[HTML]{329A9D} \textbf{72.20 $\pm$ 0.16}} & 68.39 $\pm$ 0.95                                 & 66.11 $\pm$ 0.71 & 70.57 $\pm$ 0.18 & 70.37 $\pm$ 0.23                                 & 68.09 $\pm$ 0.43                                 & 67.83 $\pm$ 1.46 & {\color[HTML]{9A0000} \textbf{92.84 $\pm$ 1.21}} \\
                           & Can. Parl.  & 62.44 $\pm$ 1.11                                 & 70.16 $\pm$ 1.70                                 & 70.86 $\pm$ 0.94                                 & 73.23 $\pm$ 3.08                                 & 72.06 $\pm$ 3.94 & 69.95 $\pm$ 3.70 & 79.03 $\pm$ 1.01                                 & {\color[HTML]{9A0000} \textbf{97.61 $\pm$ 0.40}} & 77.59 $\pm$ 6.22 & {\color[HTML]{329A9D} \textbf{87.22 $\pm$ 1.85}} \\
                           & US Legis.   & 67.47 $\pm$ 6.40                                 & {\color[HTML]{9A0000} \textbf{91.44 $\pm$ 1.18}} & 73.47 $\pm$ 5.25                                 & 83.53 $\pm$ 4.53                                 & 78.62 $\pm$ 7.46 & 83.97 $\pm$ 3.71 & 85.17 $\pm$ 0.70                                 & {\color[HTML]{329A9D} \textbf{90.77 $\pm$ 1.96}} & 63.49 $\pm$ 13.04 & 76.64 $\pm$ 3.38                                 \\
                           & UN Trade    & 68.92 $\pm$ 1.40                                 & 64.36 $\pm$ 1.40                                 & 60.37 $\pm$ 0.68                                 & 63.93 $\pm$ 5.41                                 & 63.09 $\pm$ 0.74 & 61.43 $\pm$ 1.04 & 63.20 $\pm$ 1.54                                 & {\color[HTML]{329A9D} \textbf{73.86 $\pm$ 1.13}} & - & {\color[HTML]{9A0000} \textbf{87.67 $\pm$ 1.80}} \\
                           & UN Vote     & {\color[HTML]{9A0000} \textbf{76.84 $\pm$ 1.01}} & {\color[HTML]{329A9D} \textbf{74.72 $\pm$ 1.43}} & 53.95 $\pm$ 3.15                                 & 73.40 $\pm$ 5.20                                 & 51.27 $\pm$ 0.33 & 52.29 $\pm$ 2.39 & 52.61 $\pm$ 1.44                                 & 64.27 $\pm$ 1.78                                 & 52.76 $\pm$ 2.85 & 72.92 $\pm$ 1.73                                 \\
                           & Contact     & {\color[HTML]{329A9D} \textbf{96.35 $\pm$ 0.92}} & 96.00 $\pm$ 0.23                                 & 95.39 $\pm$ 0.43                                 & 93.76 $\pm$ 1.29                                 & 83.06 $\pm$ 0.32 & 93.34 $\pm$ 0.19 & 93.14 $\pm$ 0.34                                 & {\color[HTML]{9A0000} \textbf{97.17 $\pm$ 0.05}} & 97.12 $\pm$ 0.24 & 90.99 $\pm$ 1.80                                 \\ \cline{2-12} 
\multirow{-14}{*}{ROC-AUC} & Avg. Rank   


& 5.38& 5.69& 6.92& 4.31& 8.54& 7.15& 5.69& {\color[HTML]{329A9D} \textbf{3.69}}& 4.92& {\color[HTML]{9A0000} \textbf{2.69}} \\

\hline
\end{tabular}
}
}
\label{tab:transductive_hist}
\end{table}

\begin{table}[h]
\centering
\caption{Performance comparison in the \textit{inductive setting} with \textit{historical negative sampling strategy}.}
\vspace{1mm}
\resizebox{1.01\textwidth}{!}
{
\setlength{\tabcolsep}{0.9mm}
{
\begin{tabular}{c|c|cccccccccc}
\hline
Metric                     & Datasets    & JODIE                      & DyRep                                            & TGAT                                                       & TGN                                                        & CAWN                                             & TCL                                                        & GraphMixer                                                 & DyGFormer                                              & FreeDyG    & \name\                                    \\ \hline
                           & Wikipedia   & 68.69 $\pm$ 0.39           & 62.18 $\pm$ 1.27                                 & 84.17 $\pm$ 0.22                                 & 81.76 $\pm$ 0.32                                           & 67.27 $\pm$ 1.63                                 & 82.20 $\pm$ 2.18                                           & {\color[HTML]{329A9D} \textbf{87.60 $\pm$ 0.30}}           & 71.42 $\pm$ 4.43                                           &  82.78 $\pm$0.30 & {\color[HTML]{9A0000} \textbf{98.45 $\pm$ 0.88}} \\
                           & Reddit      & 62.34 $\pm$ 0.54           & 61.60 $\pm$ 0.72                                 & 63.47 $\pm$ 0.36                                           & 64.85 $\pm$ 0.85                                 & 63.67 $\pm$ 0.41                                 & 60.83 $\pm$ 0.25                                           & 64.50 $\pm$ 0.26                                           & 65.37 $\pm$ 0.60           & {\color[HTML]{329A9D} \textbf{66.02 $\pm$ 0.41}} & {\color[HTML]{9A0000} \textbf{85.74 $\pm$ 2.46}} \\
                           & MOOC        & 63.22 $\pm$ 1.55           & 62.93 $\pm$ 1.24                                 & 76.73 $\pm$ 0.29                                           & 77.07 $\pm$ 3.41                                 & 74.68 $\pm$ 0.68                                 & 74.27 $\pm$ 0.53                                           & 74.00 $\pm$ 0.97                                           & 80.82 $\pm$ 0.30           & {\color[HTML]{329A9D} \textbf{81.63 $\pm$ 0.33}} & {\color[HTML]{9A0000} \textbf{95.17 $\pm$ 2.78}} \\
                           & LastFM      & 70.39 $\pm$ 4.31           & 71.45 $\pm$ 1.76                                 & 76.27 $\pm$ 0.25                                           & 66.65 $\pm$ 6.11                                           & 71.33 $\pm$ 0.47                                 & 65.78 $\pm$ 0.65                                           & 76.42 $\pm$ 0.22           & 76.35 $\pm$ 0.52                                 & {\color[HTML]{329A9D} \textbf{77.28 $\pm$ 0.21}} & {\color[HTML]{9A0000} \textbf{92.96 $\pm$ 1.08}} \\
                           & Enron       & 65.86 $\pm$ 3.71           & 62.08 $\pm$ 2.27                                 & 61.40 $\pm$ 1.31                                           & 62.91 $\pm$ 1.16                                           & 60.70 $\pm$ 0.36                                 & 67.11 $\pm$ 0.62                                 & 72.37 $\pm$ 1.37           & 67.07 $\pm$ 0.62                                           & {\color[HTML]{329A9D} \textbf{73.01 $\pm$ 0.88}} & {\color[HTML]{9A0000} \textbf{89.37 $\pm$ 1.01}} \\
                           & Social Evo. & 88.51 $\pm$ 0.87           & 88.72 $\pm$ 1.10                                 & 93.97 $\pm$ 0.54                                           & 90.66 $\pm$ 1.62                                           & 79.83 $\pm$ 0.38                                 & {\color[HTML]{329A9D} \textbf{94.10 $\pm$ 0.31}} & 94.01 $\pm$ 0.47                                           & {\color[HTML]{9A0000} \textbf{96.82 $\pm$ 0.16}}           & 96.69 $\pm$ 0.14 & 85.13 $\pm$ 1.05                                 \\
                           & UCI         & 63.11 $\pm$ 2.27           & 52.47 $\pm$ 2.06                                 & 70.52 $\pm$ 0.93                                           & 70.78 $\pm$ 0.78                                           & 64.54 $\pm$ 0.47                                 & 76.71 $\pm$ 1.00 & {\color[HTML]{329A9D} \textbf{81.66 $\pm$ 0.49}}           & 72.13 $\pm$ 1.87                                           & {\color[HTML]{9A0000} \textbf{82.35 $\pm$ 0.39}} & 75.97 $\pm$ 1.13                                 \\
                           & Flights     & 61.01 $\pm$ 1.65           & 62.83 $\pm$ 1.31                                 & 64.72 $\pm$ 0.36                                 & 59.31 $\pm$ 1.43                                           & 56.82 $\pm$ 0.57                                 & 64.50 $\pm$ 0.25                                           & {\color[HTML]{329A9D} \textbf{65.28 $\pm$ 0.24}}           & 57.11 $\pm$ 0.21                                           & 56.72 $\pm$ 1.30 &  {\color[HTML]{9A0000} \textbf{71.54 $\pm$ 2.12}} \\
                           & Can. Parl.  & 52.60 $\pm$ 0.88           & 52.28 $\pm$ 0.31                                 & 56.72 $\pm$ 0.47                                           & 54.42 $\pm$ 0.77                                           & 57.14 $\pm$ 0.07                       & 55.71 $\pm$ 0.74                                           & 55.84 $\pm$ 0.73                                           & {\color[HTML]{9A0000} \textbf{87.40 $\pm$ 0.85}}           & 52.15 $\pm$ 0.84 & {\color[HTML]{329A9D} \textbf{68.91 $\pm$ 1.63}} \\
                           & US Legis.   & 52.94 $\pm$ 2.11           & {\color[HTML]{9A0000} \textbf{62.10 $\pm$ 1.41}} & 51.83 $\pm$ 3.95                                           & {\color[HTML]{329A9D} \textbf{61.18 $\pm$ 1.10}} & 55.56 $\pm$ 1.71                                 & 53.87 $\pm$ 1.41                                           & 52.03 $\pm$ 1.02                                           & 56.31 $\pm$ 3.46                                           & 53.63 $\pm$ 4.38 & 49.51 $\pm$ 1.09                                 \\
                           & UN Trade    & 55.46 $\pm$ 1.19           & 55.49 $\pm$ 0.84                       & 55.28 $\pm$ 0.71                                           & 52.80 $\pm$ 3.19                                           & 55.00 $\pm$ 0.38                                 & {\color[HTML]{329A9D} \textbf{55.76 $\pm$ 1.03}}           & 54.94 $\pm$ 0.97                                           & 53.20 $\pm$ 1.07                                           & - & {\color[HTML]{9A0000} \textbf{79.09 $\pm$ 1.86}} \\
                           & UN Vote     & 61.04 $\pm$ 1.30 & 60.22 $\pm$ 1.78                                 & 53.05 $\pm$ 3.10                                           & {\color[HTML]{9A0000} \textbf{63.74 $\pm$ 3.00}}           & 47.98 $\pm$ 0.84                                 & 54.19 $\pm$ 2.17                                           & 48.09 $\pm$ 0.43                                           & 52.63 $\pm$ 1.26                                           & 50.23 $\pm$ 1.65  & {\color[HTML]{329A9D} \textbf{61.89 $\pm$ 1.28}} \\
                           & Contact     & 90.42 $\pm$ 2.34           & 89.22 $\pm$ 0.66                                 & {\color[HTML]{9A0000} \textbf{94.15 $\pm$ 0.45}}           & 88.13 $\pm$ 1.50                                           & 74.20 $\pm$ 0.80                                 & 90.44 $\pm$ 0.17                                           & 89.91 $\pm$ 0.36                                           & {\color[HTML]{329A9D} \textbf{93.56 $\pm$ 0.52}} & 93.26 $\pm$ 1.42 & 79.08 $\pm$ 2.24                                 \\ \cline{2-12}
\multirow{-14}{*}{AP}      & Avg. Rank   


& 6.85& 6.85& 5.31& 5.85& 7.54& 5.23& 4.92& {\color[HTML]{329A9D}\textbf{4.38}}& 4.77& {\color[HTML]{9A0000}\textbf{3.31}} \\

\hline
                           & Wikipedia   & 61.86 $\pm$ 0.53           & 57.54 $\pm$ 1.09                                 & 78.38 $\pm$ 0.20                                           & 75.75 $\pm$ 0.29                                           & 62.04 $\pm$ 0.65                                 & 79.79 $\pm$ 0.96                                 & {\color[HTML]{329A9D} \textbf{82.87 $\pm$ 0.21}}           & 68.33 $\pm$ 2.82                                           & 82.08 $\pm$ 0.32 & {\color[HTML]{9A0000} \textbf{98.59 $\pm$ 0.82}} \\
                           & Reddit      & 61.69 $\pm$ 0.39           & 60.45 $\pm$ 0.37                                 & 64.43 $\pm$ 0.27                                           & 64.55 $\pm$ 0.50                                           & 64.94 $\pm$ 0.21 & 61.43 $\pm$ 0.26                                           & 64.27 $\pm$ 0.13                                           & 64.81 $\pm$ 0.25                                 &  {\color[HTML]{329A9D} \textbf{66.79 $\pm$ 0.31}} & {\color[HTML]{9A0000} \textbf{92.04 $\pm$ 1.87}} \\
                           & MOOC        & 64.48 $\pm$ 1.64           & 64.23 $\pm$ 1.29                                 & 74.08 $\pm$ 0.27                                           & 77.69 $\pm$ 3.55                                 & 71.68 $\pm$ 0.94                                 & 69.82 $\pm$ 0.32                                           & 72.53 $\pm$ 0.84                                           & 80.77 $\pm$ 0.63           & {\color[HTML]{329A9D} \textbf{81.52 $\pm$ 0.37}} & {\color[HTML]{9A0000} \textbf{96.85 $\pm$ 1.59}} \\
                           & LastFM      & 68.44 $\pm$ 3.26           & 68.79 $\pm$ 1.08                                 & 69.89 $\pm$ 0.28                                           & 66.99 $\pm$ 5.62                                           & 67.69 $\pm$ 0.24                                 & 55.88 $\pm$ 1.85                                           & 70.07 $\pm$ 0.20                                 & 70.73 $\pm$ 0.37           & {\color[HTML]{329A9D} \textbf{72.63 $\pm$ 0.16}} & {\color[HTML]{9A0000} \textbf{96.34 $\pm$ 1.68}} \\
                           & Enron       & 65.32 $\pm$ 3.57           & 61.50 $\pm$ 2.50                                 & 57.84 $\pm$ 2.18                                           & 62.68 $\pm$ 1.09                                           & 62.25 $\pm$ 0.40                                 & 64.06 $\pm$ 1.02                                           & 68.20 $\pm$ 1.62           & 65.78 $\pm$ 0.42                                 & {\color[HTML]{329A9D} \textbf{70.09 $\pm$ 0.65}} & {\color[HTML]{9A0000} \textbf{92.14 $\pm$ 0.70}} \\
                           & Social Evo. & 88.53 $\pm$ 0.55           & 87.93 $\pm$ 1.05                                 & 91.87 $\pm$ 0.72                                           & 92.10 $\pm$ 1.22                                           & 83.54 $\pm$ 0.24                                 & 93.28 $\pm$ 0.60                                           & 93.62 $\pm$ 0.35 & {\color[HTML]{329A9D} \textbf{96.91 $\pm$ 0.09}}           & {\color[HTML]{9A0000} \textbf{96.94 $\pm$ 0.17}} & 90.08 $\pm$ 2.91                                 \\
                           & UCI         & 60.24 $\pm$ 1.94           & 51.25 $\pm$ 2.37                                 & 62.32 $\pm$ 1.18                                           & 62.69 $\pm$ 0.90                                           & 56.39 $\pm$ 0.10                                 & 70.46 $\pm$ 1.94                                 & 75.98 $\pm$ 0.84           & 65.55 $\pm$ 1.01                                           & {\color[HTML]{329A9D} \textbf{76.01 $\pm$ 0.75}} & {\color[HTML]{9A0000} \textbf{84.63 $\pm$ 1.23}} \\
                           & Flights     & 60.72 $\pm$ 1.29           & 61.99 $\pm$ 1.39                                 & 63.38 $\pm$ 0.26                                 & 59.66 $\pm$ 1.04                                           & 56.58 $\pm$ 0.44                                 & {\color[HTML]{329A9D} \textbf{63.48 $\pm$ 0.23}}           & 63.30 $\pm$ 0.19                                           & 56.05 $\pm$ 0.21                                           & 55.20 $\pm$ 1.16 & {\color[HTML]{9A0000} \textbf{85.60 $\pm$ 1.47}} \\
                           & Can. Parl.  & 51.62 $\pm$ 1.00           & 52.38 $\pm$ 0.46                                 & 58.30 $\pm$ 0.61                                           & 55.64 $\pm$ 0.54                                           & 60.11 $\pm$ 0.48                       & 57.30 $\pm$ 1.03                                           & 56.68 $\pm$ 1.20                                           & {\color[HTML]{9A0000} \textbf{88.68 $\pm$ 0.74}}           & 51.20 $\pm$ 1.99 & {\color[HTML]{329A9D} \textbf{80.73 $\pm$ 1.09}}                                 \\
                           
                           & US Legis.   & 58.12 $\pm$ 2.94           & {\color[HTML]{9A0000}\textbf{67.94 $\pm$ 0.98}}                        & 49.99 $\pm$ 4.88                                           & {\color[HTML]{329A9D}\textbf{64.87 $\pm$ 1.65}}                                 & 54.41 $\pm$ 1.31                                 & 52.12 $\pm$ 2.13                                           & 49.28 $\pm$ 0.86                                           & 56.57 $\pm$ 3.22                                           & 53.75 $\pm$ 5.92 & 55.74 $\pm$ 1.53                                 \\
                           & UN Trade    & 58.73 $\pm$ 1.19           & 57.90 $\pm$ 1.33                                 & 59.74 $\pm$ 0.59                                           & 55.61 $\pm$ 3.54                                           & 60.95 $\pm$ 0.80                       & {\color[HTML]{329A9D} \textbf{61.12 $\pm$ 0.97}}           & 59.88 $\pm$ 1.17                                           & 58.46 $\pm$ 1.65                                           & - & {\color[HTML]{9A0000} \textbf{86.26 $\pm$ 1.78}} \\
                           
                           & UN Vote     & 65.16 $\pm$ 1.28 & 63.98 $\pm$ 2.12                                 & 51.73 $\pm$ 4.12                                           & {\color[HTML]{329A9D} \textbf{68.59 $\pm$ 3.11}}           & 48.01 $\pm$ 1.77                                 & 54.66 $\pm$ 2.11                                           & 45.49 $\pm$ 0.42                                           & 53.85 $\pm$ 2.02                                           & 48.98 $\pm$ 2.44 & {\color[HTML]{9A0000} \textbf{72.98 $\pm$ 1.22}} \\
                           
                           & Contact     & 90.80 $\pm$ 1.18           & 88.88 $\pm$ 0.68                                 & 93.76 $\pm$ 0.41 & 88.84 $\pm$ 1.39                                           & 74.79 $\pm$ 0.37                                 & 90.37 $\pm$ 0.16                                           & 90.04 $\pm$ 0.29                                           & {\color[HTML]{329A9D} \textbf{94.14 $\pm$ 0.26}}           & {\color[HTML]{9A0000} \textbf{94.27 $\pm$ 0.77}} & 88.69 $\pm$ 2.24                                 \\ \cline{2-12}
\multirow{-14}{*}{ROC-AUC} & Avg. Rank   


& 6.54& 7.46& 5.77& 5.92& 7.08& 5.54& 5.23& {\color[HTML]{329A9D}\textbf{4.38}}& 4.62& {\color[HTML]{9A0000}\textbf{2.46}} \\

\hline

\end{tabular}
}
}
\label{tab:inductive_hist}
\end{table}

\begin{table}[h]
\centering
\caption{Performance comparison in the \textit{transductive setting} with \textit{inductive negative sampling strategy}.}
\vspace{1mm}
\resizebox{1.01\textwidth}{!}
{
\setlength{\tabcolsep}{0.9mm}
{
\begin{tabular}{c|c|cccccccccc}
\hline
Metric                     & Datasets    & JODIE                                            & DyRep                                            & TGAT                                             & TGN                                              & CAWN             & TCL              & GraphMixer                                       & DyGFormer                                        & FreeDyG & \name\                                           \\ \hline
                           & Wikipedia   & 75.65 $\pm$ 0.79                                           & 70.21 $\pm$ 1.58                                 & 87.00 $\pm$ 0.16                                 & 85.62 $\pm$ 0.44                                           & 74.06 $\pm$ 2.62                                           & 86.76 $\pm$ 0.72                                           & 88.59 $\pm$ 0.17 & 78.29 $\pm$ 5.38                                           & {\color[HTML]{329A9D} \textbf{90.05 $\pm$ 0.79}} & {\color[HTML]{9A0000} \textbf{98.66 $\pm$ 0.94}} \\
                           
                           & Reddit      & 86.98 $\pm$ 0.16                                           & 86.30 $\pm$ 0.26                                 & 89.59 $\pm$ 0.24                                           & 88.10 $\pm$ 0.24                                           & {\color[HTML]{329A9D} \textbf{91.67 $\pm$ 0.24}}           & 87.45 $\pm$ 0.29                                           & 85.26 $\pm$ 0.11                                 & 91.11 $\pm$ 0.40                                 & 90.74 $\pm$ 0.17 & {\color[HTML]{9A0000} \textbf{98.15 $\pm$ 0.48}} \\
                           & MOOC        & 65.23 $\pm$ 2.19                                           & 61.66 $\pm$ 0.95                                 & 75.95 $\pm$ 0.64                                           & 77.50 $\pm$ 2.91                                 & 73.51 $\pm$ 0.94                                           & 74.65 $\pm$ 0.54                                           & 74.27 $\pm$ 0.92                                 & 81.24 $\pm$ 0.69           & {\color[HTML]{329A9D} \textbf{83.01 $\pm$ 0.87}} & {\color[HTML]{9A0000} \textbf{97.49 $\pm$ 0.99}} \\
                           & LastFM      & 62.67 $\pm$ 4.49                                           & 64.41 $\pm$ 2.70                                 & 71.13 $\pm$ 0.17                                           & 65.95 $\pm$ 5.98                                           & 67.48 $\pm$ 0.77                                           & 58.21 $\pm$ 0.89                                           & 68.12 $\pm$ 0.33                                 & {\color[HTML]{329A9D} \textbf{73.97 $\pm$ 0.50}} & 72.19 $\pm$ 0.24 & {\color[HTML]{9A0000} \textbf{93.92 $\pm$ 1.10}} \\
                           & Enron       & 68.96 $\pm$ 0.98                                           & 67.79 $\pm$ 1.53                                 & 63.94 $\pm$ 1.36                                           & 70.89 $\pm$ 2.72                                           & 75.15 $\pm$ 0.58                                 & 71.29 $\pm$ 0.32                                           & 75.01 $\pm$ 0.79                                 & 77.41 $\pm$ 0.89           & {\color[HTML]{329A9D} \textbf{77.81 $\pm$ 0.65}}& {\color[HTML]{9A0000} \textbf{98.98 $\pm$ 0.53}} \\
                           & Social Evo. & 89.82 $\pm$ 4.11                                           & 93.28 $\pm$ 0.48                                 & 94.84 $\pm$ 0.44                                           & {\color[HTML]{329A9D} \textbf{95.13 $\pm$ 0.56}} & 88.32 $\pm$ 0.27                                           & 94.90 $\pm$ 0.36                                           & 94.72 $\pm$ 0.33                                 & {\color[HTML]{9A0000} \textbf{97.68 $\pm$ 0.10}}           & 97.57 $\pm$ 0.15 & 90.53 $\pm$ 1.88                                 \\
                           & UCI         & 65.99 $\pm$ 1.40                                           & 54.79 $\pm$ 1.76                                 & 68.67 $\pm$ 0.84                                           & 70.94 $\pm$ 0.71                                           & 64.61 $\pm$ 0.48                                           & 76.01 $\pm$ 1.11 & {\color[HTML]{329A9D} \textbf{80.10 $\pm$ 0.51}} & 72.25 $\pm$ 1.71                                           & {\color[HTML]{9A0000} \textbf{82.35 $\pm$ 0.73}} & 74.08 $\pm$ 1.84                                 \\
                           
                           & Flights     & 69.07 $\pm$ 4.02                                           & 70.57 $\pm$ 1.82                                 & {\color[HTML]{329A9D}\textbf{75.48 $\pm$ 0.26}}                                 & 71.09 $\pm$ 2.72                                           & 69.18 $\pm$ 1.52                                           & 74.62 $\pm$ 0.18                                           & 74.87 $\pm$ 0.21                                 & 70.92 $\pm$ 1.78                                          & 65.56 $\pm$ 1.10  & {\color[HTML]{9A0000}\textbf{80.75 $\pm$ 1.51}}                             \\
                           
                           & Can. Parl.  & 48.42 $\pm$ 0.66                                           & 58.61 $\pm$ 0.86                                 & 68.82 $\pm$ 1.21                                           & 65.34 $\pm$ 2.87                                           & 67.75 $\pm$ 1.00                                           & 65.85 $\pm$ 1.75                                           & 69.48 $\pm$ 0.63                       & {\color[HTML]{9A0000} \textbf{95.44 $\pm$ 0.57}}           & 59.83 $\pm$ 5.15 & {\color[HTML]{329A9D} \textbf{73.79 $\pm$ 1.26}} \\
                           
                           & US Legis.   & 50.27 $\pm$ 5.13                                           & {\color[HTML]{9A0000} \textbf{83.44 $\pm$ 1.16}} & 61.91 $\pm$ 5.82                                           & 67.57 $\pm$ 6.47                                           & 65.81 $\pm$ 8.52                                           & 78.15 $\pm$ 3.34                                           & 79.63 $\pm$ 0.84                                 & {\color[HTML]{329A9D} \textbf{81.25 $\pm$ 3.62}} & 51.64 $\pm$ 5.36 & 57.67 $\pm$ 3.10                                 \\
                           
                           & UN Trade    & 60.42 $\pm$ 1.48                                           & 60.19 $\pm$ 1.24                                 & 60.61 $\pm$ 1.24                                           & 61.04 $\pm$ 6.01                                           & {\color[HTML]{329A9D} \textbf{62.54 $\pm$ 0.67}} & 61.06 $\pm$ 1.74                                           & 60.15 $\pm$ 1.29                                 & 55.79 $\pm$ 1.02                                           & - & {\color[HTML]{9A0000} \textbf{80.96 $\pm$ 1.12}} \\
                           
                           & UN Vote     & {\color[HTML]{9A0000} \textbf{67.79 $\pm$ 1.46}}           & 67.53 $\pm$ 1.98                                 & 52.89 $\pm$ 1.61                                           & {\color[HTML]{329A9D} \textbf{67.63 $\pm$ 2.67}} & 52.19 $\pm$ 0.34                                           & 50.62 $\pm$ 0.82                                           & 51.60 $\pm$ 0.73                                 & 51.91 $\pm$ 0.84                                           & 51.14 $\pm$ 1.56 & 59.42 $\pm$ 3.73                                 \\
                           
                           & Contact     & 93.43 $\pm$ 1.78                                           & 94.18 $\pm$ 0.10                                 & 94.35 $\pm$ 0.48 & 90.18 $\pm$ 3.28                                           & 89.31 $\pm$ 0.27                                           & 91.35 $\pm$ 0.21    & 90.87 $\pm$ 0.35                                 & {\color[HTML]{329A9D} \textbf{94.75 $\pm$ 0.28}}           & {\color[HTML]{9A0000} \textbf{94.89 $\pm$ 0.89}} & 80.15 $\pm$ 1.48                                 \\ \cline{2-12} 
\multirow{-14}{*}{AP}      & Avg. Rank   


& 7.69& 7.23& 5.08& 5.38& 6.46& 5.69& 5.38& {\color[HTML]{329A9D}\textbf{3.92}} & 4.85 & {\color[HTML]{9A0000}\textbf{3.31}} \\

\hline

                           & Wikipedia   & 70.96 $\pm$ 0.78                                           & 67.36 $\pm$ 0.96                                 & 81.93 $\pm$ 0.22                                           & 80.97 $\pm$ 0.31                                           & 70.95 $\pm$ 0.95                                           & 82.19 $\pm$ 0.48                                 & {\color[HTML]{329A9D} \textbf{84.28 $\pm$ 0.30}} & 75.09 $\pm$ 3.70                                           &82.74 $\pm$ 0.32 & {\color[HTML]{9A0000} \textbf{98.78 $\pm$ 0.91}} \\
                           
                           & Reddit      & 83.51 $\pm$ 0.15                                           & 82.90 $\pm$ 0.31                                 & 87.13 $\pm$ 0.20                                 & 84.56 $\pm$ 0.24                                           & {\color[HTML]{329A9D} \textbf{88.04 $\pm$ 0.29}}           & 84.67 $\pm$ 0.29                                           & 82.21 $\pm$ 0.13                                 & 86.23 $\pm$ 0.51                                           & 84.38 $\pm$ 0.21 & {\color[HTML]{9A0000} \textbf{98.51 $\pm$ 0.40}} \\
                           & MOOC        & 66.63 $\pm$ 2.30                                           & 63.26 $\pm$ 1.01                                 & 73.18 $\pm$ 0.33                                           & 77.44 $\pm$ 2.86                                 & 70.32 $\pm$ 1.43                                           & 70.36 $\pm$ 0.37                                           & 72.45 $\pm$ 0.72                                 & {\color[HTML]{329A9D} \textbf{80.76 $\pm$ 0.76}}           & 78.47 $\pm$ 0.94 & {\color[HTML]{9A0000} \textbf{98.31 $\pm$ 0.59}} \\
                           & LastFM      & 61.32 $\pm$ 3.49                                           & 62.15 $\pm$ 2.12                                 & 63.99 $\pm$ 0.21                                           & 65.46 $\pm$ 4.27                                           & 67.92 $\pm$ 0.44                                           & 46.93 $\pm$ 2.59                                           & 60.22 $\pm$ 0.32                                 & 69.25 $\pm$ 0.36                                 & {\color[HTML]{329A9D} \textbf{72.30 $\pm$ 0.59}} & {\color[HTML]{9A0000} \textbf{97.08 $\pm$ 1.47}} \\
                           & Enron       & 70.92 $\pm$ 1.05                                           & 68.73 $\pm$ 1.34                                 & 60.45 $\pm$ 2.12                                           & 71.34 $\pm$ 2.46                                           & 75.17 $\pm$ 0.50           & 67.64 $\pm$ 0.86                                           & 71.53 $\pm$ 0.85                                 & 74.07 $\pm$ 0.64                                 & {\color[HTML]{329A9D} \textbf{77.27 $\pm$ 0.61}}  & {\color[HTML]{9A0000} \textbf{95.79 $\pm$ 0.33}} \\
                           & Social Evo. & 90.01 $\pm$ 3.19                                           & 93.07 $\pm$ 0.38                                 & 92.94 $\pm$ 0.61                                           & 95.24 $\pm$ 0.56 & 89.93 $\pm$ 0.15                                           & 93.44 $\pm$ 0.72                                           & 94.22 $\pm$ 0.32                                 & {\color[HTML]{329A9D} \textbf{97.51 $\pm$ 0.06}}           & {\color[HTML]{9A0000} \textbf{98.47 $\pm$ 0.02}} & 93.73 $\pm$ 2.22                                 \\
                           & UCI         & 64.14 $\pm$ 1.26                                           & 54.25 $\pm$ 2.01                                 & 60.80 $\pm$ 1.01                                           & 64.11 $\pm$ 1.04                                           & 58.06 $\pm$ 0.26                                           & 70.05 $\pm$ 1.86                                 & 74.59 $\pm$ 0.74 & 65.96 $\pm$ 1.18                                           & {\color[HTML]{329A9D} \textbf{75.39 $\pm$ 0.57}} & {\color[HTML]{9A0000} \textbf{84.09 $\pm$ 1.61}} \\
                           
                           & Flights     & 69.99 $\pm$ 3.10                                           & 71.13 $\pm$ 1.55                                 & {\color[HTML]{329A9D}\textbf{73.47 $\pm$ 0.18}}                                 & 71.63 $\pm$ 1.72                                           & 69.70 $\pm$ 0.75                                           & 72.54 $\pm$ 0.19                                           & 72.21 $\pm$ 0.21                                 & 69.53 $\pm$ 1.17                                           & 64.04 $\pm$ 0.55 & {\color[HTML]{9A0000}\textbf{92.32 $\pm$ 1.51}}                             \\
                           & Can. Parl.  & 52.88 $\pm$ 0.80                                           & 63.53 $\pm$ 0.65                                 & 72.47 $\pm$ 1.18                                           & 69.57 $\pm$ 2.81                                           & {\color[HTML]{329A9D} \textbf{72.93 $\pm$ 1.78}} & 69.47 $\pm$ 2.12                                           & 70.52 $\pm$ 0.94                                 & {\color[HTML]{9A0000} \textbf{96.70 $\pm$ 0.59}}           & 65.58 $\pm$ 5.71  & 85.80 $\pm$ 1.51                                 \\
                           
                           & US Legis.   & 59.05 $\pm$ 5.52                                           & {\color[HTML]{9A0000} \textbf{89.44 $\pm$ 0.71}} & 71.62 $\pm$ 5.42                                           & 78.12 $\pm$ 4.46                                           & 76.45 $\pm$ 7.02                                           & 82.54 $\pm$ 3.91                                           & 84.22 $\pm$ 0.91                                 & {\color[HTML]{329A9D} \textbf{87.96 $\pm$ 1.80}} & 60.71 $\pm$ 9.72 & 70.81 $\pm$ 3.72                                 \\
                           
                           & UN Trade    & 66.82 $\pm$ 1.27                                           & 65.60 $\pm$ 1.28                                 & 66.13 $\pm$ 0.78                                           & 66.37 $\pm$ 5.39                                           & {\color[HTML]{329A9D} \textbf{71.73 $\pm$ 0.74}} & 67.80 $\pm$ 1.21                                           & 66.53 $\pm$ 1.22                                 & 62.56 $\pm$ 1.51                                           & - & {\color[HTML]{9A0000} \textbf{87.92 $\pm$ 1.61}} \\
                           
                           & UN Vote     & {\color[HTML]{9A0000} \textbf{73.73 $\pm$ 1.61}}           & {\color[HTML]{329A9D} \textbf{72.80 $\pm$ 2.16}} & 53.04 $\pm$ 2.58                                           & 72.69 $\pm$ 3.72                                           & 52.75 $\pm$ 0.90                                           & 52.02 $\pm$ 1.64                                           & 51.89 $\pm$ 0.74                                 & 53.37 $\pm$ 1.26                               & 51.95 $\pm$ 2.14 & 70.70 $\pm$ 3.58                                 \\
                           
                           & Contact     & 94.47 $\pm$ 1.08 & 94.23 $\pm$ 0.18                                 & 94.10 $\pm$ 0.41                                           & 91.64 $\pm$ 1.72                                           & 87.68 $\pm$ 0.24                                           & 91.23 $\pm$ 0.19                                           & 90.96 $\pm$ 0.27                                 & {\color[HTML]{329A9D} \textbf{95.01 $\pm$ 0.15}}           & {\color[HTML]{9A0000} \textbf{95.41 $\pm$ 0.49}} & 90.68 $\pm$ 1.59                                 \\ \cline{2-12} 
\multirow{-14}{*}{ROC-AUC} & Avg. Rank   


& 6.92& 7.0& 5.85& 5.23& 6.23& 5.92& 5.69& {\color[HTML]{329A9D}\textbf{4.23}} & 5.15 & {\color[HTML]{9A0000}\textbf{2.77}} \\

\hline

\end{tabular}
}
}
\label{tab:transductive_ind}
\end{table}

\begin{table}[!htbp]
\centering
\caption{Performance comparison in the \textit{inductive setting} with \textit{inductive negative sampling strategy}.}
\vspace{1mm}
\resizebox{1.01\textwidth}{!}
{
\setlength{\tabcolsep}{0.9mm}
{
\begin{tabular}{c|c|cccccccccc}
\hline
Metric                 & Datasets    & JODIE                      & DyRep                                            & TGAT                                                       & TGN                                                        & CAWN                                             & TCL                                                        & GraphMixer                                                 & DyGFormer              & FreeDyG                                    & \name\                                           \\ \hline
                       & Wikipedia   & 68.70 $\pm$ 0.39           & 62.19 $\pm$ 1.28                                 & 84.17 $\pm$ 0.22                                 & 81.77 $\pm$ 0.32                                           & 67.24 $\pm$ 1.63                                 & 82.20 $\pm$ 2.18                                           & {\color[HTML]{329A9D} \textbf{87.60 $\pm$ 0.29}}           & 71.42 $\pm$ 4.43                                           & 87.54 $\pm$ 0.26 & {\color[HTML]{9A0000} \textbf{98.45 $\pm$ 0.88}} \\
                       
                       & Reddit      & 62.32 $\pm$ 0.54           & 61.58 $\pm$ 0.72                                 & 63.40 $\pm$ 0.36                                           & 64.84 $\pm$ 0.84                                 & 63.65 $\pm$ 0.41                                 & 60.81 $\pm$ 0.26                                           & 64.49 $\pm$ 0.25                                           & {\color[HTML]{329A9D} \textbf{65.35 $\pm$ 0.60}}           & 64.98 $\pm$ 0.20 & {\color[HTML]{9A0000} \textbf{85.76 $\pm$ 2.46}} \\
                       
                       & MOOC        & 63.22 $\pm$ 1.55           & 62.92 $\pm$ 1.24                                 & 76.72 $\pm$ 0.30                                           & 77.07 $\pm$ 3.40                                 & 74.69 $\pm$ 0.68                                 & 74.28 $\pm$ 0.53                                           & 73.99 $\pm$ 0.97                                           & 80.82 $\pm$ 0.30           & {\color[HTML]{329A9D} \textbf{81.41 $\pm$ 0.31}} & {\color[HTML]{9A0000} \textbf{95.17 $\pm$ 2.78}} \\
                       
                       & LastFM      & 70.39 $\pm$ 4.31           & 71.45 $\pm$ 1.75                                 & 76.28 $\pm$ 0.25                                           & 69.46 $\pm$ 4.65                                           & 71.33 $\pm$ 0.47                                 & 65.78 $\pm$ 0.65                                           & 76.42 $\pm$ 0.22           & 76.35 $\pm$ 0.52                                   & {\color[HTML]{329A9D} \textbf{77.01 $\pm$ 0.43}} &  {\color[HTML]{9A0000} \textbf{92.96 $\pm$ 1.08}} \\
                       
                       & Enron       & 65.86 $\pm$ 3.71           & 62.08 $\pm$ 2.27                                 & 61.40 $\pm$ 1.30                                           & 62.90 $\pm$ 1.16                                           & 60.72 $\pm$ 0.36                                 & 67.11 $\pm$ 0.62                                 & 72.37 $\pm$ 1.38          & 67.07 $\pm$ 0.62                                     & {\color[HTML]{329A9D} \textbf{72.85 $\pm$ 0.81}} & {\color[HTML]{9A0000} \textbf{89.36 $\pm$ 1.01}} \\
                       
                       & Social Evo. & 88.51 $\pm$ 0.87           & 88.72 $\pm$ 1.10                                 & 93.97 $\pm$ 0.54                                           & 90.65 $\pm$ 1.62                                           & 79.83 $\pm$ 0.39                                 & 94.10 $\pm$ 0.32 & 94.01 $\pm$ 0.47                                           & {\color[HTML]{329A9D} \textbf{96.82 $\pm$ 0.17}}           & {\color[HTML]{9A0000} \textbf{96.91 $\pm$ 0.12}} & 85.13 $\pm$ 1.05                                 \\
                       & UCI         & 63.16 $\pm$ 2.27           & 52.47 $\pm$ 2.09                                 & 70.49 $\pm$ 0.93                                           & 70.73 $\pm$ 0.79                                           & 64.54 $\pm$ 0.47                                 & 76.65 $\pm$ 0.99 & {\color[HTML]{329A9D} \textbf{81.64 $\pm$ 0.49}}           & 72.13 $\pm$ 1.86                                           & {\color[HTML]{9A0000} \textbf{82.06 $\pm$ 0.58}} & 75.97 $\pm$ 1.13                                 \\
                       
                       & Flights     & 61.01 $\pm$ 1.66           & 62.83 $\pm$ 1.31                                 & 64.72 $\pm$ 0.37 & 59.32 $\pm$ 1.45                                           & 56.82 $\pm$ 0.56                                 & 64.50 $\pm$ 0.25                                           & {\color[HTML]{329A9D} \textbf{65.29 $\pm$ 0.24}}           & 57.11 $\pm$ 0.20                                          & 56.70 $\pm$ 1.17  & {\color[HTML]{9A0000} \textbf{69.14 $\pm$ 1.19}}                             \\ 
                       
                       & Can. Parl.  & 52.58 $\pm$ 0.86           & 52.24 $\pm$ 0.28                                 & 56.46 $\pm$ 0.50                                           & 54.18 $\pm$ 0.73                                           & 57.06 $\pm$ 0.08                       & 55.46 $\pm$ 0.69                                           & 55.76 $\pm$ 0.65                                           & {\color[HTML]{9A0000} \textbf{87.22 $\pm$ 0.82}}           & 52.33 $\pm$ 0.87 & {\color[HTML]{329A9D} \textbf{69.59 $\pm$ 1.45}} \\
                       
                       & US Legis.   & 52.94 $\pm$ 2.11           & {\color[HTML]{9A0000} \textbf{62.10 $\pm$ 1.41}} & 51.83 $\pm$ 3.95                                           & {\color[HTML]{329A9D} \textbf{61.18 $\pm$ 1.10}} & 55.56 $\pm$ 1.71                                 & 53.87 $\pm$ 1.41                                           & 52.03 $\pm$ 1.02                                           & 56.31 $\pm$ 3.46                                     & 53.63 $\pm$ 4.38 &  49.51 $\pm$ 1.09                                 \\
                       
                       & UN Trade    & 55.43 $\pm$ 1.20           & 55.42 $\pm$ 0.87                                 & 55.58 $\pm$ 0.68                                 & 52.80 $\pm$ 3.24                                           & 54.97 $\pm$ 0.38                                 & {\color[HTML]{329A9D} \textbf{55.66 $\pm$ 0.98}}           & 54.88 $\pm$ 1.01                                           & 52.56 $\pm$ 1.70                                           & - & {\color[HTML]{9A0000} \textbf{79.12 $\pm$ 1.84}} \\
                       
                       & UN Vote     & 61.17 $\pm$ 1.33 & 60.29 $\pm$ 1.79                                 & 53.08 $\pm$ 3.10                                           & {\color[HTML]{9A0000} \textbf{63.71 $\pm$ 2.97}}           & 48.01 $\pm$ 0.82                                 & 54.13 $\pm$ 2.16                                           & 48.10 $\pm$ 0.40                                           & 52.61 $\pm$ 1.25                                     & 50.21 $\pm$ 1.74  & {\color[HTML]{329A9D} \textbf{61.88 $\pm$ 1.27}} \\
                       
                       & Contact     & 90.43 $\pm$ 2.33           & 89.22 $\pm$ 0.65                                 & {\color[HTML]{9A0000} \textbf{94.14 $\pm$ 0.45}}           & 88.12 $\pm$ 1.50                                           & 74.19 $\pm$ 0.81                                 & 90.43 $\pm$ 0.17                                           
                       & 89.91 $\pm$ 0.36                                           & {\color[HTML]{329A9D} \textbf{93.55 $\pm$ 0.52}} & 93.26 $\pm$ 1.41 &  79.09 $\pm$ 1.25                                 \\ \cline{2-12} 
\multirow{-14}{*}{ind} & Avg. Rank   

& 6.85& 7.08& 5.23& 5.77& 7.54& 5.23& 4.92& {\color[HTML]{329A9D}\textbf{4.46}}& 4.62& {\color[HTML]{9A0000}\textbf{3.31}} \\

\hline

                       & Wikipedia   & 61.87 $\pm$ 0.53           & 57.54 $\pm$ 1.09                                 & 78.38 $\pm$ 0.20                                           & 75.76 $\pm$ 0.29                                           & 62.02 $\pm$ 0.65                                 & 79.79 $\pm$ 0.96                                 & {\color[HTML]{329A9D} \textbf{82.88 $\pm$ 0.21}}           & 68.33 $\pm$ 2.82                                          & 83.17 $\pm$ 0.31 & {\color[HTML]{9A0000} \textbf{98.59 $\pm$ 0.82}} \\
                       
                       & Reddit      & 61.69 $\pm$ 0.39           & 60.44 $\pm$ 0.37                                 & 64.39 $\pm$ 0.27                                           & 64.55 $\pm$ 0.50                                           & {\color[HTML]{329A9D} \textbf{64.91 $\pm$ 0.21}} & 61.36 $\pm$ 0.26                                           & 64.27 $\pm$ 0.13                                           & 64.80 $\pm$ 0.25                                 & 64.51 $\pm$ 0.19 & {\color[HTML]{9A0000} \textbf{92.05 $\pm$ 1.87}} \\
                       
                       & MOOC        & 64.48 $\pm$ 1.64           & 64.22 $\pm$ 1.29                                 & 74.07 $\pm$ 0.27                                           & 77.68 $\pm$ 3.55                                 & 71.69 $\pm$ 0.94                                 & 69.83 $\pm$ 0.32                                           & 72.52 $\pm$ 0.84                                           & {\color[HTML]{329A9D} \textbf{80.77 $\pm$ 0.63}}           & 75.81 $\pm$ 0.69 & {\color[HTML]{9A0000} \textbf{96.85 $\pm$ 1.59}} \\
                       
                       & LastFM      & 68.44 $\pm$ 3.26           & 68.79 $\pm$ 1.08                                 & 69.89 $\pm$ 0.28                                           & 66.99 $\pm$ 5.61                                           & 67.68 $\pm$ 0.24                                 & 55.88 $\pm$ 1.85                                           & 70.07 $\pm$ 0.20                                 & 70.73 $\pm$ 0.37           & {\color[HTML]{329A9D} \textbf{71.42 $\pm$ 0.33}} & {\color[HTML]{9A0000} \textbf{96.34 $\pm$ 1.68}} \\
                       
                       & Enron       & 65.32 $\pm$ 3.57           & 61.50 $\pm$ 2.50                                 & 57.83 $\pm$ 2.18                                           & 62.68 $\pm$ 1.09                                           & 62.27 $\pm$ 0.40                                 & 64.05 $\pm$ 1.02                                           & 68.19 $\pm$ 1.63          & 65.79 $\pm$ 0.42                                & {\color[HTML]{329A9D} \textbf{68.79 $\pm$ 0.91}} & {\color[HTML]{9A0000} \textbf{92.14 $\pm$ 0.70}} \\
                       
                       & Social Evo. & 88.53 $\pm$ 0.55           & 87.93 $\pm$ 1.05                                 & 91.88 $\pm$ 0.72                                           & 92.10 $\pm$ 1.22                                           & 83.54 $\pm$ 0.24                                 & 93.28 $\pm$ 0.60                                           & 93.62 $\pm$ 0.35 & {\color[HTML]{9A0000} \textbf{96.91 $\pm$ 0.09}}           & {\color[HTML]{329A9D} \textbf{96.79 $\pm$ 0.17}} & 90.08 $\pm$ 3.91                                 \\
                       
                       & UCI         & 60.27 $\pm$ 1.94           & 51.26 $\pm$ 2.40                                 & 62.29 $\pm$ 1.17                                           & 62.66 $\pm$ 0.91                                           & 56.39 $\pm$ 0.11                                 & 70.42 $\pm$ 1.93                                 & {\color[HTML]{329A9D} \textbf{75.97 $\pm$ 0.85}}           & 65.58 $\pm$ 1.00                                    & 73.41 $\pm$ 0.88 & {\color[HTML]{9A0000} \textbf{84.63 $\pm$ 1.22}} \\
                       
                       & Flights     & 60.72 $\pm$ 1.29           & 61.99 $\pm$ 1.39                                 & 63.40 $\pm$ 0.26                                 & 59.66 $\pm$ 1.05                                           & 56.58 $\pm$ 0.44                                 & {\color[HTML]{329A9D}\textbf{63.49 $\pm$ 0.23}}                                  & 63.32 $\pm$ 0.19                                           & 56.05 $\pm$ 0.22                           & 55.18 $\pm$ 1.04 & {\color[HTML]{9A0000}\textbf{83.87 $\pm$ 1.30}}                             \\
                       
                       & Can. Parl.  & 51.61 $\pm$ 0.98           & 52.35 $\pm$ 0.52                                 & 58.15 $\pm$ 0.62                                           & 55.43 $\pm$ 0.42                                           & 60.01 $\pm$ 0.47                       & 56.88 $\pm$ 0.93                                           & 56.63 $\pm$ 1.09                                           & {\color[HTML]{9A0000} \textbf{88.51 $\pm$ 0.73}}           & 51.37 $\pm$ 2.23 & {\color[HTML]{329A9D} \textbf{81.26 $\pm$ 3.92}} \\
                       
                       & US Legis.   & 58.12 $\pm$ 2.94           & {\color[HTML]{9A0000} \textbf{67.94 $\pm$ 0.98}} & 49.99 $\pm$ 4.88                                           & {\color[HTML]{329A9D} \textbf{64.87 $\pm$ 1.65}} & 54.41 $\pm$ 1.31                                 & 52.12 $\pm$ 2.13                                           & 49.28 $\pm$ 0.86                                           & 56.57 $\pm$ 3.22                                     & 53.75 $\pm$ 5.92 & 55.74 $\pm$ 1.53                                 \\
                       
                       & UN Trade    & 58.71 $\pm$ 1.20           & 57.87 $\pm$ 1.36                                 & 59.98 $\pm$ 0.59                                           & 55.62 $\pm$ 3.59                                           & 60.88 $\pm$ 0.79                       & {\color[HTML]{329A9D} \textbf{61.01 $\pm$ 0.93}}           & 59.71 $\pm$ 1.17                                           & 57.28 $\pm$ 3.06                                           & - & {\color[HTML]{9A0000} \textbf{86.28 $\pm$ 1.76}} \\
                       
                       & UN Vote     & 65.29 $\pm$ 1.30 & 64.10 $\pm$ 2.10                                 & 51.78 $\pm$ 4.14                                           & {\color[HTML]{329A9D} \textbf{68.58 $\pm$ 3.08}}           & 48.04 $\pm$ 1.76                                 & 54.65 $\pm$ 2.20                                           & 45.57 $\pm$ 0.41                                           & 53.87 $\pm$ 2.01                                           & 48.93 $\pm$ 2.53 & {\color[HTML]{9A0000} \textbf{72.98 $\pm$ 1.22}} \\
                       
                       & Contact     & 90.80 $\pm$ 1.18           & 88.87 $\pm$ 0.67                                 & 93.76 $\pm$ 0.40 & 88.85 $\pm$ 1.39                                           & 74.79 $\pm$ 0.38                                 & 90.37 $\pm$ 0.16                                           & 90.04 $\pm$ 0.29                                           & {\color[HTML]{329A9D} \textbf{94.14 $\pm$ 0.26}}                 & {\color[HTML]{9A0000} \textbf{94.27 $\pm$ 0.76}} &  88.69 $\pm$ 3.25                                 \\ \cline{2-12} 
\multirow{-14}{*}{ind} & Avg. Rank   


& 6.54& 7.38& 5.69& 5.77& 7.0& 5.54& 5.31& {\color[HTML]{329A9D}\textbf{4.23}}& 5.08& {\color[HTML]{9A0000}\textbf{2.46}} \\

\hline

\end{tabular}
}
}
\label{tab:inductive_ind}
\end{table}

\clearpage

\subsection{Demonstrating the MLPs and Transformers on the proposed LPE positional encoding}
\label{sec:exp_ablation_study}
Here, we empirically show that simple architectures like MLPs can fully harness our learnable positional encoding to finish temporal link prediction tasks.

By performing an ablation study of using Transformers (instead of MLPs) to help \name\ learn positional encodings, as shown in Table~\ref{tab:transformer}, we can discern (1) in most cases (i.e., 9 of 13 datasets), our vanilla \name\ can outperform, and the gain is not marginal; (2) in a few cases, adding the attention mechanism on \name\ to replace MLPs can boost the performance to a small extent, and it is prone to cost more computational memory, like OOM in the Flights dataset.

\begin{table*}[h]
\centering
\caption{Comparison of replacing MLPs in \name\ with Transformers.}
\vspace{1mm}
\resizebox{0.8\textwidth}{!}
{
\setlength{\tabcolsep}{0.7mm}
{
\begin{tabular}{c|c|c|c|c|c}
\hline
Dataset & Method & Transductive AP &	Transductive ROC-AUC &	Inductive AP & 	Inductive ROC-AUC \\
\hline
\multirow{2}{*}{Wikipedia}   & \name\ + Attention &	98.50 $\pm$ 0.73 &	98.85 $\pm$ 0.50 &	98.11 $\pm$ 0.83 &	98.49 $\pm$ 0.62 \\
            & \name\ &	\textbf{99.34 $\pm$ 0.04} &	\textbf{99.48 $\pm$ 0.03} &	\textbf{99.15 $\pm$ 0.04} &	\textbf{99.30 $\pm$ 0.03} \\
\hline
\multirow{2}{*}{Reddit}      & \name\ + Attention &	98.90 $\pm$ 0.10 &	99.13 $\pm$ 0.08 &	97.75 $\pm$ 0.26 &	98.29 $\pm$ 0.16 \\
            & \name\ &	\textbf{99.37 $\pm$ 0.04} &	\textbf{99.49 $\pm$ 0.03} &	\textbf{98.02 $\pm$ 0.09} &	\textbf{98.49 $\pm$ 0.06} \\
\hline
\multirow{2}{*}{MOOC}        & \name\ + Attention	& 82.46 $\pm$ 3.61	& 85.38 $\pm$ 2.93 &	82.57 $\pm$ 4.24 &	85.80 $\pm$ 3.18 \\
            & \name\ &	\textbf{86.94 $\pm$ 0.34} &	\textbf{89.28 $\pm$ 0.28} &	\textbf{88.49 $\pm$ 0.24} & 	\textbf{90.63 $\pm$ 0.21} \\
\hline
\multirow{2}{*}{LastFM}      & \name\ + Attention &	95.50 $\pm$ 0.70 &	97.19 $\pm$ 0.50 &	96.07 $\pm$ 0.36 &	97.62 $\pm$ 0.25 \\
            & \name\ &	\textbf{96.06 $\pm$ 0.30} &	\textbf{97.52 $\pm$ 0.17} &	\textbf{96.25 $\pm$ 0.43} &	\textbf{97.66 $\pm$ 0.21} \\
\hline
\multirow{2}{*}{Enron}       & \name\ + Attention &	92.56 $\pm$ 0.53 &	95.02 $\pm$ 0.33 &	 86.41 $\pm$ 1.15 &	90.70 $\pm$ 0.72\\
            & \name\ &	\textbf{93.96 $\pm$ 0.36} &	\textbf{95.98 $\pm$ 0.23} &	\textbf{89.17 $\pm$ 0.88} &	\textbf{92.30 $\pm$ 0.57} \\
\hline
\multirow{2}{*}{SocialEvo}   & \name\ + Attention &	90.26 $\pm$ 0.69 &	92.61 $\pm$ 0.58 &	90.75 $\pm$ 0.31 &	93.50 $\pm$ 0.30 \\
            & \name\ & 	\textbf{92.22 $\pm$ 0.25} & 	\textbf{94.33 $\pm$ 0.21} & 	\textbf{91.71 $\pm$ 0.32} & 	\textbf{94.09 $\pm$ 0.24} \\
\hline
\multirow{2}{*}{UCI}         &	\name\ + Attention & 	96.06 $\pm$ 0.53 &	97.12 $\pm$ 0.41 &	93.00 $\pm$ 1.30 &	94.59 $\pm$ 0.85 \\
            & \name\	& \textbf{96.67 $\pm$ 0.47} &	\textbf{97.62 $\pm$ 0.23} &	\textbf{94.60 $\pm$ 0.41} &	\textbf{95.88 $\pm$ 0.17} \\
\hline
\multirow{2}{*}{Flights}     &	\name\ + Attention &	OOM &	OOM &	OOM &	OOM \\
            & \name\ &	98.94 $\pm$ 0.10 &	99.38 $\pm$ 0.04 &	97.43 $\pm$ 0.15 &	98.46 $\pm$ 0.07 \\
\hline
\multirow{2}{*}{Can. Parl.}  &	\name\ + Attention &	\textbf{98.36 $\pm$ 0.12} & \textbf{99.01 $\pm$ 0.07} &	\textbf{92.59 $\pm$ 0.32} &	\textbf{95.37 $\pm$ 0.25} \\
            & \name\ &	98.24 $\pm$ 0.11 &	98.97 $\pm$ 0.06 &	92.25$\pm$ 0.24 &	95.06 $\pm$ 0.11 \\
\hline
\multirow{2}{*}{US Legis.}   &	\name\ + Attention &	\textbf{76.84 $\pm$ 0.61} &	\textbf{84.16 $\pm$ 0.40} &	\textbf{64.07 $\pm$ 1.31} &	\textbf{68.13 $\pm$ 1.78} \\
            & \name\	 & 76.74 $\pm$ 0.60 & 	83.89 $\pm$ 0.43 &	61.98 $\pm$ 1.31 &	66.18 $\pm$ 1.19 \\
\hline 
\multirow{2}{*}{UN Trade.}   &	\name\ + Attention &	\textbf{77.45 $\pm$ 3.27} &	\textbf{83.94 $\pm$ 2.23} &	\textbf{79.83 $\pm$ 3.28} &	\textbf{85.95 $\pm$ 2.13} \\
            & \name\ &	75.84 $\pm$ 2.08 &	83.17 $\pm$ 1.28 &	76.80 $\pm$ 2.01 &	83.95 $\pm$ 1.76 \\
\hline 
\multirow{2}{*}{UN Vote}     &	\name\ + Attention &	\textbf{74.65 $\pm$ 0.46} &	\textbf{80.73 $\pm$ 0.40} &	\textbf{76.10 $\pm$ 0.88} &	\textbf{82.89 $\pm$ 0.64} \\
            & \name\	& 73.40 $\pm$ 0.03 &	79.59 $\pm$ 0.02 &	75.09 $\pm$ 0.34 &	82.01 $\pm$ 0.21 \\
\hline 
\multirow{2}{*}{Contact}     &	\name\ + Attention &	97.32 $\pm$ 0.15 &	98.12 $\pm$ 0.10 &	96.05 $\pm$ 0.41 &	97.18 $\pm$ 0.21 \\
            & \name\	& \textbf{98.16 $\pm$ 0.09}	& \textbf{98.72 $\pm$ 0.06} &	\textbf{97.25 $\pm$ 0.07} &	\textbf{97.99 $\pm$ 0.07} \\
\hline
\end{tabular}
}
}
\label{tab:transformer}
\end{table*}

Based on the above analysis, our design gets verified in-depth: the proposed learnable positional encoding is effective and friendly to lightweight neural architectures, and even simple MLPs can fully exploit its expressive power.


\subsection{Empirical runtime comparison with SOTAs}
\label{app:scalability}
\begin{table}[!htbp]
\centering
\caption{Comparison in runtime (in seconds) and number of training epochs of \name\ and DyGFormer across 5 runs on US Legis and UN Trade.}
\vspace{1mm}
\resizebox{1.01\textwidth}{!}
{
\setlength{\tabcolsep}{0.9mm}
{
\begin{tabular}{c|c|c|c}
\hline
Dataset                     & Method        & Total time for 5 runs (in seconds)               & Number of training epochs  \\ 
\hline
\multirow{3}{*}{US Legis}   & \name\     & \textbf{[287.19, 257.18, 251.49, 248.92 , 305.09]}     &	\textbf{[21, 18, 17, 17, 21]} \\
                            & DyGFormer     & [1251.97, 1697.56, 1112.50, 756.80, 1092.77] &	[27, 37, 24, 16, 23] \\ 
                            & FreeDyG & [1019.91, 871.21, 1061.15, 1070.41, 868.03] & [17, 13, 18, 18, 13] \\
\hline
\multirow{2}{*}{UN Trade}   & \name\     & \textbf{[2922.14 ,2857.13 , 2450.84 , 2593.40, 2740.26]}         &	\textbf{[21, 21, 18, 18, 19]} \\
                            & DyGFormer     & [12847.83, 21167.01, 20380.42, 9901.51, 39374.45]    &	[33, 55, 53, 25, 95] \\ 
                            & FreeDyG & - & - \\
\hline
\end{tabular}
}
}
\label{tab:runtime_comparison}
\end{table}

We present the runtime comparison between \name~and DyGFormer~\citep{DBLP:conf/nips/0004S0L23}, FreeDyG~\citep{DBLP:conf/iclr/TianQG24}, recent SOTA methods, on \textit{US Legis} and \textit{UN Trade} datasets, in Table~\ref{tab:runtime_comparison}. As shown in Table~\ref{tab:runtime_comparison}, our running time is lower than SOTAs in terms of convergence time.




\subsection{Empirical results for ablation study}
\label{app:ablation_study}

Next, we conduct an ablation study to examine the ability of the learnable positional encoding in the link prediction task.

Specifically, we consider the following cases: (1) only leveraging LPE to make predictions (i.e., we omit node/edge features and only employ positional encoding learned by LPE to obtain link predictions); (2) omitting LPE module (i.e., only relying on node and link features).

First, we provide the performance comparison between \name\ and only using LPE module (denoted as ``Only LPE'') to obtain link predictions under transductive and inductive settings with random negative sampling strategy. As shown in Table~\ref{tab:ablation_study_1}, on most datasets, the performance obtained by only leveraging LPE is already competitive, which suggests the positional encoding is informative enough to support downstream tasks. Also, we can observe that LPE positional encodings can collaborate well with node and edge features to boost performance.

\begin{table}[!htbp]
\centering
\caption{Comparison between only LPE and \name\ with random negative sampling strategy.}
\vspace{1mm}
\resizebox{1.01\textwidth}{!}
{
\setlength{\tabcolsep}{0.9mm}
{
\begin{tabular}{c|cccc|cccc}
\hline
                          & \multicolumn{4}{c|}{Transductive setting}                                                                                                    & \multicolumn{4}{c}{Inductive setting}                                                                                                                               \\ \cline{2-9} 
                          & \multicolumn{2}{c|}{AP}                                                    & \multicolumn{2}{c|}{ROC-AUC}                                    & \multicolumn{2}{c|}{AP}                                                                            & \multicolumn{2}{c}{ROC-AUC}                                    \\ \cline{2-9} 
\multirow{-3}{*}{Dataset} & \name               & \multicolumn{1}{c|}{Only LPE}                  & {\color[HTML]{1F1F1F} \name } & Only LPE                  & {\color[HTML]{1F1F1F} \name } & \multicolumn{1}{c|}{Only LPE}                                & {\color[HTML]{1F1F1F} \name } & Only LPE                 \\ \hline
Wikipedia                 & \textbf{99.34 $\pm$ 0.04} & \multicolumn{1}{c|}{99.07 $\pm$ 0.03}          & \textbf{99.48 $\pm$ 0.03}           & 99.42 $\pm$ 0.03          & \textbf{99.15 $\pm$ 0.04}           & \multicolumn{1}{c|}{98.77 $\pm$ 0.05}                        & \textbf{99.30 $\pm$ 0.03}           & 98.99$\pm$ 0.03          \\
Reddit                    & \textbf{99.37 $\pm$ 0.04} & \multicolumn{1}{c|}{98.67 $\pm$ 0.07}          & \textbf{99.49 $\pm$ 0.03}           & 99.13 $\pm$ 0.05          & \textbf{98.02 $\pm$ 0.09}           & \multicolumn{1}{c|}{{\color[HTML]{1F1F1F} 95.44 $\pm$ 0.29}} & \textbf{98.49 $\pm$ 0.06}           & 96.71$\pm$0.16           \\
MOOC                      & \textbf{86.94 $\pm$ 0.34} & \multicolumn{1}{c|}{81.55 $\pm$ 0.53}          & \textbf{89.28 $\pm$ 0.28}           & 83.04 $\pm$ 0.44          & \textbf{88.49 $\pm$ 0.24}           & \multicolumn{1}{c|}{80.04 $\pm$ 0.40}                        & \textbf{90.63 $\pm$ 0.21}           & 81.70$\pm$ 0.39          \\
Lastfm                    & \textbf{96.06 $\pm$ 0.30} & \multicolumn{1}{c|}{92.84 $\pm$ 0.54}          & \textbf{97.52 $\pm$ 0.17}           & 94.43 $\pm$ 0.38          & \textbf{96.25 $\pm$ 0.43}           & \multicolumn{1}{c|}{94.27 $\pm$ 0.36}                        & \textbf{97.66 $\pm$ 0.21}           & 96.22$\pm$ 0.22          \\
Enron                     & \textbf{93.96 $\pm$ 0.36} & \multicolumn{1}{c|}{93.82 $\pm$ 0.44}          & \textbf{95.98 $\pm$ 0.23}           & 95.95 $\pm$ 0.31          & \textbf{89.17 $\pm$ 0.88}           & \multicolumn{1}{c|}{89.15 $\pm$1.32}                         & 92.30 $\pm$ 0.57                    & \textbf{92.33 $\pm$0.79} \\
Social Evo.               & \textbf{92.22 $\pm$ 0.25} & \multicolumn{1}{c|}{91.56 $\pm$ 0.79}          & \textbf{94.33 $\pm$ 0.21}           & 93.76 $\pm$ 0.63          & \textbf{91.71 $\pm$ 0.32}           & \multicolumn{1}{c|}{90.42$\pm$ 1.68}                         & \textbf{94.09 $\pm$ 0.24}           & 93.28$\pm$ 0.97          \\
UCI                       & 96.67 $\pm$ 0.47          & \multicolumn{1}{c|}{\textbf{96.83 $\pm$ 0.19}} & 97.62 $\pm$ 0.23                    & \textbf{97.89 $\pm$ 0.10} & \textbf{94.60 $\pm$ 0.41}           & \multicolumn{1}{c|}{94.60 $\pm$ 0.46}                        & 95.88 $\pm$ 0.17                    & \textbf{96.17 $\pm$0.22} \\
Flights                  & 98.94 $\pm$ 0.10          & \multicolumn{1}{c|}{\textbf{99.10 $\pm$ 0.03}} & 99.38 $\pm$ 0.04          & \textbf{99.43 $\pm$ 0.06} & 97.43 $\pm$ 0.15          & \multicolumn{1}{c|}{\textbf{97.81 $\pm$0.25}} & 98.46 $\pm$ 0.07          & \textbf{98.62 $\pm$0.11} \\
Can. Parl.               & 98.24 $\pm$ 0.11          & \multicolumn{1}{c|}{\textbf{98.41 $\pm$ 0.06}} & 98.97 $\pm$ 0.06          & \textbf{99.03 $\pm$ 0.03} & \textbf{92.25$\pm$ 0.24}  & \multicolumn{1}{c|}{92.13 $\pm$ 0.24}         & 95.06 $\pm$ 0.11          & \textbf{95.09 $\pm$0.07} \\
US Legis.                & 76.74 $\pm$ 0.60          & \multicolumn{1}{c|}{\textbf{77.36 $\pm$ 0.26}} & 83.89 $\pm$ 0.43          & \textbf{84.36 $\pm$ 0.20} & 61.98 $\pm$ 1.31          & \multicolumn{1}{c|}{\textbf{63.47 $\pm$0.63}} & 66.18 $\pm$ 1.19          & \textbf{66.75$\pm$ 0.57} \\
UN Trade                 & 75.84 $\pm$ 2.08          & \multicolumn{1}{c|}{\textbf{78.46 $\pm$ 0.97}} & \textbf{83.17 $\pm$ 1.28} & 77.81 $\pm$ 1.78          & 76.80 $\pm$ 2.01          & \multicolumn{1}{c|}{\textbf{84.85$\pm$ 0.89}} & 83.95 $\pm$ 1.76          & \textbf{84.89$\pm$ 1.39} \\
UN Vote                  & \textbf{73.40 $\pm$ 0.03} & \multicolumn{1}{c|}{70.99 $\pm$ 0.11}          & \textbf{79.59 $\pm$ 0.02} & 76.67 $\pm$ 0.09          & \textbf{75.09 $\pm$ 0.34} & \multicolumn{1}{c|}{70.01$\pm$ 0.69}          & \textbf{82.01 $\pm$ 0.21} & 77.36 $\pm$0.51          \\
Contact                  & \textbf{98.16 $\pm$ 0.09} & \multicolumn{1}{c|}{97.73 $\pm$ 0.04}          & \textbf{98.72 $\pm$ 0.06} & 98.40 $\pm$ 0.03          & \textbf{97.25 $\pm$ 0.07} & \multicolumn{1}{c|}{96.38$\pm$ 0.09}          & \textbf{97.99 $\pm$ 0.07} & 97.37$\pm$ 0.07          \\ \hline

\end{tabular}
}
}
\label{tab:ablation_study_1}
\end{table}

Second, we present the comparison between \name\ and only using node/edge features, i.e., omitting the LPE module from \name\ (denoted as ``W/o LPE''). As shown in Table~\ref{tab:ablation_study_no_pe}, \name\ performs the best on all datasets, further suggesting that performance would be negatively affected without employing LPE. 

\begin{table}[!htbp]
\centering
\caption{Comparison between not using LPE and \name\ with random negative sampling strategy.}
\vspace{1mm}
\resizebox{1.01\textwidth}{!}
{
\setlength{\tabcolsep}{0.9mm}
{
\begin{tabular}{c|cccc|cccc}
\hline
& \multicolumn{4}{c|}{Transductive setting}                                         &\multicolumn{4}{c}{Inductive setting}                              \\ \cline{2-9} 

& \multicolumn{2}{c|}{AP} & \multicolumn{2}{c|}{ROC-AUC} & \multicolumn{2}{c|}{AP}  & \multicolumn{2}{c}{ROC-AUC}                                       \\ \cline{2-9} 

\multirow{-3}{*}{Dataset} & \name & \multicolumn{1}{c|}{W/o LPE} & \name & W/o LPE & \name & \multicolumn{1}{c|}{W/o LPE} & \name & W/o LPE  \\ \hline

Wikipedia & \textbf{99.34 $\pm$ 0.04} & \multicolumn{1}{c|}{96.52 $\pm$ 0.06} & \textbf{99.48 $\pm$ 0.03} & 96.15 $\pm$ 0.08 & \textbf{99.15 $\pm$ 0.04} & \multicolumn{1}{c|}{96.11 $\pm$ 0.09} & \textbf{99.30 $\pm$ 0.03} & 95.64 $\pm$ 0.11 \\

Reddit & \textbf{99.37 $\pm$ 0.04} & \multicolumn{1}{c|}{97.07 $\pm$ 0.08} & \textbf{99.49 $\pm$ 0.03} & 96.96 $\pm$ 0.07 & \textbf{98.02 $\pm$ 0.09} & \multicolumn{1}{c|}{95.10 $\pm$ 0.06} & \textbf{98.49 $\pm$ 0.06} & 95.02 $\pm$ 0.06 \\

MOOC & \textbf{86.94 $\pm$ 0.34} & \multicolumn{1}{c|}{81.89 $\pm$ 0.32} & \textbf{89.28 $\pm$ 0.28} & 83.24 $\pm$ 0.29 & \textbf{88.49 $\pm$ 0.24} & \multicolumn{1}{c|}{80.65 $\pm$ 0.28} & \textbf{90.63 $\pm$ 0.21} & 82.19 $\pm$ 0.26 \\

LastFM & \textbf{96.06 $\pm$ 0.30} & \multicolumn{1}{c|}{74.63 $\pm$ 0.41} & \textbf{97.52 $\pm$ 0.17} & 73.48 $\pm$ 0.40 & \textbf{96.25 $\pm$ 0.43} & \multicolumn{1}{c|}{82.08 $\pm$ 0.46} & \textbf{97.66 $\pm$ 0.21} & 80.74 $\pm$ 0.37 \\

Enron & \textbf{93.96 $\pm$ 0.36} & \multicolumn{1}{c|}{82.49 $\pm$ 0.85} & \textbf{95.98 $\pm$ 0.23} & 84.18 $\pm$ 0.53 & \textbf{89.17 $\pm$ 0.88} & \multicolumn{1}{c|}{76.14 $\pm$ 0.82} & \textbf{92.30 $\pm$ 0.57} & 76.63 $\pm$ 0.65 \\

Social Evo. & \textbf{92.22 $\pm$ 0.25} & \multicolumn{1}{c|}{91.17 $\pm$ 0.21} & \textbf{94.33 $\pm$ 0.21} & 93.30 $\pm$ 0.17 & \textbf{91.71 $\pm$ 0.32} & \multicolumn{1}{c|}{89.38 $\pm$ 0.30} & \textbf{94.09 $\pm$ 0.24} & 91.85 $\pm$ 0.24 \\

UCI & \textbf{96.67 $\pm$ 0.47} & \multicolumn{1}{c|}{91.79 $\pm$ 4.08} & \textbf{97.62 $\pm$ 0.23} & 90.51 $\pm$ 3.45 & \textbf{94.60 $\pm$ 0.41} & \multicolumn{1}{c|}{89.58 $\pm$ 3.41} & \textbf{95.88 $\pm$ 0.17} & 87.88 $\pm$ 3.10 \\

Flights & \textbf{98.94 $\pm$ 0.10} & \multicolumn{1}{c|}{90.91 $\pm$ 0.02} & \textbf{99.38 $\pm$ 0.04} & 91.03 $\pm$ 0.01 & \textbf{97.43 $\pm$ 0.15} & \multicolumn{1}{c|}{82.72 $\pm$ 0.05} & \textbf{98.46 $\pm$ 0.07} & 82.05 $\pm$ 0.04 \\

Can. Parl. & \textbf{98.24 $\pm$ 0.11} & \multicolumn{1}{c|}{67.19 $\pm$ 1.39} & \textbf{98.97 $\pm$ 0.06} & 76.51 $\pm$ 1.37 & \textbf{92.25 $\pm$ 0.24} & \multicolumn{1}{c|}{52.47 $\pm$ 0.88} & \textbf{95.06 $\pm$ 0.11} & 50.99 $\pm$ 1.72 \\

US Legis. & \textbf{76.74 $\pm$ 0.60} & \multicolumn{1}{c|}{60.98 $\pm$ 5.43} & \textbf{83.89 $\pm$ 0.43} & 65.76 $\pm$ 7.15 & \textbf{61.98 $\pm$ 1.31} & \multicolumn{1}{c|}{52.59 $\pm$ 2.13} & \textbf{66.18 $\pm$ 1.19} & 53.75 $\pm$ 1.07 \\

UN Trade & \textbf{75.84 $\pm$ 2.08} & \multicolumn{1}{c|}{62.04 $\pm$ 0.72} & \textbf{83.17 $\pm$ 1.28} & 65.91 $\pm$ 0.62 & \textbf{76.80 $\pm$ 2.01} & \multicolumn{1}{c|}{61.84 $\pm$ 0.53} & \textbf{83.95 $\pm$ 1.76} & 63.54 $\pm$ 0.54 \\

UN Vote & \textbf{73.40 $\pm$ 0.03} & \multicolumn{1}{c|}{52.29 $\pm$ 0.44} & \textbf{79.59 $\pm$ 0.02} & 52.55 $\pm$ 0.61 & \textbf{75.09 $\pm$ 0.34} & \multicolumn{1}{c|}{50.74 $\pm$ 0.82} & \textbf{82.01 $\pm$ 0.21} & 49.39 $\pm$ 0.75 \\

Contact & \textbf{98.16 $\pm$ 0.09} & \multicolumn{1}{c|}{91.39 $\pm$ 0.05} & \textbf{98.72 $\pm$ 0.06} & 93.57 $\pm$ 0.04 & \textbf{97.25 $\pm$ 0.07} & \multicolumn{1}{c|}{89.71 $\pm$ 0.05} & \textbf{97.99 $\pm$ 0.07} & 92.24 $\pm$ 0.05 \\ \hline
\end{tabular}
}
}
\label{tab:ablation_study_no_pe}
\end{table}

Combining the results from Tables~\ref{tab:ablation_study_1} and~\ref{tab:ablation_study_no_pe}, we discern that (1) in most cases, the full design of \name\ is the best and removing any component can lead to suboptimal results, (2)
in a few cases, only using LPE can also perform competitively, which suggests that the Learnable Positional Encoding module in our \name\ is relatively dominating than input node and edge features.

\subsection{Empirical results for parameter analysis}\label{app:param_analysis}

\subsubsection{Parameter Analysis for $t_{gap}$} \label{app:param_analysis_t_gap}
Here, we examine how varying the values of $t_{gap}$ affects the performance of \name\ under both \textit{transductive} and \textit{inductive} settings with random negative sampling.

\begin{table}[h]
\centering
\caption{AP and ROC-AUC  with various values of $t_{gap}$ under \textit{transductive setting} and random negative sampling.}
\vspace{1mm}
\begin{tabular}{c|c|cccc}
\hline
Dataset                  & Metric  & $t_{gap} = 10$            & $t_{gap} = 50$   & $t_{gap} = 1000$          & $t_{gap} = 2000$ \\ \hline
\multirow{2}{*}{Enron}   & AP      & 93.77 $\pm$ 0.28          & 93.82 $\pm$ 0.11 & \textbf{93.96 $\pm$ 0.36} & 93.70 $\pm$ 0.43 \\
                         & ROC-AUC & 95.83 $\pm$ 0.14          & 95.86 $\pm$ 0.07 & \textbf{95.98 $\pm$ 0.23} & 95.83 $\pm$ 0.21 \\ \hline
\multirow{2}{*}{UN Vote} & AP      & \textbf{73.40 $\pm$ 0.03} & 73.38 $\pm$ 0.07 & 73.39 $\pm$ 0.08          & 73.28 $\pm$ 0.33 \\
                         & ROC-AUC & \textbf{79.59 $\pm$ 0.02} & 79.54 $\pm$ 0.08 & 79.57 $\pm$ 0.06          & 79.53 $\pm$ 0.22 \\ \hline
\end{tabular}
\label{tab:param_analysis_transductive}
\end{table}

\begin{table}[h]
\centering
\caption{AP and ROC-AUC  with various values of $t_{gap}$ under \textit{inductive setting} and random negative sampling.}
\vspace{1mm}
\begin{tabular}{c|c|cccc}
\hline
Dataset                 & Metric  & $t_{gap} = 10$            & $t_{gap} = 50$   & $t_{gap} = 1000$          & $t_{gap} = 2000$ \\ \hline
\multirow{2}{*}{Enron}  & AP      & 88.43 $\pm$ 0.52          & 88.76 $\pm$ 0.62 & \textbf{89.17 $\pm$ 0.88} & 88.41 $\pm$ 0.21 \\
                        & ROC-AUC & 91.83 $\pm$ 0.37          & 92.02 $\pm$ 0.37 & \textbf{92.30 $\pm$ 0.57} & 91.86 $\pm$ 0.24 \\ \hline
\multirow{2}{*}{Un Vote} & AP      & \textbf{75.09 $\pm$ 0.34} & 75.01 $\pm$ 0.29 & 74.92 $\pm$ 0.48          & 74.73 $\pm$ 0.44 \\
                        & ROC-AUC & \textbf{82.01 $\pm$ 0.21} & 79.54 $\pm$ 0.08 & 81.85 $\pm$ 0.32          & 81.77 $\pm$ 0.16 \\ \hline
\end{tabular}
\label{tab:param_analysis_inductive}
\end{table}

We conduct the parameter analysis on two datasets: \textit{Enron} and \textit{UN Vote}, and adjusting $t_{gap}$ to different values: $\{10, 50, 1000, 2000\}$. Due to the page limit, we report the results in Table~\ref{tab:param_analysis_transductive} for \textit{transductive} and Table~\ref{tab:param_analysis_inductive} for \textit{inductive} setting, respectively. The best results are emphasized with \textbf{bold}. Firstly, it worth noting that the ratio 
\#Links / (Unique Steps) of \textit{UN Vote} is much larger than that of \textit{Enron}, since \textit{UN Vote} has significantly more links and fewer number of unique steps than \textit{Enron}, as stated in Table~\ref{tab:data_statistics}, implying \textit{UN Vote} is much denser that \textit{Enron}. From Tables~\ref{tab:param_analysis_transductive} and~\ref{tab:param_analysis_inductive}, we can see that the best performance on \textit{UN Vote} is derived from $t_{gap} = 10$, which is the smallest value, while the best result on \textit{Enron} is obtained by a much larger value of $t_{gap}$, which is $1000$. These findings indicate that a small $t_{gap}$ value functions effectively for dense graphs, whereas a larger $t_{gap}$ value is suited for sparse graphs. The detailed configurations are reported in Table~\ref{tab:hyperparams} further suggest this finding, as the $t_{gap}$ values associated with the sparse dataset (\textit{Wikipedia, Reddit, MOOC, LastFM, Enron, Social Evo., UCI}) mostly are in the range of $[1000, 2000]$, while the $t_{gap}$ values employed for \textit{Can. Parl., US Legis., UN Trade, UN Vote,} and \textit{Contact} are only less than or equal to $10$.

We report the results for the parameter analysis regarding how the performance change with different values of $t_{gap}$ Tables~\ref{tab:param_analysis_transductive} and~\ref{tab:param_analysis_inductive}.

\subsubsection{Parameter analysis for $K$} \label{app:param_analysis_K}

Next, we conduct parameter analysis for the neighborhood size, $K$, with random negative sampling, and report the results under \textit{transductive} and \textit{inductive} settings in Table~\ref{tab:param_analysis_neighborhood_transductive} and Table~\ref{tab:param_analysis_neighborhood_inductive}, respectively.

\begin{table}[h]
\centering
\caption{AP and ROC-AUC  with various values of $K$ under \textit{transductive setting} and random negative sampling.}
\vspace{1mm}
\begin{tabular}{c|c|cccc}
\hline
Dataset                  & Metric  & $K = 5$            & $K = 20$          & $K = 50$          & $K = 100$ \\ \hline
\multirow{2}{*}{Enron}   & AP      & 93.32 $\pm$ 0.25       &	\textbf{93.96 $\pm$ 0.36}    &	93.53 $\pm$ 0.51    &	93.22 $\pm$ 0.38 \\
                         & ROC-AUC & 95.50 $\pm$ 0.16       &	\textbf{95.98 $\pm$ 0.23}	&   95.71 $\pm$ 0.31	&   95.48 $\pm$ 0.19 \\ \hline
\multirow{2}{*}{UN Vote} & AP      & 72.08 $\pm$ 0.73       &	73.40 $\pm$ 0.03	&   \textbf{73.49 $\pm$ 0.10}	&   73.31 $\pm$ 0.13 \\
                         & ROC-AUC & 78.92 $\pm$ 0.33       & 	79.59 $\pm$ 0.02	&   \textbf{79.69 $\pm$ 0.10}	&   79.58 $\pm$ 0.08 \\ \hline
\end{tabular}
\label{tab:param_analysis_neighborhood_transductive}
\end{table}

\begin{table}[h]
\centering
\caption{AP and ROC-AUC  with various values of $K$ under \textit{inductive setting} and random negative sampling.}
\vspace{1mm}
\begin{tabular}{c|c|cccc}
\hline
Dataset                  & Metric  & $K = 5$            & $K = 20$   & $K = 50$          & $K = 100$ \\ \hline
\multirow{2}{*}{Enron}   & AP      & 89.72 $\pm$ 0.50       &	\textbf{94.60 $\pm$ 0.41}  &	 88.99 $\pm$ 1.04     &	88.29 $\pm$ 0.58 \\
                         & ROC-AUC & 92.65 $\pm$ 0.33       &	\textbf{95.88 $\pm$ 0.17}  &  92.11 $\pm$ 0.65     &	91.64 $\pm$ 0.36 \\ \hline
\multirow{2}{*}{UN Vote} & AP      & \textbf{77.76 $\pm$ 0.49}       &	75.09 $\pm$ 0.34  &	 75.10 $\pm$ 0.34     &	73.92 $\pm$ 0.75 \\
                         & ROC-AUC & 83.73 $\pm$ 0.23       &	\textbf{83.95 $\pm$ 1.76}  &	 81.96 $\pm$ 0.25     &	80.94 $\pm$ 0.82 \\ \hline
\end{tabular}
\label{tab:param_analysis_neighborhood_inductive}
\end{table}

According to the tables, we can discern that: (1) in a few cases, increasing $K$ can slightly increase the performance, but it is prone to cost more computational complexity; (2) in most cases, increasing $K$ does not add much to the performance, which suggests that close neighbors have already had enough information to support downstream tasks.

\subsubsection{Parameter analysis for $L$}

Here, we report the parameter analysis for $L$ as follows on Can. Parl. dataset in Table~\ref{tab:param_analysis_L}.

\textcolor{orange}{\begin{table}[!htbp]
\centering
\caption{AP and ROC-AUC  with various values of $L$ under \textit{transductive setting} and \textit{inductive setting} with random negative sampling.}
\vspace{1mm}
\resizebox{1.01\textwidth}{!}
{
\setlength{\tabcolsep}{0.9mm}
{
\begin{tabular}{c|c|c|c|c|c}
\hline
Dataset      &                & Transductive AP &	Transductive ROC-AUC &	Inductive AP &	Inductive ROC-AUC  \\ 
\hline
\multirow{4}{*}{Can. Parl}      & $L = 20$    & 98.24 $\pm$ 0.11	&98.97 $\pm$ 0.06&	92.25 $\pm$ 0.24&	95.06 $\pm$ 0.11 \\

& $L = 50$      & 98.26 $\pm$ 0.06&	98.94 $\pm$ 0.03&	92.15 $\pm$ 0.34&	95.11 $\pm$ 0.17\\ 

& $L = 100$      & 98.11 $\pm$ 0.21&	98.85 $\pm$ 0.11&	92.16 $\pm$ 0.98&	95.25 $\pm$ 0.56 \\

& $L = 200$      & 97.04 $\pm$ 0.22&	98.37 $\pm$ 0.09&	91.96 $\pm$ 0.59&	95.19 $\pm$ 0.38 \\

\hline
\end{tabular}
}
}
\label{tab:param_analysis_L}
\end{table}
}

From Table~\ref{tab:param_analysis_L}, we can observe that our proposed \name\ just relies on looking back a few historical information to make the correct information, as $L$ increases, the performance can be downgraded, and the best performance exists between $20$ and $50$, which discovery demonstrates that the near past history plays a more important role in the current decision making than the long past history.

\subsubsection{Parameter analysis for $\alpha_{neg}$}

We present the parameter analysis for $\alpha_{neg}$ on Can. Parl. dataset in Table~\ref{tab:param_analysis_neg}.

\begin{table}[!htbp]
\centering
\caption{AP and ROC-AUC  with various values of $\alpha_{neg}$ under \textit{transductive setting} and \textit{inductive setting} with random negative sampling.}
\vspace{1mm}
\resizebox{1.01\textwidth}{!}
{
\setlength{\tabcolsep}{0.9mm}
{
\begin{tabular}{c|c|c|c|c|c}
\hline
Dataset      &                & Transductive AP &	Transductive ROC-AUC &	Inductive AP &	Inductive ROC-AUC  \\ 
\hline
\multirow{5}{*}{Can. Parl}      & $\alpha_{neg} = 0.05$    & 98.26 $\pm$ 0.13 &	98.98 $\pm$ 0.05 &	92.24 $\pm$ 0.09 &	95.09 $\pm$ 0.11 \\

& $\alpha_{neg} = 0.1$    & 98.28 $\pm$ 0.09 &	98.98 $\pm$ 0.05 &	92.35 $\pm$ 0.21 &	95.09 $\pm$ 0.09 \\

& $\alpha_{neg} = 0.3$      & 98.24 $\pm$ 0.11 &	98.97 $\pm$ 0.06&	92.25 $\pm$ 0.24&	95.06 $\pm$ 0.11\\

& $\alpha_{neg} = 0.5$      & 98.28 $\pm$ 0.13&	98.97 $\pm$ 0.05&	91.48 $\pm$ 0.23&	94.75 $\pm$ 0.16
\\

& $\alpha_{neg} = 0.95$      & 98.24 $\pm$ 0.18&	98.95 $\pm$ 0.09&	91.09 $\pm$ 0.49&	94.53 $\pm$ 0.32\\

\hline
\end{tabular}
}
}
\label{tab:param_analysis_neg}
\end{table}

In general, Table~\ref{tab:param_analysis_neg} shows (1) our proposed method \name\ is robust towards different choices of $\alpha_{neg}$, and (2)
usually a smaller weight of negative pairs can help the model concentrate more on the positive pairs and boost the performance.

\subsubsection{Parameter analysis for $\alpha_{pe}$}

We present the parameter analysis for $\alpha_{neg}$ on Enron dataset in Table~\ref{tab:param_analysis_pe_weight}.

\begin{table}[!htbp]
\centering
\caption{AP and ROC-AUC  with various values of $\alpha_{pe}$ under \textit{transductive setting} and \textit{inductive setting} with random negative sampling.}
\vspace{1mm}
\resizebox{1.01\textwidth}{!}
{
\setlength{\tabcolsep}{0.9mm}
{
\begin{tabular}{c|c|c|c|c|c}
\hline
Dataset      &                & Transductive AP &	Transductive ROC-AUC &	Inductive AP &	Inductive ROC-AUC  \\ 
\hline
\multirow{4}{*}{Enron}      & $\alpha_{pe} = 0.05$    & 92.81 $\pm$ 0.82 &	95.32 $\pm$ 0.38&	86.49 $\pm$ 1.52&	90.91 $\pm$ 0.61 \\

& $\alpha_{pe} = 0.3$      & 93.96 $\pm$ 0.36&	95.98 $\pm$ 0.23&	89.17 $\pm$ 0.88&	92.30 $\pm$ 0.57\\

& $\alpha_{pe} = 0.5$      & 93.94 $\pm$ 0.10&	95.90 $\pm$ 0.06&	88.92 $\pm$ 0.39&	92.12 $\pm$ 0.22
\\

& $\alpha_{pe} = 0.95$      &  94.00 $\pm$ 0.28	&95.92 $\pm$ 0.17&	88.67 $\pm$ 0.34&	91.91 $\pm$ 0.30\\

\hline
\end{tabular}
}
}
\label{tab:param_analysis_pe_weight}
\end{table}

$\alpha_{neg}$ stands for the weight of the loss function of learnable positional encoding. As shown in the Table~\ref{tab:param_analysis_pe_weight}, we can observe that (1) our \name\ is robust towards different choices of $\alpha_{pe}$, and (2) a considerable weight, e.g., $0.5$, is optimal, too large or too small weight can induce suboptimal results.

\subsection{Different positional encoding initialization}
\label{app:different_pe}
In this section, we demonstrate that \name~can be equipped with different positional encoding method, other than Laplacian PE. Specifically, our work aims to introduce a general framework on how to approximate positional encodings at a current time solely based on information from previous timestamps to outperform in temporal link prediction task, and we do not want to constraint the type of positional encodings. In Table~\ref{tab:rwpe}, we provide the comparison between initializing the positional encodings at initial timestamp with Laplacian PE and Random Walk PE \citep{DBLP:conf/iclr/DwivediL0BB22} under two settings, \textit{transductive} and \textit{inductive}, with random negative sampling.

\begin{table}[!htbp]
\centering
\caption{Comparison in Positional Encoding initialization with Laplacian and Random Walk Positional Encoding under \textit{transductive} and \textit{inductive} settings with random negative sampling.}
\vspace{1mm}
\resizebox{1.01\textwidth}{!}
{
\setlength{\tabcolsep}{0.9mm}
{
\begin{tabular}{c|c|c|c|c|c}
\hline
Dataset      & Positional Encoding               & Transductive AP &	Transductive ROC-AUC &	Inductive AP &	Inductive ROC-AUC  \\ 
\hline
\multirow{2}{*}{Enron}      & Random Walk PE    & 93.79 $\pm$ 0.67 &	95.88 $\pm$ 0.40 &	89.11 $\pm$ 0.83 &	92.43 $\pm$ 0.64 \\
                            & Laplacian PE      & 93.96 $\pm$ 0.36 &	95.98 $\pm$ 0.23 &	89.17 $\pm$ 0.88 &	92.30 $\pm$ 0.57  \\ 
\hline
\end{tabular}
}
}
\label{tab:rwpe}
\end{table}

Based on Table~\ref{tab:rwpe}, we can discern that (1) \name~can handle different positional encodings and perform robustly; (2) different positional encodings bring slightly different performances and finding a powerful positional encoding is a promising future direction to trigger more interesting works.

\section{Reproducibility}
\label{app:reproducibility}

\subsection{Different Negative Sampling Strategies (NSS)}
\label{app:nss}
In brief, random NSS samples possible node pairs uniformly at random, historical NSS samples negative edges from the edges occurring in past timestamps but are absent in the present time, and inductive NSS samples edges that are unseen during the training time. We refer readers to \citep{DBLP:conf/nips/PoursafaeiHPR22} for more details regarding these three sampling strategies.




\textbf{Configuration and implementation details.} Following the training procedure of \citep{DBLP:conf/nips/0004S0L23}, \name\ is trained with a maximum of $200$ epochs, and we employ early stopping with patience $10$ during the training process. We leverage the Adam optimizer with learning rate of $0.0001$. For a more detailed description of the model's implementation, computational resources, and configurations of hyper-parameters over all $13$ datasets, we refer readers to Appendix~\ref{app:hyper-param},~\ref{app:evaluation_ctdg}.

Across Table~\ref{tab:transductive_random}, Table~\ref{tab:inductive_random}, Table~\ref{tab:transductive_hist}, Table~\ref{tab:inductive_hist}, Table~\ref{tab:transductive_ind}, and Table~\ref{tab:inductive_ind}, the model with the best performance, in terms of metric scores AP and AUC-ROC, on the validation set will be selected for testing. We run \name\ 5 times with different random seeds from $0$ to $4$ and report the average metric score. Results from all other baselines are also obtained in the same manner \citep{DBLP:conf/nips/0004S0L23}.    

\subsection{Model Configurations, Hyper-parameters, and Computing Resources} \label{app:hyper-param}

We first report the configuration and hyper-parameters that are unchanged for all $13$ datasets: 
\begin{itemize}
    \item Dimension of time encoding: $d_T = 100$.
    \item Dimension of node encoding: $d_N = 172$.
    \item Dimension of edge encoding: $d_E = 172$.
    \item Dimension of positional encoding: $d_P = 172$ (only for \textit{Social Evo.}, $d_P = 72$).
    \item Hyper-parameters for time encoding function: $\alpha = 10, \beta 
    = 10$.
    \item Weight of negative samples in positional encoding loss: $\alpha_{neg} = 0.3$.
    \item Weight of positional encoding loss in objective loss function of \name\: $\alpha_{pe} = 0.5$.
\end{itemize}

\begin{table}[!htbp]
\centering
\caption{Configurations of the number of recent snapshots for computing LPE, time window for node features, recent interactions for edge features, and batch size over $13$ datasets}
\vspace{1mm}
\begin{tabular}{c|cccc}
\hline
Dataset    & $L$ & $t_{gap}$ & $K$ & Batch size \\ \hline
Wikipedia  & 100 & 1000      & 15  & 128        \\
Reddit     & 100 & 1000      & 20  & 200        \\
MOOC       & 100 & 2000      & 30  & 128        \\
Lastfm     & 100 & 1000      & 30  & 128        \\
Enron      & 100 & 1000      & 20  & 64         \\
Social Evo.  & 100 & 1000      & 20  & 128        \\
UCI        & 200 & 500       & 30  & 100        \\
Flights    & 100 & 1000      & 30  & 128        \\
Can. Parl. & 20  & 2         & 10  & 64         \\
US Legis.   & 50  & 2         & 10  & 200        \\
UN Trade    & 200 & 6         & 30  & 200        \\
UN Vote     & 100 & 10        & 20  & 128        \\
Contact    & 200 & 10        & 20  & 128        \\ \hline
\end{tabular}
\label{tab:hyperparams}
\end{table}

The experiments are coded by Python and are performed on a Linux machine with a single NVIDIA Tesla V100 32GB GPU. The code will be released upon paper's publication.

Next, for reproducibility, we present detailed hyper-parameters across $13$ datasets in Table~\ref{tab:hyperparams}.

\subsection{Running \name~on Continuous Time Dynamic Graphs}
\label{app:evaluation_ctdg}

The definition of discrete time and continuous time dynamic graphs can be referred to the survey paper~\citep{DBLP:journals/jmlr/KazemiGJKSFP20}, where
\begin{itemize}
    \item Continuous Time Dynamic Graph (CTDG) is represented as $((v_2, v_5), t_1)$, $((v_1, v_2), t_2)$, …, as shown in its Example 2;
    \item Discrete Time Dynamic Graph (DTDG) is represented as a set of $G_1, G_2, …, G_T$, where $G_t = (V_t, E_t)$ is the graph at snapshot $t$, $V_t$ is the set of nodes in $G_t$, and $E_t$ is the set of edges in $G_t$.
\end{itemize}

In this viewpoint, our method is designed for discrete time. However, we can see a clear transformation between CTDG and DTDG: if we aggregate edges that happen at the same timestamp $t$ into a set, then it is the graph snapshot $G_t$. This is not invented by us, a similar example can be seen in Example 3 of the survey paper~\citep{DBLP:journals/jmlr/KazemiGJKSFP20}.

Because the introduction and theoretical derivation on the snapshot level are much easier to understand by people, and using “a set of continuous-time edges” every time when we do theoretical derivation and illustration can be wordy and hurt the presentation. Since graph snapshot and relevant concepts are already existing in the community and cover the meaning, then we choose to follow them.

Then, in this section, we elaborate how \name~can be evaluated on Continuous Time Dynamic Graphs. Firstly, we explain the data batching procedure for Continuous Time Dynamic Graphs as follows. Suppose we are given a CTDG with $T$ interactions, $G = \{u_i, v_i, t_i\}_{i = 1}^T$, where $t_1 \leq \dots t_T$ and $(u_i, v_i, t_i)$ denotes the link occurrence between $u_i, v_i$ at time $t_i$. Let the batch size be $B$, then a data batch would be $B$ consecutive interactions from $\{u_i, v_i, t_i\}_{i = 1}^T$. Specifically, suppose $T = B \cdot K$ (for some $K$), then our $1$st data batch would be the first $B$ link occurrences of $G$: $\{u_i, v_i, t_i\}_{i = 1}^B$. The $2$nd data batch would be the next $B$ consecutive interactions, $\{u_i, v_i, t_i\}_{i = B + 1}^{2B}$, and so on. In this way, now we obtain $K$ data batches. Moreover, as a batch is a stream of $B$ events, so we can regard a data batch as a mini temporal graph with $B$ link occurrences. This data batching technique is also employed by other baselines.

Next, we explain the evaluation process of \name. Consider an arbitrary data batch $\{u_j, v_j, \tau_j\}_{j = 1}^B$. If this is the first data batch, then we first extract all links occurrence (without the timestamps), $\{u_j, v_j\}_{j = 1}^B$, transform this to a static graph, and obtain the Laplacian eigenvectors of this graph to initialize the positional encoding for each node. If a node of $G$ does not belong to this graph then we initialize the positional encoding of that node with a zero vector. Here, let the "initial" positional encoding for node $u$ as $\pe^{1}_u$. As we advance to the $2$nd data batch, we first compute the approximated positional encoding for all nodes, $\approxpe^2_{.}$, as stated in Eq.~\ref{eq:transform_pe},~\ref{eq:approx_pe}, based on $\pe^1_{.}$. In this way, following Eq.~\ref{eq:transform_pe},~\ref{eq:approx_pe}, we can iteratively compute the approximate positional encoding for the $k$-th batch, $\approxpe^{k}_{.}$, based on the sequence of previous $L$ positional encodings, $\pe^{k - L}_{.}, \dots, \pe^{k - 1}_{.}$, corresponding to previous $L$ data batches.

For our current data batch (suppose this is the $k$-th batch), $\{u_j, v_j, \tau_j\}_{j = 1}^B$, we now describe how to obtain the link predictions. For each query $(u_j, v_j, \tau_j)$, we obtain $\h^{\tau_j}_{u_j, N || E}$, as defined in Eq.~\ref{eq:node-link-encoding}, and $\h^{\tau_j}_{u_j, P}$, as stated in Eq.~\ref{eq: pe}, using $\approxpe^{k}_{u_j}$. Finally, we combine $\h^{\tau_j}_{u_j, N || E}, \h^{\tau_j}_{u_j, P}$ to obtain the temporal representation at time $\tau_j$ for $u_j$, $\h^{\tau_j}_{u_j}$. Similarly, we obtain the temporal representation for $v_j$, $h^{\tau_j}_{v_j}$. Finally, following Eq.~\ref{eq: link_prediction}, we obtain the link prediction for $(u_j, v_j, \tau_j)$ using $\h^{\tau_{j}}_{u_j}, \h^{\tau_j}_{v_j}$.
  
In summary, we ideally would want to obtain the positional encoding of a node $u$, $\pe^t_{u} \in \mathbb{R}^{d_P}$ for every temporal graph snapshot. However, for training efficiency, our implementation computes the positional encoding for the graph corresponding to a data batch.


\end{document}